\documentclass[floatfix,rmp,twocolumn,twoside,a4paper]{revtex4}
\usepackage{amsmath}
\usepackage{amsfonts}
\usepackage{amsthm}
\usepackage{reptheorem}

\usepackage{graphicx} 
\usepackage[english]{babel}
\usepackage{makeidx}

\usepackage{booktabs}
\usepackage{cellspace}

\usepackage{hyperref}
\usepackage{xcolor}

\definecolor{darkblue}{RGB}{0,0,192}
\hypersetup{
    colorlinks=true,
    linkcolor=darkblue,
    filecolor=magenta,      
    urlcolor=blue,
    citecolor=blue,
    }
    
\newfont{\chancery}{pzcmi scaled 1200}

\newcommand{\textquote}[3] {
   \begin{quotation}
      {\chancery #1}
      \begin{flushright}
        \normalsize \textsc{#2} \\ {\em #3}
      \end{flushright}
   \end{quotation}
}

\newtheoremstyle{mystyle}
  {6pt}
  {6pt}
  {\it}
  {}
  {\bf}
  {}
  { }
  {{\sc \thmname{#1} \thmnumber{#2}:\thmnote{ #3}}}
\theoremstyle{mystyle}

\newtheorem{theorem}{Theorem}[section]
\newtheorem{lemma}[theorem]{Lemma}
\newtheorem{corollary}[theorem]{Corollary}
\newtheorem{proposition}{Proposition}

\newtheorem{definition}{Definition}[section]

\theoremstyle{remark}
\newtheorem*{remark}{Remark}

\bibpunct{(}{)}{,}{n}{}{,}
\addto\captionsenglish{
}

\makeatletter
\renewcommand\tableofcontents{%
    \@starttoc{toc}%
}
\makeatother

\makeindex
        
\begin{document}

\title{DL101 The Representation Capability of Neural Network Models}
\author{Fernando Berzal \\ \url{berzal@acm.org} }
\affiliation{Department of Computer Science and Artificial Intelligence, University of Granada, Spain}

\maketitle

\tableofcontents
\vspace{6mm}

\vspace{6mm}

\textquote{
All models are wrong, but some are useful.
}{George E.P. Box}{}

Deep learning models \cite{goodfellow2016,berzal2019a,berzal2019b,aggarwal2023,bishop2024}
have a layered structure. 
A feedforward deep neural network with \(L\) layers can be defined as a function  
\[
f: \mathbb{R}^{n_0} \to \mathbb{R}^{n_L}
\]  
constructed by composing affine transformations and nonlinear activations:
\[
f(x) = f^{(L)} \circ f^{(L-1)} \circ \cdots \circ f^{(1)}(x),
\]
For each layer \(l = 1, \dots, L\),
\[
f^{(l)}(x) = \sigma^{(l)}(W^{(l)} x + b^{(l)}).
\]
where 
\(x \in \mathbb{R}^{n_0}\) is an input vector, 
\(n_l\) is the number of units (width) in layer \(l\)  ,
\(W^{(l)} \in \mathbb{R}^{n_l \times n_{l-1}}\) is the weight matrix of layer \(l\),
\(b^{(l)} \in \mathbb{R}^{n_l}\) is the bias vector of layer \(l\),
and
\(\sigma^{(l)}: \mathbb{R} \to \mathbb{R}\) is the activation function  of layer \(l\), which is applied element-wise.

The depth of the network is \(L\), and the parameters of the deep learning model are the collection of all weights and biases:  
\[
\theta = \{ W^{(l)}, b^{(l)} \}_{l=1}^L.
\]

Training a deep learning model consists of finding \(\theta\) that minimizes a loss function \( \mathcal{L}(f(x), y) \) over a dataset \(\{(x_i, y_i)\}\).

Formally, a deep learning model is just a composition of affine maps and nonlinearities parameterized by weights and biases, which are optimized via loss minimization. As we will see, multilayered neural networks are universal approximators \cite{cybenko1989, funahashi1989, hornik1989, leshno1993, montufar2014}.

\begin{figure*}[h!t!]
    \centering
    \includegraphics[width=0.9\linewidth]{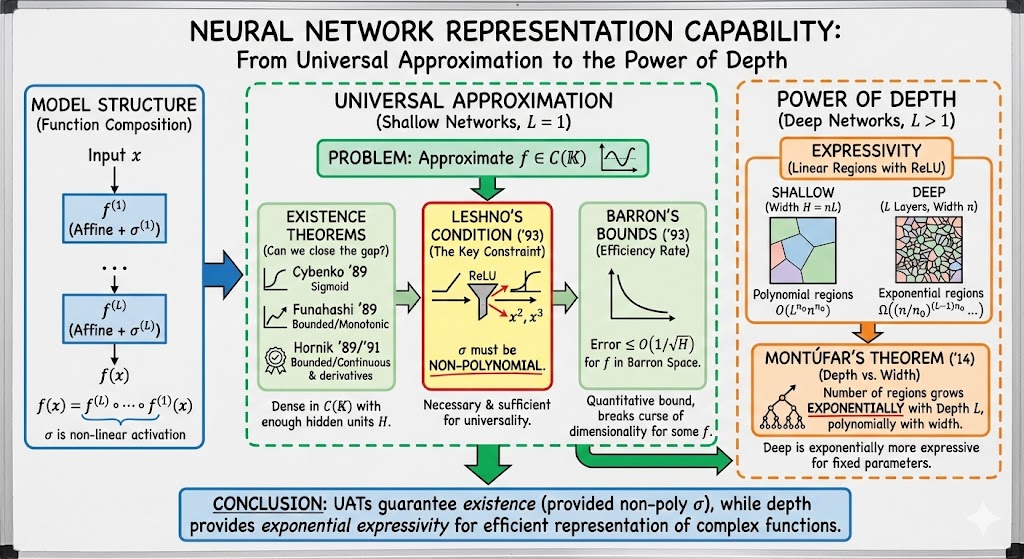}
    \caption{Main results discussed in this technical report, according to Google Nano Banana.}
    \label{fig:placeholder}
\end{figure*}

\section{Universal Approximators}

Let us consider the problem of representing general functions of an n-dimensional real variable, $x \in \mathbb{R}^n$, by finite linear combinations:
$$f: \mathbb{R}^n \to \mathbb{R}$$
$$f(x) = \sum_{j=1}^h \alpha_j \sigma \left( \sum_{i=1}^n w_{ij} x_i  + b_j \right) $$
which corresponds to a feed-forward neural network with $n$ inputs, $h$ hidden neurons, and a single output $y=f(x)$. $W=[w_{ij}]$ is the $n \times h$ weight matrix for the hidden layer, $(b_j)$ is the $h \times 1$ vector of biases for the hidden neurons, and $(\alpha_j)$ is the $h \times 1$ weight vector of the output neuron. Finally, $\sigma$ is the univariate activation function of the hidden units. 

\subsection*{Cybenko's Universal Approximation Theorem}

If $\sigma(z)$ is a sigmoidal function, i.e. $\lim_{z \to +\infty} \sigma(z) =1$ and $\lim_{z \to -\infty} \sigma(z) =0$ (as in the logistic function), George Cybenko \cite{cybenko1989} proved that functions defined as $f(x)$ are dense in the space of continuous functions on the unit cube if $\sigma$ is any continuous sigmoidal function, such as the logistic function. Cybenko, then at the University of Illinois at Urbana-Champaign, proved the universal approximation theorem for artificial neural networks with sigmoid activation functions:

\begin{makethm}{theorem}{thm:cybenko}[Universal approximation theorem for artificial neural networks with sigmoid activation functions]  
Let $\sigma$ be any continuous discriminatory function. The finite sums of the form
$$ G(x) = \sum_{j=1}^h \alpha_j \sigma \left( \sum_{i=1}^n w_{ij} x_i  + b_j \right) $$
are dense in $C(I_n)$. In other words, given any $f \in C(I_n)$ and $\epsilon>0$, there is a sum, $G(x)$, of the above form, for which
$$ |G(x)-f(x)| < \epsilon \quad \forall x \in I_n$$
\end{makethm}

$I_n$ is the n-dimensional unit hypercube, whereas $C(I_n)$ is the space of continuous functions in $I_n$. The theorem above states that finite superpositions of a fixed, univariate function that is discriminatory can uniformly approximate any continuous function of $n$ real variables with support in the unit hypercube.

Given that sigmoid activation functions are continuous and discriminatory (lemma \ref{lemma:sigmoidal-discriminatory}), any continuous function can be uniformly approximated by a continuous neural network having only one hidden sigmoid layer with an arbitrary continuous sigmoidal nonlinearity. Therefore, a feed-forward neural network with a single hidden sigmoid layer, acting as a universal approximator, can represent any continuous function, provided that we have enough hidden units.

\subsection*{Funahashi's Universal Approximation Theorem}

Independently from Cybenko, Ken-ichi Funahashi \cite{funahashi1989} also proved that any continuous mapping can be approximately realized multilayer neural networks with at least one hidden layer with sigmoid activation functions.

\begin{makethm}{theorem}{thm:funahashi}[Universal approximation theorem for artificial neural networks with bounded, monotonically increasing activation functions]  
Let $\phi(x)$ be a non-constant, bounded, and monotonically increasing continuous function. Let $K$ be a compact subset (bounded closed subset) of $\mathbb{R}^n$ and $f$ a real valued continuous function on $K$. Then, for an arbitrary $\epsilon > 0$, there exists an integer $N$ and real constants $\alpha_i$, $b_i$, and $w_ij$ such that
$$
\tilde{f}(x) = \sum_{j=1}^h \alpha_j \phi\left( \sum_{i=1}^n w_{ij} x_i + b_j \right) 
$$
satisfies $\max_{k\in K} |f(x)-\tilde{f}(x)| < \epsilon$
\end{makethm}

The above theorem proves that, for an arbitrary $\epsilon>0$, there exists
a three-layer network, with a $\phi(x)$ activation function in the hidden layer and a linear output layer, whose output function $\tilde{f}(x)$ approximates $f(x)$ as close as needed. Obviously, the sigmoid function $\sigma(z)=1/(1-e^{-z})$ typically used in neural networks satisfies the requirements of $\phi(x)$, i.e. non-constant, bounded, and monotonically increasing.

\begin{table*}[!ht]
\begin{tabular}{SlScScSc} 
\toprule[1.5pt]
 & Activation function & Range & Derivative \\ 
\midrule[1pt]
Linear (identity)  
  & $f(z) = z$ 
  & $(-\infty, \infty) $ 
  & $f'(z) = 1$ \\
Logistic (standard sigmoid)
  & $f(z) = \sigma(z) = \dfrac{1}{1+e^{-z}}$ 
  & $(0,1)$ 
  & $f'(z) = f(z) (1- f(z))$\\
Hyperbolic tangent (tanh)
  & $f(z) = \tanh(z) = \dfrac{e^{z} - e^{-z}}{e^{z} + e^{-z}}$ 
  & $(-1,1)$
  & $f'(z) = 1- f^2(z) $ \\ 
Softmax
  & $ f_i(\vec{z}) = \dfrac{e^{z_i}}{\sum_{j=1}^{K} e^{z_j}}$
  & [0,1] 
  & $\dfrac{\partial y_i}{\partial x_j} = \dfrac{\partial f_i(\vec{z})}{\partial x_j}  = y_i ( \delta_{ij} - y_j)$ \\
Softplus
  & $f(z) = \ln (1+e^z)$
  & $[0, \infty) $
  & $f'(z) = \dfrac{e^x}{1 + e^{z}} = \dfrac{1}{1+e^{-z}} = \sigma(z) $ \\ 
ReLU    
  & $f(z) = \max \{0, z \}$
  & $[0, \infty) $
  & $ 
    f'(z) = H(z) = 
        \begin{cases} 
            1 & \text{if } z > 0 \\ 
            0 & \text{if } z \le 0
        \end{cases}
    $ \\
Leaky ReLU 
  & $ 
    f(z) = 
        \begin{cases} 
            z & \text{if } z > 0 \\ 
            \alpha z & \text{if } z \le 0
        \end{cases}
    $
  & $(-\infty, \infty)$ 
  & $ 
    f'(z) = 
        \begin{cases} 
            1 & \text{if } z > 0 \\ 
            -\alpha & \text{if } z \le 0
        \end{cases}
    $ \\
\bottomrule[1.5pt]
\end{tabular}
\caption{Common activation functions in neural networks.}
\label{activation}
\end{table*}

\begin{remark}
Funahashi's theorem uses bounded functions. Rectified linear units, $ReLU(z)=max\{0,z\}$, are not bounded. However, the linear combination $\phi(z) = ReLU(z-1)-ReLU(z)$ falls within the scope of Funahashi's theorem, so its conclusion is also valid for $\phi(z)=ReLU(z)$.
\end{remark}

While Cybenko's and Funahashi's techniques differs, both support the same conclusion for sigmoidal activations. Funahashi's proof guarantees the existence of parameters approximating arbitrary spectra, paralleling Cybenko’s measure-separation proof. Both are non-constructive in the sense of parameter selection.

\section{Alternative Activation Functions}

Apart from sigmoid functions, such as the logistic function, and the hyperbolic tangent, other activation functions are common in neural networks (see Table \ref{activation}). Cybenko and Funahashi proved the universal approximation capabilities of multi-layered neural networks when using sigmoid activation functions in the hidden layer. What happens with other activation functions?

\subsection*{Hornik's Approximation Theorems}

The Hornik theorems \cite{hornik1989,hornik1991} formalize the universal approximation capabilities of multilayer feedforward neural networks for bounded and continuous activation functions:

\begin{makethm}{theorem}{thm:hornik}[Universal approximation theorem for artificial neural networks with bounded and continuous activation functions]  
Let \(\sigma: \mathbb{R} \to \mathbb{R}\) be a bounded, non-constant, and continuous activation function. Then, for any compact set \(K \subset \mathbb{R}^n\), the set of functions of the form
$$
H(x) =  \sum_{j=1}^h \alpha_j \sigma \left( \sum_{i=1}^n w_{ij} x_i  + b_j \right)
$$  
is dense in $C(K)$, the space of continuous functions on $K$ with the uniform norm.
\end{makethm}



Hornik's results \cite{hornik1989} build on the Stone–Weierstrass theorem, which states that certain algebraic structures (like neural networks with nonlinear activations) can approximate any continuous function. The neural network acts as a flexible basis expansion, where the hidden units serve as nonlinear basis functions.

A feedforward neural network with a single hidden layer using a non-constant, bounded, and continuous activation function can approximate any continuous function on a compact subset $K \subset \mathbb{R}^n$ to arbitrary precision. 

Not only can such networks approximate functions, but they can also approximate their derivatives \cite{hornik1991}:

\begin{makethm}{theorem}{thm:hornik-derivative}[Universal approximation theorem for derivatives]  
Let \(\sigma: \mathbb{R} \to \mathbb{R}\) be a bounded, non-constant, and continuous activation function.
Then, for any \(f \in C^k(K)\), the space of \(k\)-times continuously differentiable functions on compact \(K \subset \mathbb{R}^n\), there exists a neural network $\tilde{f}_N$ such that  
$$
\| f(x) - \tilde{f}_N(x) \|_{C^k(K)} < \varepsilon
$$ 
for any \(\varepsilon > 0\). 
\end{makethm}

\subsection*{Leshno's Approximation Theorem}

Hornik’s approximation theorems are sufficient: they tells us when universal approximation is possible. Leshno’s approximation theorem is necessary and sufficient: it tells us exactly which activation functions work and which don’t.
So while Hornik opened the door to understanding neural networks as universal approximators, Leshno sharpened the criteria and closed the door on polynomial activations \cite{leshno1993}:


\begin{makethm}{theorem}{thm:leshno}[Leshno's theorem: Universal approximation theorem for artificial neural networks with non-polynomial activation functions]  
Let \(\sigma: \mathbb{R} \to \mathbb{R}\) be a continuous activation function. Then the set of functions defined by 
$$
L(x) = \sum_{j=1}^h \alpha_j \sigma \left( \sum_{i=1}^n w_{ij} x_i  + b_j \right)
$$  
is dense in \(C(K)\) for every compact \(K \subset \mathbb{R}^n\) if and only if \(\sigma\) is not a polynomial almost everywhere.
\end{makethm}


Leshno used algebraic and functional analysis to prove non-polynomiality is essential.  The universal approximation property holds for a wide class of activation functions, not just sigmoids. The key requirement is that the activation function must be non-polynomial. If \(\sigma\) is a polynomial (e.g., \(\sigma(x) = x^2\)), then the multilayered network cannot universally approximate. If \(\sigma\) is non-polynomial, even if unbounded or discontinuous in places (e.g., ReLU), it can be a universal approximator.

As a consequence of Leshno's theorem, ReLU, softplus, and many others activation functions qualify. Since they satisfy the non-polynomial constraint, they can be used as universal approximators.

\subsection*{Barron's Approximation Error Bounds}

Andrew Barron, also from the University of Illinois at Urbana Champaign, provided explicit convergence rates for approximation in \(L^2\), unlike previous results, which are existence theorems.

\begin{makethm}{theorem}{thm:barron}[Barron’s Theorem]
Let \(f: \mathbb{R}^n \to \mathbb{R}\) be a function with Fourier transform \(\mathcal{F}\) such that its Barron norm is finite, i.e.
\[
\int_{\mathbb{R}^n} \|\omega\| |\mathcal{F}(\omega)| \, d\omega < \infty.
\]
Then there exists a single-hidden-layer neural network \(f_H\) with \(H\) hidden units such that
\[
\|f - f_H\|_{L^2} \leq \frac{C}{\sqrt{H}},
\]
where \(C\) depends on the Barron norm of \(f\).
\end{makethm}

Cybenko, Funahashi, Hornik, and Leshno proved that neural networks are universal approximators, but did not quantify how fast the approximation improves with more neurons. Barron \cite{barron1993} shows that, for a specific class of functions (Barron space), the approximation error decays as \(O(1/\sqrt{H})\),

Leshno’s condition (non-polynomial activation) is still necessary for Barron’s bounds to hold (polynomial activations cannot generate the required Fourier richness). 

The Barron space is a strict subset of \(L^2\), characterized by Fourier decay and low-frequency dominance. Functions in Barron space can be approximated efficiently, even in high dimensions, breaking the curse of dimensionality.


\section{The Power of Deep Networks}

The universal approximation properties of neural networks ar powerful, but focus only on existence. An important question is ``how many neurons are required to yield an approximation of a given quality?''

Funahashi \cite{funahashi1989} already suggested that ``it should be studied theoretically studied in the future that the possibility of $k > 3$-layer networks can realize a given mapping with less costs (number of units or connections) than three-layer networks, within error $\epsilon$.''

Guido Montúfar \cite{montufar2014} studied the expressive power of deep feedforward neural networks, particularly focusing on piecewise linear activations like ReLU.


Consider a feedforward neural network with input dimension $n_0$ and $L$ hidden layers with widths $n_1..n_L$ using the ReLU activation function (i.e. $\sigma(z)=\max \{0, z\}$).
The network computes a piecewise linear function $f: \mathbb{R}^{n_0} \to \mathbb{R}$, partitioning the input space into linear regions where $f$is affine.
(i.e. linear regions correspond to distinct affine pieces of the function computed by the network).

\begin{makethm}{theorem}{thm:montufar}[Expressivity via Linear Regions]
Let \( N \) be a feedforward ReLU network with input dimension \( n_0 \), \( L \) hidden layers, and widths \( n_1.. n_L \), $n_l \ge n_0$ $\forall l$. Then, 
the maximal number of linear regions of the function computed by the neural network $N$ is lower bounded by
$$
\left(\prod_{l=1}^{L-1} \left\lfloor \frac{n_l}{n_0} \right\rfloor^{n_{0}} \right) \sum_{j=0}^{n_0} {n_L \choose j}
$$
\end{makethm}

Assuming $n_0 \in O(1)$ and $n_i=n \quad\forall i\ge1$, the number of regions of a single hidden layer model with $Ln$ hidden units behaves as $O(L^{n_0} n^{n_0})$. A deep ReLU network can compute functions that have $\Omega((n/n_0)^{(L-1)n_0} n^{n_0})$ linear regions. 

Montúfar's lower bound shows that depth increases expressivity exponentially, while width contributes polynomially.
The number of regions grows exponentially with depth, confirming that, using the same number of parameters, deep networks are exponentially more expressive than shallow ones. The number of linear regions of deep models grows exponentially in $L$ and polynomially in $n$, which is much faster than that of shallow models with $nL$ hidden units ($(n/n_0)^{(L-1)n_0}$ vs. $L^{n_0}$).

Depth is not just convenient, it is formally necessary for efficient representation of certain functions: some functions can be represented efficiently by deep networks but require exponentially more units in shallow networks \cite{telgarsky2016}.

\cite{serra2018} provides tighter upper and lower bounds on the number of linear regions achievable by ReLU networks.  

Montúfar quantified expressivity via combinatorial complexity (number of linear regions), but there are alternative measures of expressivity beyond linear region counting \cite{raghu2017}. Depth increases expressivity in a way consistent with Montúfar’s exponential linear region separation.  


\section{Conclusion}

Universal approximation theorems (Cybenko, Funahashi, Hornik \& Leshno) provide the mathematical justification for why neural networks are powerful enough to model complex, nonlinear phenomena. Shallow networks can approximate any continuous function on compact sets. Leshno’s theorem clarified that non-polynomial activations are necessary for universality, which explains why ReLU, logistic, hyperbolic tangent, and other common activations work (but pure polynomials do not).

However, approximation theorems provide no efficiency, training, nor generalization guarantees. They do not indicate how many neurons are needed for a given accuracy, whether gradient descent will find the right parameters, or whether networks trained on finite data will generalize well.

Barron provided quantitative error bounds for particular function classes (approximation rates for functions with bounded Fourier norm), whereas Montúfar was the first to show that expressivity grows exponentially with depth (in terms of the number of linear regions). Deep networks are strictly more expressive than shallow ones with the same number of parameters. Depth is fundamentally more powerful than width, providing a rigorous mathematical explanation for the use of deep learning models.

\appendix

\section{Mathematical Background}

\begin{definition}
The n-dimensional unit hypercube is $I_n = [0,1]^n$. More precisely,
\begin{align*} 
  I_n &= [0,1]^n \\
      &= \{(x_1,\ldots,x_n)\in\mathbb{R}^n \;|\; 0\le x_i\le 1,\; i=1,\ldots,n\}
\end{align*}

\end{definition}

\begin{definition} \label{def:CIn}
Let $\mathbb{K}$ denote the field of scalars, typically $\mathbb{R}$ or $\mathbb{C}$. The space of continuous functions on $I_n$ is 
\[ C(I_n) = \{f:I_n\rightarrow\mathbb{K} \;|\; f \text{ is continuous}\}. \] 
Under point-wise addition and scalar multiplication, $C(I_n)$ is a vector space. 
\end{definition}

\begin{remark}
$C(\mathbb R^n)$ is the set of all continuous functions on $\mathbb R^n$, whereas $C^\infty(\mathbb R^n)$ is the set of all infinitely differentiable functions on $\mathbb R^n$.
\end{remark}

In computer science, a functional usually refers to a higher-order function, i.e. a function that takes one or more functions as arguments or returns a function as a result. 

In mathematics, a \emph{functional} is a map whose codomain is the field of scalars, \[ F:X\rightarrow\mathbb{K}, \] where $X$ is typically a space of functions. The input elements of $X$ are often themselves functions, which explains the similarity between the mathematical and computer-science uses of the term. 

\begin{definition} 
Let $V$ and $W$ be vector spaces over the same field $\mathbb{K}$. A map \[ L:V\rightarrow W \] is said to be \emph{linear} if for all $f,g\in V$ and all scalars $\alpha,\beta\in\mathbb{K}$, \[ L(\alpha f+\beta g) = \alpha L(f)+\beta L(g). \] \end{definition} 

A functional need not be linear. 
In linear algebra, a linear form (also known as a linear functional, a one-form, or a covector) is a linear map from a vector space to its field of scalars. 
In functional analysis, it refers to a mapping from a space $X$ into the field of real or complex numbers. A linear functional on $C(I_n)$ is a scalar-valued linear map defined on $C(I_n)$, in concordance with the computer science interpretation of functional.

\begin{definition} 
A \emph{linear functional} on $C(I_n)$ is a linear map \[ L:C(I_n)\rightarrow\mathbb{K}. \] That is, \[ L(\alpha f+\beta g) = \alpha L(f)+\beta L(g) \] for all $f,g\in C(I_n)$ and all $\alpha,\beta\in\mathbb{K}$. 
\end{definition} 

\subsection*{Measures}

A measure is a generalization and formalization of geometrical measures (length, area, volume) and other common notions, such as magnitude, mass, and probability of events. In mathematical analysis and in probability theory, a $\sigma$-algebra is part of the formalism for defining sets that can be measured (also $\sigma$-field, where the $\sigma$ comes from the German ``Summe,'' meaning ``sum''):

\begin{definition}
A $\sigma$-algebra  on a set $X$ is a nonempty collection $\Sigma$ of subsets of $X$ closed under complement, countable unions, and countable intersections:\footnote{This definition is mathematically correct but slightly redundant: closure under countable intersections already follows from closure under complements and countable unions via De Morgan's laws.} 
$$ \Sigma \subseteq \mathcal{P} (X) $$
such that
\begin{enumerate}
    \item $X \in \Sigma$,
    \item $E \in \Sigma \Rightarrow E^c \in \Sigma$,
    \item If $E_1..E_n \in \Sigma$, then $\cup_k E_k \in \Sigma$.
\end{enumerate}
Countable intersections then follow automatically.
\end{definition}

\begin{definition}
The ordered pair $(X,\Sigma)$ is called a measurable space.
\end{definition}

\begin{definition}
Any element of the chosen $\sigma$-algebra $\Sigma$ is a measurable set.
\end{definition}

\begin{definition}
Let $X$ be a set and $\Sigma$ a $\sigma$-algebra over $X$, defining subsets of $X$
that are measurable. A measure $\mu: \Sigma \to [0, +\infty)$ is a set function from the $\sigma$-algebra that satisfies the following conditions:

\begin{itemize}

\item 
$\mu(\emptyset)=0$.

\item
Non-negativity: For all $E\in \Sigma$, $\mu(E) \ge 0$.

\item
Countable additivity (or $\sigma$-additivity):  For all countable collections 
$\{E_k\}_{k=1}^\infty$ of pairwise disjoint sets in $\Sigma$,
$
\mu \left( \cup_{k=1}^{\infty} E_k\right) = \sum_{k=1}^{\infty} \mu(E_k)
$
\end{itemize}

\end{definition}

\begin{definition}[Borel set]
Let $X$ be a topological space. The Borel $\sigma$-algebra of $X$, denoted by $\mathcal{B}(X)$, is the smallest $\sigma$-algebra containing all open subsets of $X$. The elements of $\mathcal{B}(X)$ are called Borel sets. 
\end{definition}

\begin{definition}[Borel measure]
Let $X$ be a topological space and let $\mathcal{B}(X)$ denote its Borel $\sigma$-algebra. A Borel measure on $X$ is a measure $\mu : \mathcal{B}(X) \to [0,\infty]$ defined on the Borel $\sigma$-algebra. 
\end{definition}

Measures on topological spaces are usually defined on the Borel $\sigma$-algebra because open and closed sets have geometric meaning, the Borel $\sigma$-algebra is rich enough for most applications (e.g. analysis, probability, and statistics), and continuous functions interact naturally with Borel sets (the preimage of a Borel set under a continuous function is again a Borel set).

\begin{definition}[Signed measure] 
Let $(X,\Sigma)$ be a measurable space. A signed measure is a countably additive set function $\mu:\Sigma\to[-\infty,\infty]$ such that $\mu(\emptyset)=0$ and at most one of the values $+\infty$ and $-\infty$ is attained. Countable additivity means that for every sequence $\{E_k\}_{k=1}^{\infty}$ of pairwise disjoint measurable sets, \[ \mu\!\left(\bigcup_{k=1}^{\infty}E_k\right) = \sum_{k=1}^{\infty}\mu(E_k). \] Unlike ordinary measures, a signed measure may take both positive and negative values. \end{definition}

In mathematics, specifically general topology, compactness is a property that seeks to generalize the notion of a closed and bounded subset of Euclidean space. Roughly speaking, compact sets are those that can be covered by arbitrary collections of open sets, while still admitting a finite subcover.\footnote{Intuitively, a compact space has no punctures or missing endpoints. But this description is mathematically dangerous. For instance, the circle $S^1$ has a ``hole'' yet it is compact. 
Compactness actually means that, given a topological space $X$. a subset $K\subseteq X$ is compact if every open cover of $K$ admits a finite subcover. Whenever $ K \subseteq \bigcup_{\alpha\in A} U_\alpha$, where each $U_\alpha$ is open, there exist finitely many indices $\alpha_1,\ldots,\alpha_n \in A$ such that $ K \subseteq U_{\alpha_1}\cup\cdots\cup U_{\alpha_n}$.
In Euclidean spaces, a subset of $\mathbb{R}^n$ is compact iff it is closed and bounded (Heine--Borel theorem).\cite{heine1872}\cite{borel1895}}

\begin{definition}[Regular Borel measure]
Let $X$ be a topological space. A Borel measure $\mu$ on $X$ is regular if every Borel set $E\in\mathcal B(X)$ can be approximated from inside by compact sets
and from outside by open sets:
\[
\mu(E)
=
\sup\{\mu(K): K\subseteq E,\; K \text{ compact}\},
\]
and
\[
\mu(E)
=
\inf\{\mu(U): E\subseteq U,\; U \text{ open}\}.
\]
The first property is called inner regularity and the second property is called outer regularity.
\end{definition}

\begin{definition} \label{def:MIn}
The space of finite signed regular Borel measures on the compact space $I_n = [0,1]^n$ is denoted by $M(I_n)$.

That is, $M(I_n) =
\bigl\{ \mu: \mathcal B(I_n)\to\mathbb R\bigr\}$, where  $\mu$ is a finite signed regular Borel measure and finiteness means $|\mu|(I_n)<\infty$, with $|\mu|$ denoting the total variation measure of $\mu$.

Since the measure is finite, i.e. $\mu(I_n)<\infty$, local finiteness is automatic on compact subsets: $\mu(K)<\infty$ for every compact $K$, since $\mu(I_n)<\infty$ and every compact set satisfies $K \subseteq I_n$, so $\mu(K) \leq \mu(I_n)<\infty$.
\end{definition}

\begin{definition}[Measurable function]
A measurable function $ f: X \rightarrow \mathbb{R}$ satisfies
$f^{-1}(B) \in \Sigma$ for every Borel set $B \subseteq \mathbb{R}$.
\end{definition}
A function $f:X\to\mathbb{R}$ is measurable if it is compatible with the underlying notion of measure, allowing quantities such as $\int_X f\,d\mu$ to be defined whenever $f$ is integrable.

\begin{definition}
L-infinity, $L^\infty(X) = L^\infty(X,\Sigma,\mu)$, is the vector space of essentially bounded measurable functions with the essential supremum norm.
$L^\infty(X)$ depends on $\Sigma$ and $\mu$, yet we omit them for notational convenience.

\begin{itemize}
\item 
A function $f \in L^\infty(X)$ is essentially bounded when there is a constant $c<\infty$ such that $|f(x)| \le c$ except on a set of measure zero.
\item
The essential supremum norm\footnote{$\|f\|_{\infty,\mu}= \inf \{ c > 0: \mu \{ x \in \mathbb R^n: |f(x)| > c \} = 0 \}$} 
is defined as $\|f\|_{\infty} = \inf \{ c \ge 0: |f(x)| \le c \text{ for almost every } x \in X \}$.
\end{itemize}
\end{definition}

$L^\infty(X)$ is a Banach space under the norm $\| \cdot\|_{\infty}$. The vector space $L^\infty(X)$ consists of equivalence classes of essentially bounded measurable functions, where two functions are identified if they are equal almost everywhere (functions that differ only on a set of measure zero regarded as equivalent).

\begin{definition}
A simple function is a measurable function that takes only finitely many values.    
\end{definition}

\begin{theorem} \label{thm:simple-dense}
Simple functions are dense in $L^\infty(X)$ with respect to the essential-supremum norm for any measure space $(X,\Sigma,\mu)$.
\end{theorem}




\begin{proof}
To prove that simple functions are dense in $L^\infty(X)$ with respect to the essential-supremum norm, we need to show that for every essentially bounded measurable function $f \in L^\infty(X)$ and every $\varepsilon>0$, there exists a simple function 
that $\| f-s \|_{\infty} <\varepsilon$.

Let $[f] \in L^\infty(X)$, and choose a measurable representative $f$ of the equivalence class. Since $f$ is essentially bounded, there exists $c>0$ such that $|f(x)| \leq c$ almost everywhere (modifying on a null set, i.e. a set of measure zero, if needed) only because $L^\infty$ consists of equivalence classes modulo almost-everywhere equality.
Replacing $f$ by an equivalent representative if necessary, we may assume that $|f(x)|\le c$ for all $x\in X$. 

Given $\varepsilon > 0$, we can partition the interval $[ -c, c ]$ into finitely many subintervals of length less than $\varepsilon > 0$. For example, 
$I_k = \left[ a_k, a_{k+1} \right)$, with $k = 1..N-1$ and $I_N=\left[ a_N, a_{N+1} \right]$, where $a_k = -c + (k-1)\delta$, and $\delta < \varepsilon$.

Now, we can define the simple function $s(x) = a_k \:\: \text{whenever } f(x) \in I_k$ (i.e. a representative value $a_k\in I_k$ for each interval $I_k$).
This function takes only finitely many values $\{a_1 .. a_N\}$, and each level set $\{x \in X: f(x)  I_k\}$ is measurable, so $s$ is a simple measurable function.

For every $x \in X$, we have: $|f(x) - s(x)| < \delta < \varepsilon$. Hence, $|f - s| < \varepsilon$ for almost every $x\in X$. 
By the definition of the essential supremum norm, $\|f-s\|_{L^\infty} = \operatorname*{ess\,sup}_{x\in X}|f(x)-s(x)| <\varepsilon$. 
Therefore, $\|f - s\|_{L^\infty} < \varepsilon$. Since $\varepsilon > 0$ was arbitrary, simple functions are dense in $L^\infty(X)$  with respect to the essential-supremum norm. QED.
\end{proof}

 
The Riesz Representation Theorem is a fundamental result in functional analysis that characterizes the dual of certain function spaces. There are several versions depending on the context. In measure theory, the Riesz–Markov–Kakutani representation theorem relates linear functionals on spaces of continuous functions on a locally compact space to measures in measure theory.

\begin{definition} [Compact support]
Let $X$ be a topological space and let $f:X\to\mathbb{R}$. The support of $f$ is the closure of the set $\{x\in X : f(x)\neq 0\}$. The function $f$ is said to have compact support if its support is compact. 
\end{definition}

\begin{definition} [Positive linear functional]
Let $X$ be a set and let $V$ be a vector space of real-valued functions. 
A linear functional $L:V\to\mathbb R$ is positive if, for every $f\in V$, $f(x)\ge 0 \quad \forall x\in X$ implies $L(f)\ge 0$.
\end{definition}

\begin{definition} [Locally compact space]
A topological space $X$ is locally compact if every point $x\in X$ has an open neighborhood whose closure is compact.
Intuitively, every point has a small neighborhood that behaves like a compact set.
\end{definition}

\begin{definition} [Hausdorff space]
A topological space $X$ is Hausdorff if for every pair of distinct points $x,y \in X$, there exist disjoint open sets $U$ and $V$ such that $x\in U, y\in V$. 
\end{definition}

In a Hausdorff space, points can be separated by open neighborhoods. This ensures that limits of sequences (or nets) are unique when they exist. Most familiar spaces, such as the real line $\mathbb{R}$ or metric spaces, are Hausdorff spaces.

\begin{definition} [Radon measure]
Let $X$ be a locally compact Hausdorff space. A Radon measure on $X$ is a Borel measure that is 
\begin{itemize} 
\item locally finite, i.e. every point $x\in X$ has an open neighborhood $U$ such that $\mu(U)<\infty$ (on locally compact Hausdorff spaces, this is equivalent to finite measure on compact sets, i.e. $\mu(K)<\infty$ for every compact set $K$); 
\item inner regular, i.e. for every Borel set $E\in\mathcal B(X)$: $$\mu(E) = \sup\{\mu(K):K\subseteq E,\;K\text{ compact}\};$$
\item outer regular, i.e. for every Borel set $E\in\mathcal B(X)$: $$\mu(E) = \inf\{\mu(U):E\subseteq U,\;U\text{ open}\}.$$ 
\end{itemize} 
\end{definition}

Informally, Radon measures can be viewed as well-behaved Borel measures. Their regularity properties allow the measure of complicated Borel sets to be approximated using compact subsets from within and open supersets from outside.

On a locally compact Hausdorff space, every finite regular Borel measure is a Radon measure. 


\begin{theorem}[Riesz-Markov-Kakutani Representation Theorem (measure theory)] 
\label{thm:riesz}
Let \( K \) be a locally compact Hausdorff space, and let \( C_c(K) \) be the space of continuous functions with compact support. Then every positive linear functional \( L: C_c(K) \to \mathbb{R} \) is represented by a unique Radon measure \( \mu \) such that:
\[
L(f) = \int_K f \, d\mu \quad \forall f \in C_c(K).
\]
\end{theorem}

The Riesz-Markov-Kakutani theorem\footnote{Frigyes Riesz introduced it for continuous functions on the unit interval in 1909 \cite{riesz1909}, Andrey Markov extended the result to some non-compact spaces in 1938 \cite{markov1938}, and Shizuo Kakutani extended the result to compact Hausdorff spaces in 1941 \cite{kakutani1941}.} establishes a one-to-one correspondence between positive linear functionals on continuous functions and Radon measures. In other words, every positive linear functional can be represented as integration against a unique measure.

Since $I_n=[0,1]^n$ is compact, every continuous function on $I_n$ has compact support and, therefore, $C_c(I_n)=C(I_n)$. For the compact space $I_n$, the Riesz-Markov-Kakutani Representation Theorem states that every positive linear functional $L:C(I_n)\to\mathbb{R}$ can be represented uniquely as \[ L(f) = \int_{I_n} f\,d\mu, \] for some finite regular Borel measure $\mu\in M(I_n)$.

Since $I_n=[0,1]^n$ is a compact metric space, every finite Borel measure on $I_n$ is automatically regular. Regularity guarantees that the measure of any Borel set can be approximated arbitrarily well using compact subsets from within and open supersets from outside. Consequently, every finite Borel measure on $I_n$ is a Radon measure.




\subsection*{Swapping limits and integrals}

The Lebesgue Bounded Convergence Theorem \cite{lebesgue1904} gives mathematicians a safe way to swap limits and integrals, a powerful tool when analyzing functions that change over time.
The Lebesgue Bounded Convergence Theorem is a special case of the Dominated Convergence Theorem, where the dominating function is the constant function $g(x)=c$.

\begin{itemize}

\item
Let $E\subset \mathbb{R}^d$ be a measurable set and let $\mu$ denote Lebesgue measure (length in one dimension, area in two dimensions, volume in three dimensions, and their higher-dimensional analogues).

\item
A statement is said to hold almost everywhere (a.e.) on $E$ if it fails only on a subset of $E$ having measure zero.

\item
For any measurable function $g$ with finite integral $\int_E |g|\,d\mu<\infty$, its $L^1$-norm is $$ \|g\|_1=\int_E |g(x)|\,d\mu(x). $$

\end{itemize}

\begin{theorem}[Lebesgue Bounded Convergence Theorem] \label{thm:lebesgue}
Let $(f_n)_{n \in \mathbb{N}}$ be a sequence of measurable functions defined on a measurable set $E \subset \mathbb{R}^d$ with finite measure $\mu(E) < \infty$. 
Suppose that $f_n(x) \to f(x)$ pointwise almost everywhere on $E$ and also that there exists a constant $c > 0$ such that $|f_n(x)| \leq c$ for all $n$ and almost every $x \in E$. Then the limit function $f$ is measurable and integrable,
$$
\lim_{n \to \infty} \int_E f_n(x) \, d\mu(x) = \int_E f(x) \, d\mu(x),
$$
and
$$
\lim_{n \to \infty} \|f_n - f\|_1 = 0.
$$
\end{theorem}

Since $|f_n(x)|\le c$ almost everywhere and $f_n(x)\to f(x)$ almost everywhere, it follows that $|f(x)|\le c$ almost everywhere. Because $\mu(E)<\infty$, $$ \int_E |f(x)|\,d\mu(x) \le c\,\mu(E) <\infty, $$ so $f$ is integrable.

If you have a sequence of functions that get closer and closer to a final function (pointwise convergence), they are all uniformly bounded (none of them goes off to infinity), and you are working over a space with finite size (i.e. a domain whose total measure is finite, such as a bounded interval, a bounded region in the plane, or any measurable set with finite volume), then the integrals (which measure total size or area under the curve) of those functions also get closer to the integral of the final function.
The theorem provides conditions under which a limiting operation may be moved inside an integral. If the functions $f_n$ converge pointwise almost everywhere to a limiting function $f$, none of the functions becomes arbitrarily large (they are uniformly bounded by the same constant $c$), and the domain $E$ has finite measure (finite length, area, or volume), then the total accumulated quantity represented by the integral behaves continuously under the limit. Moreover, the entire error between $f_n$ and $f$, measured by the total absolute deviation $\int_E |f_n-f|\,d\mu$ (i.e. measured by the $L^1$ distance), tends to zero. Not only do the function values converge almost everywhere, but the cumulative discrepancy over the entire domain also vanishes.

For engineers, this theorem justifies exchanging a limit and an integral when approximating signals, density fields, temperature distributions, or numerical solutions by uniformly bounded approximations over a finite domain. The boundedness condition prevents small regions from developing arbitrarily large spikes that could otherwise invalidate the interchange of limit and integration.


\subsection*{Separating points by hyperplanes}

The Hahn-Banach Extension Theorem\footnote{The extension theorem for linear functionals is named for the mathematicians Hans Hahn and Stefan Banach, who proved it independently in the late 1920s.\cite{hahn1927}\cite{banach1929}\cite{banach1932} Separation results are consequences derived from it.} is one of the foundational results in functional analysis. One of its most important geometric consequences is the Hahn-Banach Separation Theorem, often called the point-set separation theorem. which provides a powerful separation principle in convex analysis.

\begin{definition}[Hyperplane]
    A hyperplane in a real vector space $X$ is a set of the form
    $$ H=\{x\in X: L(x)=\alpha\} $$
    where $L:X\to\mathbb R$ is a nonzero linear functional and $\alpha\in\mathbb R$.
\end{definition}

The sets $H^-=\{x\in X:L(x)\le \alpha\}$ and $H^+=\{x\in X:L(x)\ge \alpha\}$ are called the associated half-spaces.

\begin{definition}[Convexity]
A set $Y\subseteq X$ is convex if for every $y_1,y_2\in Y$ and every $t\in[0,1]$,
$$ty_1+(1-t)y_2\in Y$$.
\end{definition}

Geometrically, every line segment joining two points of $Y$ remains entirely inside $Y$.

\begin{definition}[Closed set] 
A subset $Y\subseteq X$ is closed if it contains all of its limit points. Equivalently, whenever a sequence $(y_n)\subseteq Y$ converges to a point $y\in X$, the limit $y$ also belongs to $Y$. 
\end{definition}

\begin{definition}[Normed vector space]
    A normed vector space is a vector space $X$ equipped with a function \[ \|\cdot\|:X\to[0,\infty) \] called a norm, satisfying \[ \|x\|=0 \iff x=0, \] \[ \|\lambda x\|=|\lambda|\|x\|, \] and \[ \|x+y\|\le \|x\|+\|y\| \] for all $x,y\in X$ and all scalars $\lambda$. The norm measures the size or length of vectors in $X$.
\end{definition}

\begin{definition}[Continuity of linear functionals]
A linear functional is a linear map $L:X\to \mathbb R$. It is continuous if it is continuous with respect to the norm topology of $X$. Intuitively, small changes in $x$ produce small changes in $L(x)$. In normed spaces, continuity is equivalent to boundedness for some constant $C>0$:
$$ |L(x)| \le C\|x\| $$
for some constant $C>0$ and all $x\in X$.
\end{definition}

\begin{theorem}[Hahn--Banach Separation Theorem] \label{thm:hahn-banach-separation} Let \(X\) be a normed vector space and let \(M\subset X\) be a linear subspace. 
If \(M\) is not dense in \(X\), then there exists a nonzero continuous linear functional $L:X\to\mathbb{R}$  such that 
\[ L(m)=0 \qquad \forall m\in M. \] 
Equivalently, 
\[ M^\perp = \{L\in X^*:L(m)=0\ \forall m\in M\} \] 
contains a nonzero element. 
\end{theorem}

The notation $\perp$ is analogous to orthogonality in Euclidean space: $M^\perp$ denotes the annihilator of $M$ in the dual space $X^*$ (the continuous dual space of the normed vector space $X$, i.e., the space of all continuous linear functionals $L:X\to\mathbb R$).

\begin{corollary} \label{cor:hahn-banach-annihilator}
Let \(X\) be a normed vector space and \(M\subset X\) a linear subspace. If the only continuous linear functional vanishing on \(M\) is the zero functional, then $\overline{M}=X$. In other words, \(M\) is dense in \(X\). 
\end{corollary}

\begin{remark}
Geometrically, if $M$ fails to be dense, then its closure $\overline M$ is a proper closed subspace of $X$. By the Hahn-Banach Separation Theorem, any point $x_0\notin\overline M$ can be separated from $\overline M$ by a continuous hyperplane. The separating hyperplane is given by a nonzero continuous linear functional $L$ satisfying $L(x)=0, x\in\overline M$, and hence $L(x)=0$ for every $x\in M$. This yields the fundamental dual characterization of density:
$$
\overline M=X \quad \Longleftrightarrow \quad M^\perp=\{0\}
$$
A linear subspace is dense if and only if the only continuous linear functional that vanishes on it is the zero functional. 
\end{remark}

\begin{theorem}[Hahn--Banach Separation Theorem (Geometric Form)] \label{thm:hahn-banach}
Let $X$ be a real normed vector space, $Y\subseteq X$  be a non-empty closed convex set, and $x_0 \in X \setminus Y$ (i.e., a point not in $Y$). Then there exists a nonzero continuous linear functional $L: X \to \mathbb{R}$, and a real number $\alpha \in \mathbb{R}$ such that:
$$
L(x_0) > \alpha \geq L(y) \quad \forall y \in Y.
$$
\end{theorem}

Equivalently, the hyperplane
$$H=\{x\in X:L(x)=\alpha\}$$
strictly separates\footnote{Strict separation has a technical meaning: $L(x_0)>\alpha\ge L(y) \quad\forall y\in Y$. Thus $x_0$ lies strictly on one side of the hyperplane, while $Y$ lies entirely on the other side.} the point $x_0$ from the convex set $Y$. 
The theorem, therefore, guarantees that convex sets can be separated from external points by hyperplanes.


\begin{proof}
Embed the convex set $Y$ into the convex and closed set $C=Y\times\{1\}\subset X\times\mathbb R$. The origin $(0,0) \in X\times\mathbb R$, yet $(0,0) \notin C$.

The convex cone generated by $C$ is $K =\{\, t(y,1): y\in Y,\ t\ge 0\,\}$. For $x_0 \notin Y$, $(x_0,1) \notin \overline K$. 
If $(x_0,1)$ belonged to $\overline K$, there would exist $t_n\ge 0$ and $y_0 \in Y$ such that $(t_n y_n,t_n) \to (x_0,1)$. Its second coordinate implies that $t_n \to 1$, so $y_n \to x_0$ (a contradiction, $x_0$ would belong to Y)

Let $E=X\times\mathbb R$ and consider the linear span $M=\operatorname{span}(\overline K)\subset E$. since $(x_0,1)\notin \overline K$, we have $(x_0,1)\notin \overline M$. Consequently $M$ is a proper closed subspace of the space $\overline M+\mathbb R(x_0,1)$. Hence $M$ is not dense there.. 

By the corollary, there exists a nonzero continuous linear functional $F:E\to\mathbb R$ such that $F(u)=0$, $u\in M$. Since $K\subset M$, $F(t(y,1))=0$ ($y\in Y$, $t \ge 0$). Therefore $F(y,1)=0$, $\forall y\in Y$.

Every continuous linear functional on $X\times\mathbb R$ has the form $F(x,t)=L(x)-\alpha t$ for some $L\in X^*$ and $alpha\in\mathbb R$.Since $F(y,1)=0$ for all $y\in Y$, $L(y)=\alpha$ $\forall y\in Y$. The functional is nonzero, so $L$ is not identically zero.

Furthermore, $F(x_0,1) \neq 0$, because $(x_0,1)\notin \overline M$. Replacing $F$ by $-F$ if necessary, we may assume $F(x_0,1)>0$ and, therefore, $L(x_0)-\alpha>0$, i.e. $L(x_0)>\alpha$.

Combining this with $L(y)=\alpha$ for all $y\in Y$, we obtain the desired result:
$$L(x_0)>\alpha\ge L(y) \qquad \forall y\in Y$$

Thus the theorem follows from the annihilator corollary by lifting the problem to $X\times\mathbb R$ and applying the non-dense-subspace result there.
\end{proof}

The geometric Hahn-Banach separation theorem asserts that every point lying outside a closed convex set can be distinguished from that set by a continuous linear measurement $L$. The values of $L$ on the set $Y$ are all bounded above by $\alpha$, whereas the value of $L$ at $x_0$ is strictly larger than $\alpha$. Consequently, $Y$ and $x_0$ lie on opposite sides of the hyperplane $L(x)=\alpha$.

\begin{remark}
The condition $L\neq0$ is essention and cannot be omitted. Otherwise, the statement becomes vacuous because the zero functional cannot provide a separating hyperplane. 
\end{remark}

\begin{remark}
Likewise, if $Y$ is not closed, strict separation can fail. The closedness assumption ensures a positive geometric gap between $x_0$ and the set; i.e. it guarantees that the distance from $x_0$ to $Y$ is strictly positive: $ d(x_0,Y)=\inf_{y\in Y}\|x_0-y\|>0$ when $Y$ is closed and $x_0\notin Y$ in metric spaces.
\end{remark}


\subsection*{Fourier analysis}


\begin{definition}[Fourier Transform] \label{def:fourier-transform} 
Let \(f\in L^1(\mathbb{R}^n)\). The (unnormalized) Fourier transform of \(f\) is defined by 
\[ \widehat{f}(\omega) = \mathcal{F}[f](\omega) = \int_{\mathbb{R}^n} f(x)e^{-i\,x\cdot\omega}\,dx, \quad \omega\in\mathbb{R}^n, \] 
where $x\cdot\omega = \sum_{k=1}^{n} x_k\omega_k$ denotes the Euclidean inner product. \end{definition}

\begin{definition}[Inverse Fourier Transform] \label{def:inverse-fourier-transform} For a sufficiently regular function \(F\), the inverse Fourier transform is given by 
\[ f(x) = \mathcal{F}^{-1}[F](x) = \frac{1}{(2\pi)^n} \int_{\mathbb{R}^n} F(\omega)e^{i\,x\cdot\omega}\,d\omega, \quad x\in\mathbb{R}^n. \] 
\end{definition}

\begin{remark} Under this convention, the inverse transform contains the factor \((2\pi)^{-n}\), which is unnecessary in the normalized Fourier transform $\int_{\mathbb{R}^n} f(x)e^{-2\pi i x\cdot \omega}\,dx$.
\end{remark}



These operators extend to \(L^2(\mathbb{R}^n)\), tempered distributions, and other function spaces by density and continuity. They also satisfy the well-known property:
$$\mathcal{F}^{-1}[\mathcal{F} f](x) = f(x) $$

\begin{definition}[Support] 
Let \(f:\mathbb{R}^n\to\mathbb{R}\) be a function. The \emph{support} of \(f\) is the closure of the set where \(f\) is nonzero: 
\[ \operatorname{supp}(f) = \overline{\{x\in\mathbb{R}^n : f(x)\neq 0\}}. \] 
A function is said to be \emph{compactly supported} if its support is a compact subset of \(\mathbb{R}^n\). 
\end{definition}

\begin{definition}[Band-Limited Function] 
A function \(f:\mathbb{R}^n\to\mathbb{C}\) is called \emph{band-limited} if its Fourier transform vanishes outside some bounded set. 
Equivalently, there exists \(R>0\) such that \[ \operatorname{supp}(\widehat{f}) \subseteq B_R(0) = \{\xi\in\mathbb{R}^n:\|\xi\|\le R\}. \] 
\end{definition}

\begin{definition}[Convolution] 
Let \(f,g:\mathbb{R}^n\to\mathbb{R}\) be functions such that the integral below is well defined. Their \emph{convolution} is the function 
\[ (f*g)(x) = \int_{\mathbb{R}^n} f(x-y)\,g(y)\,dy. \] 
Convolution averages the values of one function using the other as a kernel. \end{definition}

\begin{definition}[Schwartz Function] 
A function $\phi:\mathbb{R}^n\to\mathbb{C}$ is called a \emph{Schwartz function} if it is infinitely differentiable and all of its derivatives decay at infinity faster than any polynomial. Precisely, for every pair of multi-indices \(\alpha,\beta\), \[ \sup_{x\in\mathbb{R}^n} \bigl|x^\alpha D^\beta \phi(x)\bigr| <\infty. \] The set of all Schwartz functions is denoted by \(\mathcal{S}(\mathbb{R}^n)\). 
\end{definition}

\begin{remark} Schwartz functions are smooth functions whose values and all derivatives tend to zero at infinity faster than any inverse power of \(\|x\|\). Typical examples include the Gaussian $\phi(x)=e^{-\|x\|^2}$. Schwartz functions are particularly important in Fourier analysis because the Fourier transform maps \(\mathcal{S}(\mathbb{R}^n)\) onto itself. \end{remark}

\begin{definition}[Entire Function]
A function
\[
F:\mathbb{C}^n\to\mathbb{C}
\]
is called \emph{entire} if it is holomorphic in every complex variable
on all of \(\mathbb{C}^n\). In other words, \(F\) is complex-differentiable with respect to every complex variable throughout all of \(\mathbb{C}^n\). Equivalently, it admits a convergent power-series expansion about every point of \(\mathbb{C}^n\); i.e. it can be represented locally by a convergent power series in the complex variables \(z_1,\ldots,z_n\) at every point of \(\mathbb{C}^n\).
\end{definition}

\begin{definition}[Support Function]
Let \(K\subset\mathbb{R}^n\) be a compact convex set. The
\emph{support function} of \(K\) is
\[
h_K(y)
=
\sup_{x\in K} x\cdot y,
\qquad
y\in\mathbb{R}^n.
\]
It measures the maximal extent of \(K\) in the direction \(y\).
\end{definition}


The Paley–Wiener theorem\footnote{Theorem named after Raymond Paley (1907–1933) and Norbert Wiener (1894–1964) who, in 1934, introduced various versions of the theorem \cite{paley_wiener_1934}.} connects support properties of a function’s Fourier transform  with analytic continuation and growth conditions of the function itself: compact support in the time domain corresponds to a entire function of exponential type in the frequency domain.

\begin{theorem}[Paley–Wiener theorem (one-dimensional)] \label{thm:paley-wiener-1d}
A function \(F:\mathbb{C}\to\mathbb{C}\) is the Fourier transform of an \(L^2\)-function $f$ supported in $[-R,R] \subset \mathbb{R}$ if and only if:
\begin{itemize}
    \item $F$ is entire (holomorphic on all of $\mathbb{C}$), and
    \item Growth bound: There exists $C>0$ such that for all $z\in\mathbb{C}$,
     \[
     |F(z)| \leq C (1+\|z\|)^N \, e^{R|\Im z|}, %
     \]
     for some integer \(N\).  
\end{itemize}


\end{theorem}

\begin{theorem}[Paley–Wiener theorem (multidimensional)] \label{thm:paley-wiener}
  Let \(K\subset \mathbb{R}^n\) be a compact convex set and let \(f\in L^2(\mathbb{R}^n)\) satisfy $\operatorname{supp}(f)\subseteq K$. 
  Then its Fourier transform $F=\widehat{f}$ extends uniquely to an entire function on \(\mathbb{C}^n\) satisfying the growth estimate 
  \[ |F(z)| \le C(1+\|z\|)^N e^{h_K(\Im z)} \] 
  for suitable constants \(C,N>0\). 
  
  Conversely, every entire function satisfying such a growth condition and the appropriate \(L^2\)-integrability on \(\mathbb{R}^n\) arises as the Fourier transform of an \(L^2\)-function supported in \(K\). 
\end{theorem}

The key intuition is that entire functions are exceptionally smooth. They can be extended from real variables $x\in\mathbb R^n$ to complex variables $z\in\mathbb C^n$ without encountering singularities. 

\begin{corollary}[Band-Limited Functions are Entire] \label{cor:bandlimited-entire} 
If \(f\in L^2(\mathbb{R}^n)\) is band-limited, i.e., $\operatorname{supp}(\widehat{f}) \subseteq K$ for some compact set \(K\subset\mathbb{R}^n\), then \(f\) extends to an entire function on \(\mathbb{C}^n\). Moreover, its growth is controlled by the support function \(h_K\): \[ |f(z)| \le C(1+\|z\|)^N e^{h_K(\Im z)}. \] \end{corollary}


\begin{lemma}[Uniform approximation on compact sets by band-limited functions]
\label{lemma:approximation-band-limited}

For every compact \(K\subset\mathbb R^n\), every \(f\in C(K)\), and every \(\delta>0\), there exists a smooth band-limited function \(F\in C^\infty(\mathbb R^n)\) such that 
\[ \operatorname{supp}(\widehat F)\subset B_R(0) \] 
for some \(R>0\), and 
\[ \sup_{x\in K} |F(x)-f(x)| < \delta. \]
\end{lemma}

\section{Formal Proofs}


\subsection*{Cybenko's Universal Approximation Theorem}


Cybenko's theorem is stated on the compact unit cube $I_n = [0,1]^n \subset \mathbb R^n$ because the proof uses $C(I_n)$ as a Banach space (definition \ref{def:CIn}) and the Riesz Representation Theorem (theorem \ref{thm:riesz}). 
Given the space of finite signed measures on $I_n$, $M(I_n)$ (definition \ref{def:MIn}), the Riesz Representation Theorem identifies $C(I_n)^* \simeq M(I_n)$.

\begin{definition}
\label{def:function-discriminatory}
A function $\sigma: \mathbb{R} \to \mathbb{R}$ is discriminatory if, whenever a finite signed measure $\mu \in M(I_n)$ satisfies
$$ \int_{I_n} \sigma(w^\top x+ b) d\mu(x) = 0$$
for all $w \in \mathbb{R}^n$ and $b \in \mathbb{R}$, then $\mu=0$.
\end{definition}

\begin{definition}
We say that $\sigma: \mathbb{R} \to \mathbb{R}$ is sigmoidal if $\lim_{z \to +\infty} \sigma(z) =1$ and $\lim_{z \to -\infty} \sigma(z) =0$
\end{definition}

\begin{definition}
    The continuous function space $C(I_n) = \{f:I_n\to\mathbb R \;:\; f \text{ is continuous}\}$ is equipped with the uniform norm $\|f\|_\infty = \sup_{x\in I_n}|f(x)|$. A subset $S\subset C(I_n)$ is dense if every continuous function can be approximated arbitrarily well in this norm:
    $$\forall f\in C(I_n),\; \forall \varepsilon>0,\; \exists g\in S \text{ such that } \|f-g\|_\infty<\varepsilon$$
\end{definition}

\repthm{thm:cybenko}

\begin{remark}
The class $G(x) = \sum_{j=1}^{h} \alpha_j \sigma(w_j^\top x+b_j)$ is exactly the set of functions computed by a neural network with $n$ inputs, one hidden layer containing $h$ neurons, activation function $\sigma$, and a linear output layer.
\end{remark}

\begin{proof}
Let $S \subset C(I_n)$ be the set of functions of the form $G(x)$\footnote{Hidden assumption: Strictly speaking, $S = \left\{ \sum_{j=1}^{h} \alpha_j \sigma(w_j^\top x+b_j) : h<\infty \right\}$. If $h$ were fixed, $S$ would not even be a vector space.}. Clearly, $S$ is a linear subspace of $C(I_n)$. We prove that the closure of $S$ is $C(I_n)$ by contradiction, i.e. $\overline S=C(I_n)$.

Assume that the closure of $S$ is not all of $C(I_n)$. Then, the closure of $S$, say $R$, is a closed proper subspace of $C(I_n)$. Since $R$ is a proper closed subspace of the Banach space $C(I_n)$, the Hahn-Banach separation theorem (theorem \ref{thm:hahn-banach}) implies the existence of a nonzero continuous linear functional $L$ vanishing on $R$: $L \ne 0$ and $L(R) = L(S) = 0$.

By the Riesz Representation Theorem (theorem \ref{thm:riesz}), this bounded linear functional, $L$, is of the form
$$ L(h) = \int_{I_n} h(x) d\mu(x) \ne 0$$
for some $\mu \in M(I_n)$, for all $h \in C(I_n)$. In particular, since $\sigma(w^\top x+ b)$ is in $R$ for all $w$ and $b$,\footnote{$\sigma(w^\top x+b) = 1\cdot \sigma(w^\top x+b)$ is an element of $S$, and therefore belongs to $R=\overline S$.}  we must have that
$$ \int_{I_n} \sigma(w^\top x+ b) d\mu(x) = 0$$
for all $w$ and $b$.

However, we assumed that $\sigma$ was discriminatory so that this condition implies that $\mu=0$ contradicting our assumption. Hence, the subspace $S$ must be dense in $C(I_n)$.
\end{proof}

This demonstrates that sums of the form $G(x)$ are dense in $C(I_n)$ provided that $\sigma$ is continuous and discriminatory. We can connect this result with the sigmoid activation functions used by neural networks:

\begin{lemma} \label{lemma:sigmoidal-discriminatory}
Any bounded, measurable sigmoidal function, $\sigma$, is discriminatory. In
particular, any continuous sigmoidal function is discriminatory.
\end{lemma}

\begin{proof}
Based on \cite{cybenko1989}[pp.307-308], with correction \cite{cybenko1992}
$$
\sigma(\lambda(w^\top x+ b)+\varphi)
\begin{cases}
\to 1 & \text{for } w^\top x+ b>0 \text{ as } \lambda \to \infty \\
\to 0 & \text{for } w^\top x+ b<0 \text{ as } \lambda \to \infty \\
= \sigma(\varphi) & \text{for } w^\top x+ b=0  \: \forall \lambda\\
\end{cases}
$$
Therefore, $\sigma_\lambda(x) = \sigma(\lambda(w^\top x+ b)+\varphi)$ converges pointwise and boundedly to
$$
\gamma(x)
\begin{cases}
= 1 & \text{for } w^\top x+ b>0 \text{ as } \\
= 0 & \text{for } w^\top x+ b<0 \text{ as } \\
= \sigma(\varphi) & \text{for } w^\top x+ b=0\\
\end{cases}
$$

Let $\Pi_{w,b}$ be the hyperplane defined by $\{x | w^\top x + b=0 \}$ and let $H_{w,b}$ be the open half-space defined by $\{x | w^\top x + b>0 \}$. Then, by the Lebesgue Bounded Convergence Theorem (theorem \ref{thm:lebesgue}), for all $\varphi$, $w$, and $b$, we have\footnote{Since $0\le \sigma_\lambda\le M$ and $\sigma_\lambda\to\gamma$ pointwise, the bounded convergence theorem gives $\int \sigma_\lambda\, d\mu \to \int \gamma\, d\mu$.}
\begin{align*}
0 
& = \int_{I_n} \sigma_\lambda(x) d\mu(x) \qquad \forall \lambda \\
0
& = \lim_{\lambda\to\infty}\int_{I_n} \sigma_\lambda(x) d\mu(x) \\
& = \int_{I_n} \gamma(x) d\mu(x) \\
& = \sigma(\varphi) \mu(\Pi_{w,b}) + \mu(H_{w,b})
\end{align*}

We now show that the measure of all half-spaces being $0$ implies that the measure
$\mu$ itself must be 0. This would be trivial if $\mu$ were a positive measure but here it is not.

Why is the measure of all half-spaces equal to $0$? Since $\sigma(\varphi)\mu(\Pi_{w,b}) + \mu(H_{w,b}) = 0$ for every $\varphi$:
\begin{align*}
\sigma(\varphi_1)\mu(\Pi_{w,b}) + \mu(H_{w,b}) &= 0 \\
\sigma(\varphi_2)\mu(\Pi_{w,b}) + \mu(H_{w,b}) &= 0
\end{align*}
Because a sigmoidal function is non-constant, choose $\varphi_1,\varphi_2$ with $\sigma(\varphi_1)\neq\sigma(\varphi_2)$. Subtracting gives
$$\mu(\Pi_{w,b})=0$$
Substituting back gives
$$\mu(H_{w,b})=0$$
Thus every hyperplane and every open half-space has $\mu$-measure zero.

Now we will prove that the measure $\mu$ itself must be 0. Fix $w$. For a bounded measurable function, $h$, define the linear functional, $F: L^\infty(J) \to \mathbb C$ where $J= \{w^\top x:x\in I_n\}$, according to
$$
F(h)=\int_{I_n} h(w^\top x)d\mu(x)
$$
and note that F is a bounded functional on $L^\infty(J)$ since $\mu$ is a finite signed measure. Let $h$ be the indicator function of the half-closed interval $[b, \infty)$:
$$ 
\mathbf 1_{[b,\infty)}(u) = 
\begin{cases} 
  1,& u\ge b,\\ 
  0,& u<b. 
\end{cases}
$$

Then
$$\{x:h(w^\top x)=1\} = \{x:w^\top x\ge b\} = \Pi_{w,-b}\cup H_{w,-b}$$
where
$$\Pi_{w,-b} = \{x:w^\top x=b\}, \qquad H_{w,-b} = \{x:w^\top x>b\}$$
Therefore
$$
\begin{aligned} 
F(h) &=\int_{I_n} h(w^\top x)\,d\mu(x)\\ 
     &=\mu(\{x:w^\top x\ge b\})\\ 
     &=\mu(\Pi_{w,-b})+\mu(H_{w,-b}). 
\end{aligned}
$$

Similarly, $F(h) = 0$ if $h$ is the indicator function of the open interval $(b, \infty)$. By linearity, $F(h) = 0$ for the indicator function of any interval and hence for any simple function (that is, sum of indicator functions of intervals). 
$F$ vanishes on every simple function. Now let $h\in L^\infty(J)$ be bounded and measurable. There exists a sequence of simple functions $\{h_k\}$ such that $h_k\to h$ pointwise and $|h_k|\le \|h\|_\infty$. Since $\mu$ is finite, the Bounded Convergence Theorem (theorem \ref{thm:lebesgue}) gives $F(h) = \lim_{k\to\infty}F(h_k)$. 
Strictly speaking, since $\mu$ is a finite signed measure, the bounded convergence theorem applies to its Jordan decomposition $\mu=\mu^+-\mu^-$.
Because $F(h_k)=0$ for every $k$, it follows that $F(h)=0$. Hence $F\equiv0$ on $L^\infty(J)$.\footnote{
Original derivation: Since simple functions are dense in $L^\infty(J)$ for a compact interval, $J$, containing $\{ w^\top x  | x \in I_n\}$ with respect to the essential-supremum norm (theorem \ref{thm:simple-dense}), $F=0$.
}

In particular, the bounded measurable functions $s(u)= \sin(u)$ and $c(u)=
\cos(u)$ give
\begin{align*}
F(s+ic) 
&= \int_{I_n} \cos(\omega^\top x)+i\sin (\omega^\top x) d\mu(x) \\
&= \int_{I_n} e^{i \omega^\top x} d\mu(x) = 0 \\
\end{align*}
Since the preceding argument holds for every $\omega \in \mathbb R^n$, we obtain $\widehat{\mu}(\omega) = \int_{I_n} e^{i \omega^\top x}\,d\mu(x) = 0$ $\forall \omega \in \mathbb R^n$. Thus the Fourier transform of $\mu$ vanishes identically. 
When the Fourier transform of $\mu$ is 0, $\mu$ must be zero as well:
If $\widehat\mu(\omega) = \int e^{i \omega^\top x}\,d\mu(x)$ vanishes for all $\omega \in \mathbb R^n$, then $\mu=0$ by the uniqueness theorem for Fourier transforms of finite measures.
Hence, $\sigma$ is discriminatory.
\end{proof}

\begin{remark}
The Fourier transform of the measure $\mu$ must be zero because the $\mu$-measure of every half-space is zero. The complete logic argument is: all half-spaces have $\mu$-measure 0 $\Rightarrow$ $F=0$ $\Rightarrow$ $\widehat{\mu}=0$ $\Rightarrow$ $\mu=0$.
\end{remark}

\begin{remark}
The theorem actually proves that, if $\sigma$ is continuous and discriminatory, then single-hidden-layer networks using $\sigma$ are dense in $C(I_n)$. Since every bounded measurable sigmoidal function is discriminatory, as a corollary, every continuous sigmoidal activation function satisfies the universal approximation property.
\end{remark}


\subsection*{Funahashi's Universal Approximation Theorem}


Mollifiers (also known as approximations to the identity) are smooth functions used to create sequences of smooth functions approximating non-smooth (generalized) functions, via convolution. Intuitively, given a (generalized) function, convolving it with a mollifier "mollifies" it: its sharp features are smoothed, while still remaining close to the original.


\begin{definition}[Approximation to the Identity] 
A family of functions \( \{\rho_\varepsilon\}_{\varepsilon>0} \) is called an \emph{approximation to the identity} if \[ \rho_\varepsilon(x)\ge 0, \qquad \int_{\mathbb{R}^n}\rho_\varepsilon(x)\,dx = 1, \] and the mass of \(\rho_\varepsilon\) concentrates near the origin as \(\varepsilon\to0\). 
Equivalently, for every \(\delta>0\), \[ \int_{\|x\|>\delta}\rho_\varepsilon(x)\,dx \longrightarrow 0 \qquad (\varepsilon\to0). \] A \emph{mollifier} is a smooth approximation to the identity. 
\end{definition}

\begin{definition}[Mollifier] \label{def:mollifier}

Let $\rho(x)$ be a function on $\mathbb{R}^n$ which satisfies the
following conditions:
\begin{itemize}
\item $\rho(x) \ge 0$
\item $\rho(x)$ has continuous partial derivatives of all orders.
\item The support of $\rho(x)$ is contained in the unit ball $\|x\| \le 1$.
\item $\int_{\mathbb{R}^n} \rho(x)dx = 1$
\end{itemize}

Then, for $\epsilon>0$, $\rho_\epsilon(x) = (1/\epsilon)^n \rho(x/\epsilon)$.
If $u(x)$ is locally summable, $u \in L^1_{loc}(\mathbb{R}^n)$, define the $\rho_\epsilon \star$ operator, called a mollifier, as 
$$
\rho_\epsilon \star u(x) = \int_{\mathbb{R}^n} \rho_\epsilon(x-y) u(y) dy
$$
\end{definition}

$\rho_\epsilon \star u(x)$ has continuous partial derivatives
of all orders, i.e. $\rho_\epsilon \star u(x) \in C^\infty$.

\begin{proposition}[Differentiation of a Mollified Function] 
Let \(u\in L^1_{\mathrm{loc}}(\mathbb{R}^n)\) and let \(\rho_\varepsilon\) be a mollifier. Then \[ \rho_\varepsilon * u \in C^\infty(\mathbb{R}^n), \] and for every multi-index \(\alpha\), \[ D^\alpha (\rho_\varepsilon * u) = (D^\alpha \rho_\varepsilon) * u. \] Equivalently, \[ \frac{\partial^{|\alpha|}}{\partial x^\alpha} \left( \int_{\mathbb{R}^n} \rho_\varepsilon(x-y)u(y)\,dy \right) = \int_{\mathbb{R}^n} D^\alpha\rho_\varepsilon(x-y)\,u(y)\,dy. \] 
Hence all derivatives of \(\rho_\varepsilon*u\) exist and are continuous. 
\end{proposition}

The support of $\rho_\epsilon \star u(x)$ is contained
in the $\epsilon$ neighborhood of support of $u(x)$
when $u$ has a well-defined support (e.g. $u\in L^1_{\mathrm{loc}}$).

\begin{proposition}[Support of a Mollified Function] 
If \(u\in L^1_{\mathrm{loc}}(\mathbb{R}^n)\), then \[ \operatorname{supp}(\rho_\varepsilon*u) \subseteq \operatorname{supp}(u) + \overline{B_\varepsilon(0)}, \] where \[ A+B = \{a+b : a\in A,\ b\in B\} \] denotes the Minkowski sum of two sets. \end{proposition}

If $u(x)$ is a continuous function with compact support, then
$\rho_\epsilon \star u(x) \to u(x)$ uniformly on $\mathbb{R}^n$ as $\epsilon \to 0$.
If $u(x) \in L^p(\mathbb{R}^n)$ ($p \ge 1$), then $\rho_\epsilon \star u(x) \to u(x)$ in $L^p$ as $\epsilon \to 0$.

In topology, the Tietze extension theorem (also known as the Tietze–Urysohn–Brouwer extension theorem or Urysohn-Brouwer lemma)\footnote{L. E. J. Brouwer and Henri Lebesgue proved a special case of the theorem, when $X$ is a finite-dimensional real vector space. Heinrich Tietze extended it to all metric spaces, and Pavel Urysohn proved the theorem for normal topological spaces.} states that any real-valued, continuous function on a closed subset of a normal topological space can be extended to the entire space, preserving boundedness if necessary.

\begin{theorem}[Tietze Extension Theorem] \label{thm:tietze-extension} 
Let \(X\) be a normal topological space and let \(A\subset X\) be a closed subset. Every continuous function $f:A\to\mathbb{R}$ admits a continuous extension $ \widetilde f:X\to\mathbb{R}$ such that $\widetilde f|_A=f$. 
satisfying that $\operatorname{supp} f$ is compact.
Moreover, if \(f\) is bounded, then \(\widetilde f\) may be chosen so that $\|\widetilde f\|_\infty = \|f\|_\infty$.
\end{theorem}

\begin{corollary} 
Let \(K\subset\mathbb R^n\) be compact and let \(f\in C(K)\). 
Then there exists a function $\widetilde f\in C(\mathbb R^n)$ such that $\widetilde f(x)=f(x), \quad x\in K$. 
\end{corollary}


\begin{proof}[Proof sketch]
Based on \cite{funahashi1989} and \cite{izuki2024cma}.

\begin{itemize}

\item 
Extension:
Extend \(f\) to a continuous function \(\widetilde f\in C(\mathbb{R}^n)\) using Tietze's extension theorem, so that we pass from $f\in C(K)$ to a continuous function $\widetilde f\in C(\mathbb R^n)$ that can then be convolved with the Schwartz approximate identity.

Choose a function \(\phi\in\mathcal S(\mathbb{R}^n)\) whose Fourier transform \(\widehat{\phi}\in C_c^\infty(\mathbb{R}^n)\) satisfies \(\widehat{\phi}(0)=1\). 

For \(\varepsilon>0\), define 
\[ \phi_\varepsilon(x) = \varepsilon^{-n}\phi(x/\varepsilon). \] 

Then 
\[ \widehat{\phi_\varepsilon}(\xi) = \widehat{\phi}(\varepsilon\xi), \] 
so \(\widehat{\phi_\varepsilon}\) has compact support and \(\{\phi_\varepsilon\}_{\varepsilon>0}\) is an approximate identity. 

\item
Convolution:
Convolve \(\tilde f\) with the Schwartz function \(\phi_\varepsilon\):
\[ F_\varepsilon = \widetilde f * \phi_\varepsilon . \] 

Since \(\phi_\varepsilon\) is a Schwartz function, \(F_\varepsilon\in C^\infty(\mathbb R^n)\). 
Moreover, $ \mathcal F [{F_\varepsilon}] = \mathcal F [\widetilde f] \, \mathcal F [\phi_\varepsilon]$. 

Because $\mathcal F[{\phi_\varepsilon}]$ is compactly supported, $\mathcal F [{F_\varepsilon}]$ is compactly supported as well. Hence \(F_\varepsilon\) is band-limited. 

\item 
Aproximation:
Since \(\{\phi_\varepsilon\}\) is an approximate identity and \(\widetilde f\) is uniformly continuous on compact sets, \[ F_\varepsilon = \widetilde f * \phi_\varepsilon \longrightarrow \widetilde f \] uniformly on \(K\) as \(\varepsilon\to0\). 

Therefore, for every \(\delta>0\), there exists \(\varepsilon>0\) such that \[ \sup_{x\in K} |F_\varepsilon(x)-f(x)| < \delta. \] Thus \(f\) can be uniformly approximated on \(K\) by smooth band-limited functions.

\end{itemize}
\end{proof}



\begin{lemma}[Information on $\phi(z)$] \label{lemma:funahashi-info}
Let $\phi(z)$ be a non-constant, bounded, and monotonically increasing continuous function. Then there exist constants $\delta, \alpha >0$ such that $\psi(z)$ is integrable (i.e. $\psi(z) \in L^1(\mathbb R)$) and $\mathcal F[\psi](1) \neq 0$, where $\psi(z) = \phi(z/\delta+\alpha) - \phi(z/\delta+\alpha)$.
In particular, $\psi(z)$ is real-valued because $\phi(z)$ is real-valued.
\end{lemma}

\begin{proof}
Let $L, L' >0$ be arbitrary large numbers. Then $\psi(z)$ is integrable:
\begin{align*}
    \int_{-L'}^L \psi(z)dz 
     & = \delta \int_{-L'/\delta+\alpha}^{L/\delta+\alpha} \phi(s)ds 
       - \delta \int_{-L'/\delta-\alpha}^{L/\delta-\alpha} \phi(s)ds \\
     & = \delta \int_{L/\delta-\alpha}^{L/\delta+\alpha} \phi(s)ds 
       - \delta \int_{-L'/\delta-\alpha}^{-L'/\delta+\alpha} \phi(s)ds \\  
     & \in [0, 4 \delta \alpha \sup |\phi|]
\end{align*}

If $\mathcal F[\psi](1)=0$ for all $\delta>0$, then $\mathcal F[\phi(\cdot+\alpha)-\phi(\cdot-\alpha)]=0$, so $\phi(z+\alpha)=\phi(z-\alpha)$. If $u = z-\alpha$, when $\phi(u)=\phi(u+2\alpha)$. $\phi(z)$ would be a periodic function with period $2\alpha$. Combining this periodicity with the assumption that $\phi$ is increasing, $\phi$ would be constant on $[0, 2\alpha]$ and, again from its periodicity, $\phi(z)$ would be constant on $\mathbb R$. But this contradicts the assumption that $\phi$ is
not constant. Therefore, $\mathcal F[\psi](1) \neq 0$.
\end{proof}

Funahashi applies the continuous Fourier transform to prove a discrete theorem, so we need to transform continuous representations into discrete ones:

\begin{lemma}[From continuous to discrete] \label{lemma:funahashi-discrete}

Let $A>0$, $K \subset \mathbb R^n$ a compact set, and $h(w,x)$ a continuous function on $[-A,A]^n \times K$. Let us define the functions $H(x)$ and $H_N(x)$, with $N\in \mathbb N$ on K:
\begin{align*}
    H(x) &= \int_{[-A,A]^n} h(w,x)dw,\\
    H_N(x) &= \kappa^N \sum_{k_1 .. k_N = 0}^{N-1} h \left( -A+\kappa k_1, ..., -A+\kappa k_n, x \right)
\end{align*}
with $\kappa=2A/N$. Then, for all $\epsilon>0$, there exists $N_0 \in \mathbb N$ such that $\max_{x \in K} |H(x)-H_N(x)| < \epsilon$ for all $N \geq N_0$.
\end{lemma}

\begin{proof}
Let $\mathbf 1 = (1,1,.., 1) \in \mathbb R^n$ and $\mathbf k \in \{1,2,..,N-1\}^n$ meaning that $\mathbf k=(k_1,k_2,..,k_n)$ for every integer $k_j \in \{1,2,..,N-1\}$, $j = 1..n$. Then
$$
\sum_{\mathbf k \in \{1,2,..,N-1\}^n} = \sum_{k_1 .. k_N = 0}^{N-1}
$$
For any $\mathbf k$:
$$
\left( -A+\kappa k_1, ..., -A+\kappa k_n, x \right) = -A \mathbf 1 + \kappa \mathbf k = -A \mathbf 1 + \frac{2A}{N} \mathbf k
$$
Putting $H_N(x)$ in vector notation:
$$
H_N(x) = \kappa^N \sum_{\mathbf k \in \{1,2,..,N-1\}^n} h ( -A \mathbf 1 + \kappa \mathbf k, x)
$$
We evaluate $|H(x)-H_N(x)|$ and, by the uniform continuity of $h(w,x)$, for any $\epsilon>0$, there exists $\delta>0$ such that
$$
|h(w,x) - h(w',x)| < \frac{\epsilon}{(2A)^n}
$$
for any $w, w' \in \mathbb R^n$ satisfying $|w-w'|<\delta$.
Then, we fix $N_0\in \mathbb N$ so that $\frac{2A}{N_0} \sqrt{N} < \delta$ and let $N>N_0$:
$$
\left| w - \left(-A \mathbf 1 + \frac{2A}{N} \mathbf k \right) \right| <  \frac{2A}{N_0} \sqrt{N} < \delta
$$
$$
w \in \prod_{j=1}^n \left[ -A+\frac{2k_j A}{N}, -A+\frac{2(k_j+1) A}{N}\right]
$$
So we obtain the following approximation
$$
\left| h(w,x) - h\left(-A \mathbf 1 + \frac{2A}{N} \mathbf k,x\right)\right| < \frac{\epsilon}{(2A)^n}
$$

For each $\mathbf k \in \{1,2,..,N-1\}^n$, we define
$$
C(\mathbf k) = \prod_{j=1}^n \left[ -A+\frac{2k_j A}{N}, -A+\frac{2(k_j+1) A}{N}\right]
$$
Then, by the definition of $H_N(x)$ in vector notation and the above approximation:
\begin{align*}
& |H(x) - H_N(x)| \\
& \leq \sum_{\mathbf k} 
    \left|  \int_{C(\mathbf k)} h(w,x)dw 
    - \int_{C(\mathbf k)}h\left(-A \mathbf 1 + \frac{2A}{N} \mathbf k,x\right)dw \right| \\
& = \sum_{\mathbf k}     
    \left|  \int_{C(\mathbf k)} \left( h(w,x) 
    -  h\left(-A \mathbf 1 + \frac{2A}{N} \mathbf k,x\right) \right) dw \right| \\
& \leq \sum_{\mathbf k} \frac{\epsilon}{(2A)^n} \left( \frac{2A}{N} \right)^n \\
& \leq \epsilon
\end{align*}
\end{proof}

By the use of the Fourier transform in the real line, we can approximate the Fourier inverse transform of the Fourier transform \cite{izuki2024cma}. Let us represent the  n-dimensional hypercube centered at the origin with a side length of 2A (i.e. a closed ball under the infinity norm) as $I_A^n = B_\infty[0,A] = [-A,A]^n$ and define the following functions; 
$$
M[f](w) = \left( \frac{1}{2\pi} \right)^n \frac{\mathcal F[f](w)}{\mathcal F[\psi](1)}
$$
$$
I_{A',A}[f](x) = \int_{I_A^n} \int_{-A'}^{A'} \psi(x \cdot w-w_0) \cdot M[f](w) e^{iw_0} dw_0 dw
$$
$$
I_{\infty,A}[f](x) = \lim_{A' \to \infty} I_{A',A}[f](x) 
$$
$$
J_{A}[f](x) = \left(\frac{1}{2\pi}\right)^n \int_{I_A^n} F[f](w) e^{i w \cdot x} dw
$$

\begin{lemma}[Convergence of $I$ and $J$] \label{lemma:funahashi-convergence}
Assuming that $f(x) \in L^1(\mathbb R^n)$ satisfies $\mathcal F[f](w) \in L^1(\mathbb R^n)$ (i.e. both $f(x)$ and its Fourier transform are integrable), for all $0 < A < \infty$ and $x \in \mathbb R^n$, 
$$
I_{\infty,A}[f](x) = J_{A}[f](x)
$$
In addition, both $\{I_{\infty,A}[f](x)\}_{A>0}$ and $\{J_{A}[f](x)\}_{A>0}$ converge uniformly in $\mathbb R^n$.
\end{lemma}

\begin{proof}
Given $\psi(z)$ defined in Lemma \ref{lemma:funahashi-info}, we can define
$$
\Psi_{A'}(x) = \int_{-\infty}^{\infty} \psi(z) e^{-iz} \chi_{[xw-A',xw+A']}(z)dz
$$
where $\chi_K$ be a smooth cutoff that equals 1 on $K$ and decays outside a slightly larger compact set. By the Lebesgue dominated convergence theorem (Theorem \ref{thm:lebesgue}),
$$
\lim_{A' \to \infty} \Psi_{A'}(x)  = \mathcal F[\psi](1)
$$

To prove that $I_{\infty,A}[f](x) = J_{A}[f](x)$, it suffices to prove that
\begin{align*}
\lim_{A' \to \infty} & \int_{I_A^n} \mathcal F[f](w) e^{iw \cdot x} \Psi_{A'}(x) dw \\
& = \int_{I_A^n} \mathcal F[f](w) e^{iw \cdot x} F[\psi](1) dw
\end{align*}
We can observe that
$$
\left| \mathcal F[f](w) e^{iw \cdot x} \Psi_{A'}(x) \right| \leq |F[f](w)| \cdot || \psi ||_{L^1(\mathbb R)}
$$
and that bound is independent of $A'$ and integrable on $[-A,A]^n$. Therefore, applying the Lebesgue dominated convergence theorem again, we obtain the desired result, so $I_{\infty,A}[f](x) = J_{A}[f](x)$.

Moreover, $\{J_{A}[f](x)\}_{A>0}$ converges to $\mathcal F^{-1}[\mathcal F[f]](x)$ uniformly in $\mathbb R^n$. Since $\mathcal F[f](w)$ in integrable:
\begin{align*}
& \sup_{x\in \mathbb R`^n} \left| \mathcal F^{-1}[\mathcal F[f]](x) - J_{A}[f](x)\right| \\
& = \left( \frac{1}{2\pi} \right)^n \sup_{x\in \mathbb R`^n} \left| \int_{\mathbb R^n} \mathcal F[f](w) e^{i w \cdot x} (1-\chi_{I_A^n}(w)) dw\right| \\
& \leq \left( \frac{1}{2\pi} \right)^n \int_{\mathbb R^n} \left| \mathcal F[f](w) \right| (1-\chi_{I_A^n}(w)) dw \\
& \to 0 \quad (A \to \infty)
\end{align*}
\end{proof}

\repthm{thm:funahashi}

\begin{remark}
As in Cybenko's universal approximation theorem, $\tilde f(x)$ belongs to the space of standard three-layer feedforward neural networks: $n$ inputs, one hidden layer containing $h$ neurons with activation function $\phi$, and a single linear output neuron. The universal approximation theorem states that a multilayer network composed of at least one hidden layer can approximate any continuous function if the number of hidden units is large enough and the parameters (weights and thresholds) are appropriately adjusted.
\end{remark}

\begin{proof} \cite{izuki2024cma}
We know that $I_{\infty,A}[f](x) = J_{A}[f](x)$ for all $0 < A < \infty$ and $x \in \mathbb R^n$  (Lemma \ref{lemma:funahashi-convergence}). Let us suppose that $f \in C_c^\infty (\mathbb R^n)\subset S(\mathbb R^n)$,
$$
f(x) = \mathcal F^{-1}[\mathcal F[f]](x) = \lim_{A \to \infty} J_A[f](x) = \lim_{A \to \infty} I_{\infty,A}[f](x)
$$
which converges uniformly in $\mathbb R^n$. Therefore, there exists $A_0>0$ such that for all $A>A_0$:
$$
\max_{x \in \mathbb R^n} \left| f(x) - I_{\infty,A}[f](x) \right| < \frac{\epsilon}{3}
$$

We can approximate $I_{\infty,A}[f](x)$ on $K$ using $I_{A',A}[f](x)$ with $0<A'<\infty$. For a particular $x\in K$:
\begin{align*}
& \left| I_{\infty,A}[f](x) - I_{A',A}[f](x)\right| \\
& \leq \frac{1}{(2\pi)^n |\widehat \psi(1)|} \int_{I_A^n} |\widehat f(w)| \int_{\mathbb R \setminus [-A,A]} |\psi(x \cdot w - w_o)|dw_0 dw \\
& = \frac{1}{(2\pi)^n |\widehat \psi(1)|} \int_{I_A^n} |\widehat f(w)| \int_{-\infty}^{\infty} |\psi(t)| \chi_{D} (t) dt dw
\end{align*}
where $\widehat \psi = \mathcal F[\psi]$, $\widehat f=\mathcal F[f]$, and $D=\mathbb R \setminus[x \cdot w - A', x \cdot w + A']$. Given that $K$ is bounded, there exists $R>0$ with $K\subset B(R)$. For $w\in I_A^n = [-A,A]^n$, $|x \cdot w| \leq ||x|| \, ||w|| \leq R \cdot \sqrt{n} A$:
\begin{align*}
D &= \mathbb R \setminus[x \cdot w - A', x \cdot w + A'] \\
  &= (-\infty, x \cdot w - A') \cup (x \cdot w + A', +\infty) \\
  &\subset (-\infty, \sqrt{n} RA - A') \cup (-\sqrt{n} RA + A', +\infty) =: E\\
\end{align*}
The set $E$ is independent of $x$ and $w$. Hence:
\begin{align*}
& (2\pi)^n |\widehat \psi(1)| \max_{x\in K}\left| I_{\infty,A}[f](x) - I_{A',A}[f](x)\right| \\
& \leq \int_{I_A^n} |\widehat f(w)| dw \cdot \max_{x\in K, \, w \in I_A^n} \int_{-\infty}^{\infty} |\psi(t)| \chi_{D} (t) dt \\
& \leq \int_{I_A^n} |\widehat f(w)| dw \cdot \int_{-\infty}^{\infty} |\psi(t)| \chi_{E} (t) dt \\
\end{align*}
We can observe that
$\lim_{A' \to \infty} |\psi(t)| \chi_{E} (t) =0$, 
$|\psi(t)| \in L^1(\mathbb R)$,
and $|\psi(t)| \chi_{E} (t) \leq |\psi(t)|$. Using the Lebesgue dominated convergence theorem once more:  
$\lim_{A' \to \infty} \int_{-\infty}^{\infty} |\psi(t)| \chi_{E} (t) dt = 0$.

In consequence, there exists $A'_0 >0$ so that for all $A'>A'_0$:
$$
\max_{x\in K}\left| I_{\infty,A}[f](x) - I_{A',A}[f](x)\right| < \frac{\epsilon}{3}
$$
Combining this with our previous approximation of $f(x)$ and $I_{\infty,A}[f](x)$, we obtain
$$
\max_{x\in K}\left| f(x) - I_{A',A}[f](x)\right| < \frac{2\epsilon}{3}
$$

This result holds for $f \in C_c^\infty (\mathbb R^n)$. Let us now consider the general case, where $f$ is a continuous function defined over K.

{\it Extension: }
Let us now choose a real-valued extension $g(x) \in C_c (\mathbb R^n)$ of $f(x)$, which is valid due to the Tietze extension theorem  (Theorem \ref{thm:tietze-extension}).

Let $\rho(x) \in C_c^\infty (\mathbb R^n)$ be such that $0 \leq \rho(x) \leq \chi_{B(1)} (x)$ for all $x\in \mathbb R^n$ and $||\rho||_{L_1(\mathbb R^n)} = 1$.
Applying a mollifier $\rho_\beta(x) = \beta^{-n} \rho(\beta^{-1} x)$ (Definition \ref{def:mollifier}) to $g(x)$, we find $\beta \in (0,1)$ such that
$$
|| g - \rho_\beta * g||_{L^\infty(\mathbb R^n)} < \frac{\epsilon}{3}
$$
Geometrically, $\operatorname{supp}(g) \subset \operatorname{supp}(g * \rho_\beta)$ and $\operatorname{supp}(\rho_\beta * g)$ is in a compact set $L$. 
Since $\rho_\beta * g (x) \in C_c^\infty (\mathbb R^n)$, we can apply our previous approximation result for infinitelly differentiable functions in $\mathbb R^n$:
$$
\max_{x\in \operatorname{supp}(g * \rho_\beta)}\left| \rho_\beta * g(x) - I_{A',A}[\rho_\beta * g](x)\right| < \frac{2\epsilon}{3}
$$
Since $g(x)$ is an extension of $f(x)$:
\begin{align*}
& \max_{x\in K}\left| f(x) - I_{A',A}[\rho_\beta * g](x)\right| \\
&= \max_{x\in K} \left| \rho_\beta * g(x) - I_{A',A}[\rho_\beta * g](x)\right|
\end{align*}
Therefore,
\begin{align*}
& \max_{x\in K} \left| f(x) - I_{A',A}[\rho_\beta * g](x)\right| \\
& \leq \max_{x\in \operatorname{supp}(g)}\left| g(x) - I_{A',A}[\rho_\beta * g](x)\right| \\
& \leq \max_{x\in \operatorname{supp}(g)}\left| g(x) - \rho_\beta * g(x) \right| \\
& + \max_{x\in \operatorname{supp}(g)}\left| \rho_\beta * g(x) - I_{A',A}[\rho_\beta * g](x)\right| \\
& \leq || g(x) - \rho_\beta * g(x) ||_{L^\infty(\mathbb R^n)} \\
& + \max_{x\in \operatorname{supp}(g)}\left| \rho_\beta * g(x) - I_{A',A}[\rho_\beta * g](x)\right| \\
& < \epsilon
\end{align*}

$f(x)$ is real-valued but $I_{A',A}[\rho_\beta * g](x)$ is complex-valued. Its real part, $H(x)$, can be used to approximate $f(x)$:
\begin{align*}
& \left| f(x) - I_{A',A}[\rho_\beta * g](x)\right| \\
& \geq \left| Re(f(x) - I_{A',A}[\rho_\beta * g](x)) \right| \\
& = \left| f(x) - H(x) \right|
\end{align*}
i.e. 
$$
\max_{x\in K} \left| f(x) - H(x) \right| < \epsilon
$$
Applying Lemma \ref{lemma:funahashi-discrete} to $H(x)$:
$$
\max_{x\in K} \left| H(x) - H_N[f](x) \right| < \epsilon
$$
where
\begin{align*}
H_N[f](x) & = \left(\frac{2A}{N}\right)^n \sum_{\mathbf k \in \{1,2,..,N-1\}^n} h \left( -A \mathbf 1 + \frac{2A}{N} \mathbf k, x \right) \\
h(w,x) &= \int_{-A'}^{A'} \psi(x \cdot w - w_0) \gamma(w,w_0) d w_0 \\
\gamma(w,w_0) & = Re \left(  \left( \frac{1}{2\pi} \right)^n \frac{\mathcal F[\rho_\beta * g](w)}{\mathcal F[\psi](1)} e^{iw_0} \right)
\end{align*}
Hence,
$$
\max_{x\in K} \left| f(x) - H_N[f](x) \right| < 2\epsilon
$$

We apply Lemma \ref{lemma:funahashi-discrete} again, now to $H_N(x)$ in order to obtain our desired function $\tilde f(x)$:
\begin{align*}
& \left(\frac{2A}{N}\right)^{-n} H_N[f](x) \\
& = \sum_{\mathbf k \in \{1,2,..,N-1\}^n} h \left( -A \mathbf 1 + \frac{2A}{N} \mathbf k, x \right) \\
& = \int_{-A'}^{A'} \sum_{\mathbf k \in \{1,2,..,N-1\}^n} \psi \left(x \cdot \left( -A \mathbf 1 + \frac{2A}{N} \mathbf k \right) - w_0\right) \\
& \cdot \gamma\left(\left( -A \mathbf 1 + \frac{2A}{N} \mathbf k \right),w_0 \right) d w_0
\end{align*}
Approximating by a Riemann sum, we get 
\begin{align*}
& \left(\frac{2A}{N}\right)^{-n} H_N[f](x) \\
\approx & R_M[f](x) \\
= & \frac{2A'}{M} \sum_{m=0}^{M-1} \sum_{\mathbf k \in \{1,2,..,N-1\}^n} \\
& \psi \left(x \cdot \left( -A \mathbf 1 + \frac{2A}{N} \mathbf k \right) - \left( -A'+\frac{2mA'}{M} \right) \right) \\
& \cdot \gamma\left(\left( -A \mathbf 1 + \frac{2A}{N} \mathbf k \right), -A'+\frac{2mA'}{M} \right)
\end{align*}

Using Lemma \ref{lemma:funahashi-discrete}, we can find
$$
\max_{x\in K} \left| \left(\frac{2A}{N}\right)^{-n} H_N[f](x) - R_M[f](x) \right| < \left(\frac{2A}{N}\right)^{-n} \epsilon
$$

Combining this estimate with our previous approximation of $f(x)$ to $H_N[f](x)$, we get
$$
\max_{x\in K} \left| f(x) - \left(\frac{2A}{N}\right)^{n} R_M[f](x) \right| < 3 \epsilon
$$

In order to conclude our proof, we just have to show that $\left(\frac{2A}{N}\right)^{n} R_M[f](x)$ is precisely our desired function $\tilde f(x)$. We obtain an equivalent expression for $R_M[f](x)$ as follows:
\begin{align*}
& R_M[f](x) \\
= & \frac{2A'}{M} \sum_{m=0}^{M-1} \sum_{\mathbf k \in \{1,2,..,N-1\}^n} \\
& \psi \left( (x,-1) \cdot \left( -A \mathbf 1 + \frac{2A}{N} \mathbf k, -A'+\frac{2mA'}{M} \right) \right) \\
& \cdot \gamma\left(\left( -A \mathbf 1 + \frac{2A}{N} \mathbf k \right), -A'+\frac{2mA'}{M} \right)
\end{align*}
We define
$$
\mathbf \Omega(m,\mathbf k) =  \left( -A \mathbf 1 + \frac{2A}{N} \mathbf k, -A'+\frac{2mA'}{M} \right) \in \mathbb R^{n+1}
$$
The set $\{ \mathbf \Omega(m,\mathbf k) \}$ contains $N^n M$ vectors in $\mathbb R^{n+1}$. We can refer to them as $\mathbf \Omega(l)$, with $l=1..N^nM$, so each vector is $\mathbf \Omega(l) = (\Omega_{l,1}, .., \Omega_{l,n+1})$.

Given the definition of $\psi(x)$, we have
\begin{align*}
& R_M[f](x) \\
&= \frac{2A'}{M} \sum_{l=1}^{N^n M} 
     \psi ( (x,-1) \cdot \mathbf \Omega(l)) \gamma (\mathbf \Omega(l)) \\
&= \frac{2A'}{M} \sum_{l=1}^{N^n M} 
     \gamma (\mathbf \Omega(l)) \psi \left( \sum_{j=1}^n x_j \Omega_{l,j} - \Omega_{l,n+1} \right) \\
& = \frac{2A'}{M} \sum_{l=1}^{N^n M} 
     \gamma (\mathbf \Omega(l)) \phi \left( \sum_{j=1}^n \frac{x_j \Omega_{l,j}}{\delta} - \left( \frac{\Omega_{l,n+1}}{\delta} - \alpha \right)\right) \\
& - \frac{2A'}{M} \sum_{l=1}^{N^n M} 
     \gamma (\mathbf \Omega(l)) \phi \left( \sum_{j=1}^n \frac{x_j \Omega_{l,j}}{\delta} - \left( \frac{\Omega_{l,n+1}}{\delta} + \alpha \right)\right)  
\end{align*}
Rearranging, we can get 
\begin{align*}
\left(\frac{2A}{N}\right)^{n} R_M[f](x) 
& = \sum_{j=1}^{2N^n M} \alpha_j \phi \left( \sum_{i=1}^n w_{ij} x_i + b_j \right)\\
& = \tilde{f}
\end{align*}
Hence, our approximation of $\left(\frac{2A}{N}\right)^{n} R_M[f](x)$ concludes the proof of the theorem:
$$
\max_{x\in K} \left| f(x) - \tilde f(x) \right| < 3 \epsilon
$$
\end{proof}

\begin{remark}
Please, observe how we might need an exponential number of hidden neurons in order to obtain the desired approximation.
\end{remark}


Funahashi's original proof \cite{funahashi1989}  is based on an integral representation formula due to Bunpei Irie and Sei Miyake \cite{irie1988}, which suggests the realization of functions by three-layer (one-hidden-layer) networks with an infinite number of hidden neurons.\footnote{In Irie and Miyake's integral formula, the activation function for the hidden layer must satisfy the condition of absolute integrability, so it cannot be a sigmoid function.}
From this, the general case (for any number of hidden layers) can be proved by induction and, for the two hidden layers case,\footnote{An easy-to-understand geometric explanation of how neural networks with two hidden layers can be used to produce the desired ``bumps'' to approximate surfaces can be found in \cite{lapedes1987}.} Funahashi also included a simpler proof using the Kolmogorov-Arnold-Sprecher theorem \cite{kolmogorov1957, arnold1957, sprecher1965}, which had been already used by Robert Hetch-Nielsen to indicate that any continuous mapping can be represented by a four-layer (two-hidden-layer) neural network \cite{hechtnielsen1987}.

\repthm{thm:funahashi}

\begin{proof}[Sketch of Funahashi's original proof] \,

The following sketch presents a Fourier-analytic interpretation of why sigmoidal ridge functions are capable of approximating arbitrary continuous functions:
Finite superpositions of ridge functions $ x\mapsto \sigma(w\cdot x+b) $, where $w\in\mathbb R^n$ and $b\in\mathbb R$, are dense in \(C(K)\) for every compact \(K\subset\mathbb R^n\), 
i.e. they approximate arbitrary continuous functions on compact sets. 

Irie and Miyake \cite{irie1988} proved that arbitrary functions can be represented by a three-layer network with an infinite number of hidden units, i.e. $\lim_{A \to \infty} || I_{\infty,A}[f](x) - f(x) ||_{L^2} = 0$, provided that the activation function is absolutely integrable ($\phi(z) \in L^1(\mathbb R)$).


Funahashi's strategy is as follows:
Let \(f\in C(K)\), where \(K\subset\mathbb R^n\) is compact. By Lemma~\ref{lemma:approximation-band-limited}, \(f\) may be uniformly approximated on \(K\) by a smooth band-limited function \(F\). Since \(F\) is band-limited, the Paley--Wiener theorem implies that \(F\) extends to an entire function of exponential type. Thus the approximation problem may be viewed in terms of reconstructing a function from compactly-supported frequency data. Ridge functions $x\mapsto \sigma(w\cdot x+b)$ introduce frequency information along selected directions \(w\). By varying the directions and offsets, finite linear combinations of ridge sigmoids generate a rich family of approximating functions. Funahashi establishes that this family is sufficiently rich to be dense in \(C(K)\). Consequently, for every \(f\in C(K)\) and every \(\varepsilon>0\), there exists a finite superposition 
\[ \tilde f(x) = \sum_{j=1}^{h} \alpha_j\,\sigma(w_j\cdot x+b_j) \] 
such that 
\[ \sup_{x\in K}|f(x)-\tilde f(x)|<\varepsilon. \]

\noindent {\it Step 1:\\Reduction to approximation of band-limited functions}.

Let \(f\in C(K)\), where \(K\subset\mathbb R^n\) is compact, and let \(\varepsilon>0\). By Lemma~\ref{lemma:approximation-band-limited}, there exists a smooth band-limited function \(F\in C^\infty(\mathbb R^n)\) such that $\sup_{x\in K}|f(x)-F(x)| < \frac{\varepsilon}{2}$. Therefore, it suffices to approximate the band-limited function \(F\) on \(K\).


Since \(F\) is band-limited, there exists \(R>0\) such that $\operatorname{supp}(\widehat F) \subseteq B_R(0)$. By the Paley--Wiener theorem (Corollary \ref{cor:bandlimited-entire}), \(F\) extends to an entire function on \(\mathbb C^n\) whose growth is controlled by the bandwidth \(R\). Thus \(F\) possesses a highly structured spectral representation. The Paley--Wiener theorem (Theorem \ref{thm:paley-wiener}) provides a bridge between band-limited functions and entire functions of exponential type, which motivates the use of Fourier-analytic techniques in the approximation problem.

\noindent {\it Step 2: The family generated by sigmoids has rich frequency content}.

The key idea is that ridge sigmoids possess sufficiently rich frequency content to approximate band-limited functions. 
Consider a unit vector $u\in\mathbb{S}^{n-1}$ and functions of the form $\sigma(\lambda\,u\cdot x + b)$, with $\lambda>0$, $b\in\mathbb{R}$. As functions of the scalar variable $t=u\cdot x$, these are translations/dilations of $\sigma$.

Translations and dilations of a ridge sigmoid alter its location, offset, and characteristic scale. Varying the parameters \(w\) and \(b\) produces a rich collection of ridge functions with different spectral characteristics. From the Fourier viewpoint, translations primarily affect phase, while dilations modify the characteristic frequency scale.

Under mild regularity assumptions on \(\sigma\) (sigmoidal, non-polynomial, typically of bounded variation or absolutely integrable derivative),  the Fourier transform $\widehat{\sigma}$ of a sigmoid $\sigma$ is nontrivial and exhibits controlled decay properties.
This suggests that finite superpositions of ridge sigmoids may be used to reproduce increasingly complex spectral patterns. In other words, finite superpositions of ridge sigmoids can synthesize spectra by placing energy along chosen directions and phases, allowing the approximation of band-limited targets by finite superpositions.
\noindent {\it Step 3: Approximation in \(L^2\) by finite superpositions}.

A central result underlying Funahashi's theorem is that finite superpositions of non-polynomial ridge sigmoids form a class rich enough to approximate band-limited functions.

The spectral synthesis consists of constructing linear combinations whose Fourier transforms approximate $\mathcal{F}(F)$ in $L^2$\ over the compact support of $\mathcal{F}(F)$:
\[
\tilde F(x)=\sum_{j=1}^h \alpha_j\,\sigma(w_j\cdot x + b_j)
\]
so that $\alpha_j$, $w_j$, and $b_j$ match the amplitudes, directions, and phases of $\mathcal{F}(F)$.

\noindent {\it  Step 4: From \(L^2\) approximation to uniform approximation on compact sets}.

Since $F$ is band-limited, standard uniform bounds transfer this small $L^2$ error to uniform error on $K$ after truncation/localization.
This is a standard mathematical argument in numerical analysis that describes how to transfer a mean-error guarantee (in the $L^2$ norm) into a pointwise, uniform-error guarantee (in the $L^{\infty }$  norm) over a bounded domain $K$.\footnote{
In the $L^2$ approximation, the approximating function is close to the target function F on average (the overall energy of the error is small). However, even though $L^2$ is bounded, the functions could still match poorly at isolated points, creating massive spikes of error that $L^2$ integration ignores. In the uniform approximation on compact sets, error is strictly bounded at every single point within a closed and bounded region. The \(L^{\infty }\) norm guarantees there are no hidden spikes of error. You cannot jump from $L^2$ to uniform convergence without restrictions. The function must behave predictably, which is guaranteed by smoothness (continuous derivatives prevent the function from changing abruptly) or band-limitedness (the Fourier transform is zero outside a finite frequency range).
}

Combining the reduction to band-limited functions and the spectral approximation, we obtain our desired result: for any \(\varepsilon>0\), there exists a finite superposition $\tilde{f}(x)$  such that \(\sup_{x\in K}|f(x)-\tilde{f}(x)|<\varepsilon\). Hence, finite sums of the form $\tilde{f}(x)$ are dense in $C(K)$.


In summary, by showing that the span of shifted/dilated sigmoids is dense in the Paley–Wiener class, Funahashi concluded that they are dense in \(C(K)\).

\end{proof}

\begin{proof}[Proof by contradiction]
For simplicity, we present a Fourier-analytic proof under the additional assumption that \(\sigma\) is a non-constant sigmoidal function whose derivative belongs to \(L^1(\mathbb R)\), i.e. \(\sigma'\in L^1(\mathbb R)\). Funahashi's original argument removes this regularity assumption by a more delicate analysis.

This proof is based on a density argument combining functional analysis and Fourier analysis. The key idea is to show that any continuous linear functional that annihilates all ridge functions must vanish identically. By the Hahn--Banach theorem, this implies that the span of ridge functions is dense in \(C(K)\).

\begin{itemize}

\item
Assume, for contradiction, that the span of the ridge functions $\sigma(w\cdot x+b)$ is not dense in \(C(K)\). 

\item
Then, by the Hahn--Banach theorem (Theorem \ref{thm:hahn-banach}), there exists a nonzero continuous linear functional $L: C(K)\to\mathbb R$ such that $L(\sigma(w\cdot x+b))=0$ for all \(w\in\mathbb R^n\) and \(b\in\mathbb R\).

\item
By the Riesz representation theorem (Theorem \ref{thm:riesz}), there exists a nonzero finite signed Borel measure \(\mu\) on \(K\) such that $ L(f) = \int_K f(x)\,d\mu(x)$.
Hence $ \int_K \sigma(w\cdot x+b) \,d\mu(x) = 0 $ for all \(w\in\mathbb R^n\) and $b\in\mathbb R$.

\item
Now define $F_w(t) = \int_K \sigma(w\cdot x+t) \,d\mu(x)$. Then $F_w(t)\equiv0$.

\item
Because \(\sigma'\in L^1(\mathbb R)\), differentiation may be passed under the integral sign, yielding \[ F_w'(t) = \int_K \sigma'(w\cdot x+t)\,d\mu(x). \] Since \(F_w\equiv0\), it follows that \(F_w'\equiv0\).

\item
Taking the Fourier transform of $F_w'(t) = \int_K \sigma'(w\cdot x+t)\,d\mu(x)$ with respect to \(t\) yields \[ \widehat{F_w'}(\xi) = \widehat{\sigma'}(\xi) \int_K e^{-i\xi w\cdot x}\,d\mu(x). \] Since \(F_w'\equiv0\), we have \[ 0= \widehat{\sigma'}(\xi) \int_K e^{-i\xi w\cdot x}\,d\mu(x). \]

\item
Since \(\sigma'\in L^1(\mathbb R)\), its Fourier transform \(\widehat{\sigma'}\) is continuous. Because \(\sigma\) is nonconstant, its derivative cannot vanish almost everywhere. Hence \(\sigma'\not\equiv0\) in \(L^1(\mathbb R)\), and by injectivity of the Fourier transform on \(L^1(\mathbb R)\), $\widehat{\sigma'}\not\equiv0$. Hence there exists \(\xi_0\in\mathbb R\) such that $\widehat{\sigma'}(\xi_0)\neq0$. By continuity, there exists a nonempty open interval $I\subset\mathbb R$ containing $\xi_0$ for which $\widehat{\sigma'}(\xi)\neq 0$ ($\xi\in I$).
Hence $\int_K e^{-i\xi w\cdot x} \,d\mu(x) = 0$ for all $w$ and all $\xi\in I$.

\item
Let $\eta=\xi w$. Then $\widehat{\mu}(\eta) = \int_K e^{-i\eta\cdot x} \,d\mu(x) = 0$ on a nonempty open subset of $\mathbb R^n$.

Since $I=(a,b)$ contains some $\xi_0\neq0$, choose a smaller interval \(J\subset I\) such that \(0\notin J\). Then \[ U=\{\xi w:\xi\in J,\;w\in\mathbb R^n\} =\mathbb R^n\setminus\{0\}, \] which is a nonempty open subset of \(\mathbb R^n\). Hence \(\widehat{\mu}\) vanishes on a nonempty open set.

\item
Because \(\mu\) is supported on the compact set \(K\), the Paley--Wiener theorem (Theorem \ref{thm:paley-wiener}) implies that its Fourier transform \[ \widehat{\mu}(\eta) = \int_K e^{-i\eta\cdot x}\,d\mu(x) \] extends to an entire function on \(\mathbb C^n\).

\item
An entire function vanishing on a nonempty open set is identically zero; therefore
$\widehat{\mu}\equiv0$. 

Since $\widehat\mu$ extends to an entire function on $\mathbb C^n$ and vanishes on a nonempty open subset of $\mathbb R^n$. Therefore, it vanishes on a nonempty open subset of $\mathbb C^n$ relative to the real slice. By the identity theorem for holomorphic functions of several variables,
$\widehat\mu\equiv 0$.

\item
The Fourier transform is injective on finite signed Borel measures.
Hence $\widehat{\mu}\equiv0$ implies $\mu=0$, contradicting the choice of \(\mu\neq0\).
\end{itemize}

Therefore, no nonzero annihilating functional exists (the annihilator of the ridge-function span is trivial). 
By the Hahn--Banach theorem (Corollary \ref{cor:hahn-banach-annihilator}), \[ \overline{ \operatorname{span} \{\sigma(w\cdot x+b): w\in\mathbb R^n,\, b\in\mathbb R \} } = C(K). \] 
Hence the span of ridge functions is dense in \(C(K)\).

\end{proof}

\begin{remark}
The assumption that \(\sigma\) is non-polynomial is essential. If \(\sigma\) is a polynomial of degree \(m\), then every finite superposition $\sum_{j=1}^h \alpha_j\, \sigma(w_j\cdot x+b_j)$ is itself a polynomial of degree at most \(m\). 

Consequently, the resulting function class is contained in the finite-dimensional vector space of polynomials of degree at most \(m\) and therefore cannot be dense in \(C(K)\).
\end{remark}



A final note is in order: A. Ronald Gallant and Halbert L. White \cite{gallant1988} provided an earlier proof for the approximation capabilities of artificial neural networks. They introduced a specific, non-periodic activation function (the sigmoid-like ``cosine squasher'') and proved that a feedforward network with a single hidden layer using this activation function could perfectly replicate a ``Fourier network,'' which yields a Fourier series approximation to a given function as its output. Therefore, an approximation to any desired accuracy can be achieved by a neural network, using sufficiently many hidden units. In this sense, neural networks do not make avoidable mistakes, hence the title of their ICNN paper.

In their 1992 follow-up paper \cite{gallant1992},  Gallant and White extended their 1988 proof: Neural networks don't just approximate the underlying target function; they also consistently estimate its partial derivatives (density in Sobolev spaces), a result also obtained by Hornik \cite{hornik1990}.
\subsection*{Hornik's Universal Approximation Theorems}

In a series of papers published in the {\it Neural Networks} journal, Kurt Hornik, Maxwell Stinchcombe, and the aforementioned Halbert White established that one-hidden-layer neural networks using arbitrary squashing functions 
(e.g., logistic, hyperbolic tangent) are capable of approximating any Bore1 measurable function  
to any desired degree of accuracy \cite{hornik1989}, as well as its derivatives \cite{hornik1990, hornik1991}.

Hornik, Stinchcombe, and White's proof of multilayer feedforward networks as a class of universal approximators is based on Gallant and White's ``cosine squasher'' result \cite{gallant1988} and resorts to the Stone-Weierstrass Theorem.

A family $A$ of real functions defined on a set $E$ is an algebra if $A$ is closed
under addition, multiplication, and scalar multiplication. A family $A$ separates points on $E$ if for every $x, y \in E$, $x \neq y$, there exists a function $f$ in $A$ such that $f(x) \neq f(y)$. The family A vanishes at no point of $E$ if for each $x \in E$ there exists $f \in A$ such that $f(x) \neq 0$.

\begin{theorem}[Stone-Weierstrass Theorem] \label{thm:stone-weierstrass}
Let $A$ be an algebra of real continuous functions on a compact set $K$. 
If $A$ separates points on $K$ and if $A$ vanishes at no point of $K$, then the uniform closure $B$ of $A$ consists of all real continuous functions on $K$, i.e., $A$ is $\mu_K$-dense in the space of real continuous functions on $K$.
\end{theorem}

\begin{definition} 
A \emph{squashing function} $\Psi(z): \mathbb R \to [0,1]$ is a monotonic (non-decreasing) function with $\lim_{z \to +\infty} \Psi(z) = 1$ and $\lim_{z \to -\infty} \Psi(z) = 0$
\end{definition} 

Squashing functions are measurable because their discontinuities are at most countable (finite or countable infinity), by Froda's Theorem, and the set of points where the function is discontinuous, being countable, has a Lebesgue measure zerothey have countable many discontinuities. A function that is continuous almost everywhere (i.e., continuous everywhere except on a set of measure zero like a countable set) is Borel measurable.

Examples of squashing functions include 
the threshold function $\Psi_1(z) = 1_{\{z \geq 0\}}$, where $1_{\{\cdot\}}$ is the indicator function,
the ramp function $\Psi_{ramp}(z) = z \cdot 1_{\{0 \leq z \leq 1\}} + 1_{\{z > 1\}}$, and Gallant and White's cosine squasher $\Psi_{cosine}(z) = (1+ \cos (z+3\pi/2)) \cdot 1_{\{-\pi/2 \leq z \leq \pi/2\}} + 1_{\{z > \pi/2 \}}$,

\begin{definition}[$\Sigma$  networks]
For any measurable function $G: \mathbb R \to \mathbb R$, $\Sigma^n (G)$ is the class of functions
$$
\left\{ f: \mathbb R^n \to \mathbb R: f(x)= \sum_{j=1}^q \beta_j G(A_{j}(x)) \right\}
$$
where $A_j(x)$ is an affine function of the form $A(x)=w\cdot x + b$.

\end{definition}
\begin{definition}[$\Sigma\Pi$  networks]
For any measurable function $G: \mathbb R \to \mathbb R$, $\Sigma \Pi^n (G)$ is the class of functions
$$
\left\{ f: \mathbb R^n \to \mathbb R: f(x)=\sum_{j=1}^q \beta_j \prod_{k=1}^{l_j} G(A_{jk}(x)) \right\}
$$
where $A_{jk}(x)$ is an affine function of the form $A(x)=w\cdot x + b$.

$\Sigma$ networks are a special case of $\Sigma\Pi$ networks, with $l_j=1$ for all $j$.
\end{definition}

The classes $\Sigma^n(G)$ and $\Sigma\Pi^n(G)$ belong to the set of Borel measurable functions from $\mathbb R^n$ to $\mathbb R$ for any Borel measurable $G$, i.e. $\Sigma^n(G) \subset \Sigma\Pi^n(G) \subset M(\mathbb R^n)$. When $G$ is continuous, $\Sigma^n(G)$ and $\Sigma\Pi^n(G)$ belong to the set of continuous functions from $\mathbb R^n$ to $\mathbb R$, i.e. $\Sigma^n(G) \subset \Sigma\Pi^n(G) \subset C(\mathbb R^n)$.

Closeness of functions $f$ and $g$ belonging to $C(\mathbb R^n)$ or $M(\mathbb R^n)$
is measured by a metric $\mu$. Closeness of one class of functions to another class is described by the concept of denseness:

\begin{definition}[Dense.]
A subset $S$ of a metric space $(X,\mu)$ is $\mu$-dense in a subset $T$ when, for every $\epsilon >0$ and every $t \in T$, there is an $s \in S$ such that $\mu(s,t)<\epsilon$. 

An element of $S$ can approximate an element of $T$ to any desired degree of accuracy
\end{definition}

\begin{definition}[Uniformly Dense.]
A subset $S$ of $C(\mathbb R^n)$ is said to be uniformly dense on
compacta in $C(\mathbb R^n)$ if, for every compact subset $K \in \mathbb R^n$,
$S$ is $\mu$-dense in $C(\mathbb R^n)$ where, for $f,g \in C(\mathbb R^n)$, $\mu_K(f,g) = \sup_{x\in K}|f(x)-g(x)|$.

A sequence of functions $\{f_n\}$ converges to a function $f$ uniformly on compacta if, for all compact $K \in \mathbb R^n$, $\mu_K(f_n,f) \to 0$ as $n \to \infty$.
\end{definition}

\begin{lemma}[$\Sigma\Pi$ is uniformly dense.] 
\label{lemma:hornik-sigma-pi-continuous}

$\Sigma\Pi$  feedforward networks are capable of arbitrarily accurate approximation to any real-valued continuous function over a compact set: $\Sigma\Pi^n(G)$ is uniformly dense on compacta in $C(\mathbb R^n)$.
\end{lemma}

\begin{proof}
    By applying Stone-Weierstrass Theorem (Theorem \ref{thm:stone-weierstrass}), for any compact set $K \subset \mathbb R^n$ and any $G$, $\Sigma\Pi^n(G)$ is an algebra on $K$.

    \begin{itemize}

    \item
    If $x,y \in K$, $x \neq y$, there is an affine function $A$ such that $G(A(x)) \neq G(A(y))$. Just pick $a,b \in \mathbb R$, $a \neq b$, such that $G(a)\neq G(b)$, and pick $A$ to satisfy $A(x)=a$ and $A(y)=b$. Then $G(A(x)) \neq G(A(y))$, so $\Sigma\Pi^n(G)$ is separating on $K$.

    \item
    There are $G(A(x))$ that are constant and not equal to zero. Just pick $b\in \mathbb R$ such that $G(b)\neq 0$ and set $A(x) = 0 \cdot x + b$. For all $x \in K$, $G(A(x))=G(b)\neq 0$, so $\Sigma\Pi^n(G)$ vanishes at no point of $K$.
    
    \end{itemize}

    So the Stone-Weierstrass Theorem implies that $\Sigma\Pi^n(G)$ is $\mu_K$-dense in the space of real continuous functions on $K$ for any compact set $K$.
\end{proof}

\begin{definition}[Probability Measure.]
Let $\rho$ be a probability measure on $(\mathbb R ^n, \mathbb B^n)$, where $\mathbb B^n$ is the Borel $\sigma$-field of $\mathbb R ^n$. The measure describes the relative frequency of occurrence of input patterns $x$.

If $f, g \in M(\mathbb R^n)$, we say they are $\rho$-equivalent if $\rho \{ x \in \mathbb R^n: f(x)=g(x)\} = 1$. Functions that are $\rho$--equivalent differ only
on a set of patterns occurring with probability (measure) zero.

Given a probability measure $\rho$ on $(\mathbb R ^n, \mathbb B^n)$, we define the metric $\mu_\rho: M(\mathbb R^n) \times M(\mathbb R^n) \to \mathbb R^+$:
$$
\mu_\rho(f,g) = \inf \{ \epsilon>0: \rho \{ x: |f(x)-g(x)| > \epsilon\} < \epsilon \}
$$
\end{definition}

Two functions are close in this metric if and only if there is only a small probability that they differ significantly. In the extreme case, when $f$ and $g$ are
$\rho$-equivalent, $\mu_\rho(f,g)=0$.

There are different ways to describe that $\mu_\rho(f_n,f)$ converges to 0:
\begin{itemize}
    \item
    $\mu_\rho(f_n,f) \to 0$.
    \item
    For every $\epsilon>0$, $\rho \{ x: |f_n(x)-f(x)| > \epsilon\}\to 0$
    \item
    $\int \min \{ |f_n(x)-f(x)|, 1\} \rho(dx) \to 0$
\end{itemize}

\begin{lemma}[Uniform convergence on compacta and $\mu_\rho$-convergence.]
\label{lemma:hornik-uniform-mu-convergence}

If $\{f_n\}$ is a sequence of functions in $M(\mathbb R^n)$ that converges uniformly on compacta to the function $f$, then $\mu_\rho(f_n,f) \to 0$.
\end{lemma}

\begin{proof}
    Let $\epsilon>0$ and suppose $\rho(\mathbb R^n)=1$ for convenience. Since $\mathbb R^n$ is a locally compact metric space, $\rho$ is a regular measure. Therefore, there is a compact subset $K \in \mathbb R^n$ with $\rho(K)>1-\epsilon/2$.

    It is sufficient to find $N \in \mathbb N$ such that, $\forall n \geq N$,  $\int \min \{ |f_n(x)-f(x)|, 1\} \rho(dx) < \epsilon$. Picking $N$ such that, $\forall n \geq N$, $\sup_{x \in K} | fn(x)-f(n)| < \epsilon/2$, we have
    $ \int_{\mathbb R^n-K} \min \{ |f_n(x)-f(x)|, 1\} \rho(dx) 
    + \int_{K} \min \{ |f_n(x)-f(x)|, 1\} \rho(dx) 
    < \epsilon/2 + \epsilon/2 = \epsilon$
\end{proof}

\begin{lemma}[Finite measures are $\mu_\rho$-dense.]
\label{lemma:hornik-c-dense-in-m}

For any finite measure $\rho$, $C(\mathbb R^n)$ is $\mu_\rho$-dense in $M(\mathbb R^n)$.
\end{lemma}

\begin{proof}
    Given $\epsilon>0$ and an arbitrary $f \in M(\mathbb R^n)$, we must find $g\in C(\mathbb R^n)$ such that $\mu_\rho(f,g)<\epsilon$.

    For sufficiently large $M$, 
    $$\int \min \{|f \cdot 1_{\{|f|<M\}} - f|,1\} d\rho < \epsilon/2$$
    In fact, measure theory textbooks show that there exists a continuous function $g$ such that 
    $$\int \min \{|f \cdot 1_{\{|f|<M\}} - g|,1\} d\rho < \epsilon/2$$
    Therefore,
    $$\int \min \{|f - g|,1\} d\rho < \epsilon$$
\end{proof}

\begin{lemma}[$\Sigma\Pi$ is $\mu$-dense.] 
\label{lemma:hornik-sigma-pi-measurable}

For every $G$ and every probability measure $\rho$ on $(\mathbb R ^n, \mathbb B^n)$, $\Sigma\Pi^n(G)$ is $\mu_\rho$-dense in $M(\mathbb R^n)$.
\end{lemma}

\begin{proof}
    Given any continuous nonconstant function, it follows  from Lemmas \ref{lemma:hornik-sigma-pi-continuous} ($\Sigma\Pi$ is uniformly dense) and \ref{lemma:hornik-uniform-mu-convergence} (connecting uniform convergence to $\mu_\rho$-convergence) that $\Sigma\Pi^n(G)$ is $\mu_\rho$-dense in $C(\mathbb R^n).$

    Since $C(\mathbb R^n)$ is dense in $M(\mathbb R^n)$ by Lemma \ref{lemma:hornik-c-dense-in-m}, it follows that $\Sigma\Pi^n(G)$ is $\mu_\rho$-dense in $M(\mathbb R^n)$ by the triangle inequality.
\end{proof}

\begin{lemma}[Continuous squashing functions.] 
\label{lemma:hornik-squashing-function}

Let $F$ be a continuous squashing function and $\Psi$ an arbitrary squashing function. For every $\epsilon>0$, there is an element $H_\epsilon \in \Sigma^1(\Psi)$ such that $\sup_{\lambda \in \mathbb R} | F(\lambda) - H_\epsilon(\lambda)| < \epsilon$.
\end{lemma}

\begin{proof}
For an arbitrary $\epsilon >0$, we must find a collection of constants $\beta_j$ and affine functions $A_j$ such that $\sup_{\lambda \in \mathbb R} | F(\lambda) - \sum_{j=1}^{Q-1} \beta_j \Psi(A_j(\lambda))| < \epsilon$.

Pick $Q$ such that $1/Q < \epsilon/2$ and set $\beta_j = 1/Q$ for $j \in \{1..Q-1\}$. 

Pick $M>0$ such that $\Psi(-M)< \epsilon/2Q$ and $\Psi(M) > 1 - \epsilon/2Q$. Since $\Psi$ is a squashing function, such an $M$ exists.

Set $r_j = \sup \{ \lambda: F(\lambda)=j/Q\}$ for $j \in \{1..Q-1\}$ and $r_Q = \sup \{ \lambda: F(\lambda)=1-1/2Q\}$. Since $F$ is a continuous squashing function, such $r_j$'s exist.

For any $r<s$, let $A_{r,s}$ the unique affine function satisfying $A_{r,s}(r)=M$ and $A_{r,s}(s)=-M$.

The desired approximation is 
$$H_\epsilon(\lambda)= \sum_{j=1}^{Q-1} \beta_j \Psi(A_{r_j, r_{j+1}} (\lambda))$$
For every interval $(-\infty, r_1], (r_1, r_2]..(r_{Q-1}, r_Q], (r_Q, \infty)$, $$|F(\lambda)-H_\epsilon(\lambda)|<\epsilon$$
\end{proof}

\begin{lemma}[$\Sigma\Pi$ for squashing functions.] 
\label{lemma:hornik-sigma-pi-squashing}

For every squashing function $\Psi$ and and every probability measure $\rho$ on $(\mathbb R ^n, \mathbb B^n)$, $\Sigma^n(\Psi)$ is uniformly dense on compacta in $C(\mathbb R^n)$ and $\mu_\rho$-dense in $M(\mathbb R^n)$.
\end{lemma}

\begin{proof}
    By Lemmas \ref{lemma:hornik-uniform-mu-convergence} (connecting uniform convergence to $\mu_\rho$-convergence) and \ref{lemma:hornik-sigma-pi-measurable} ($\Sigma\Pi^n(G)$ is $\mu_\rho$-dense in $M(\mathbb R^n)$), it is sufficient to show that $\Sigma\Pi^n(\Psi)$ is uniformly dense on compacta in $\Sigma\Pi^n(F)$ for some continuous squashing function $F$.

    For this, it is sufficient to show that every function of the form $\prod_{k=1}^rF(A_k(x))$ can be uniformly approximated by members of $\Sigma\Pi^n(\Psi)$. 

    For an arbitrary $\epsilon >0$, because multiplication is continuous and $[0,1)^r$ is compact, there is a $\delta>0$ such that $|a_k-b_k|<\delta$ for $0 \leq a_k, b_k \leq 1$ for $k \in \{ 1..r\}$. Therefore, $|\prod_{k=1}^r a_k - \prod_{k=1}^r b_k| < \epsilon$. 

    By Lemma \ref{lemma:hornik-squashing-function}, there is a function
    $H_\delta(\lambda) = \sum_{t=1}^{T} \beta_t \Psi(A_{t}^r (\lambda))$ such that
    $\sup_{\lambda \in \mathbb R}|F(\lambda)-H_\delta(\lambda)|< \epsilon$. Then,
    $$
    \sup_{\lambda \in \mathbb R^n} \left| \prod_{k=1}^r F(A_k(x)) - \prod_{k=1}^r H_\delta(A_k(x)) \right|< \epsilon
    $$
    Since $A_t^r(A(x))$ is an affine function $\prod_{k=1}^r H_\delta(A_k(x)) \in \Sigma\Pi^n(\Psi)$. Therefore, $\prod_{k=1}^r F(A_k(x))$ can be uniformly approximated by elements of $\Sigma\Pi^n(\Psi)$.
\end{proof}

\begin{lemma}
\label{lemma:hornik-sigma}

For every squashing function $\Psi$ and and every probability measure $\rho$ on $(\mathbb R ^n, \mathbb B^n)$, $\Sigma^n(\Psi)$ is uniformly dense on compacta in $C(\mathbb R^n)$ and $\mu_\rho$-dense in $M(\mathbb R^n)$. 
\end{lemma}

\begin{proof}

    For every squashing function $\Psi$, $\epsilon>0$, and $M>0$, there is a function $\cos_M \in \Sigma^1(\Psi)$ such that $\sup_{\lambda \in [-M,+M]} |\cos_M(\lambda)-\cos(\lambda)|<\epsilon$.

    \begin{quote}
    Just use Gallant and White's cosine squasher \cite{gallant1988}: $\Psi_{cosine}(z) = (1+ \cos (z+3\pi/2)) \cdot 1_{\{-\pi/2 \leq z \leq \pi/2\}} + 1_{\{z > \pi/2 \}}$. By adding, subtracting and scaling a finite number of affinely shifted versions of $\Psi_{cosine}$, we can get the cosine function on any interval $[-M, +M]$. Lemma \ref{lemma:hornik-squashing-function} and the triangle inequality provide the needed $\cos_M$ function.
    \end{quote}

    Let $g(x) = \sum_{j=1}^Q \beta_j \cos(A_j(x))$, where $A_j$ is an affine function. For arbitrary squashing function $\Psi$, compact $K \subset \mathbb R^n$, and $\epsilon>0$, there is an $f \in \Sigma^n (\Psi)$ such that $\sup_{x \in K} |g(x)-f(x)| < \epsilon$.

    \begin{quote}
    Pick $M>0$ such that $A_j(K) \subset[-M,+M]$ for $j \in \{1..Q\}$. Since $Q$ is finite, $K$ is compact, and the affine functions $A_j$ are continuous, such an $M$ can be found. Let $Q' = Q \cdot \sum_{j=1}^{Q} |\beta_j|$. By the result in the previous paragraph, $| \sum_{j=1}^{Q} \beta_j \cos_M(A_j(x))-g(x)|< \epsilon$ for all $x\in K$. Since $\cos_M \in \Sigma^1(\Psi)$, $f(x)=\sum_{j=1}^{Q} \cos_M(A_j(x)) \in \Sigma^n(\Psi)$
    \end{quote}
    
    For every squashing function $\Psi$, $\Sigma^n(\Psi)$ is uniformly dense on compacta in $C(\mathbb R^n)$: 

    \begin{quote}
    By Theorem \ref{lemma:hornik-sigma-pi-continuous}, $\Sigma\Pi$ is uniformly dense. Therefore, the trigonometric polynomials $\{ \sum_{j=1}^Q \beta_j \prod_{k=1}^{r_j} \cos(A_{jk}(x))\}$ are uniformly dense on $C(\mathbb R^n)$, where $A_{jk}$ are affine functions. Repeatedly applying the trigonometric identity $\cos(a) \cos(b) = \cos(a + b) - \cos(a - b)$ allows us to rewrite every trigonometric polynomial in the form $\sum_{t=1}^T \alpha_t \cos(A_t(x))$ where $\alpha_t \in \mathbb R$ and $A_t$ is an affine function. The density result now follows from the approximation above. 
    \end{quote}
    
    As we have just seen, $\Sigma^n(\Psi)$ is uniformly dense on compacta in $C(\mathbb R^n)$. By Lemma \ref{lemma:hornik-uniform-mu-convergence} (connecting uniform convergence to $\mu_\rho$-convergence), $\Sigma^n(\Psi)$ is $\mu_\rho$-dense in $C(\mathbb R^n)$.

    The triangle inequality and Lemma \ref{lemma:hornik-c-dense-in-m} imply that $\Sigma^n(\Psi)$ is also $\mu_\rho$-dense in $M(\mathbb R^n)$.
\end{proof}

\repthm{thm:hornik}

\begin{proof}
Hornik et al. \cite{hornik1989} proved their result for $\Sigma\Pi$ networks and later restricted it to the particular case of $\Sigma$ networks. They first proved that networks can approximate continuous function, and then extended their result to Borel measurable functions. Let $G(z): \mathbb R \to \mathbb R$ be any continuous nonconstant function: 

\begin{itemize}
\item
$\Sigma\Pi^n(G)$ is uniformly dense on compacta in $C(\mathbb R^n)$ (Lemma \ref{lemma:hornik-sigma-pi-continuous}).

\item
For every $G$ and every probability measure $\rho$ on $(\mathbb R ^n, \mathbb B^n)$, $\Sigma\Pi^n(G)$ is $\mu_\rho$-dense in $M(\mathbb R^n)$ (Lemma \ref{lemma:hornik-sigma-pi-measurable}).
\end{itemize}

The activation function $G$ is not required to be a squashing function. The continuity requirement discards the threshold function $\Psi_1(z) = 1_{\{z \geq 0\}}$.

However, for squashing functions continuity is not necessary.
For every squashing function $\Psi$ and and every probability measure $\rho$ on $(\mathbb R ^n, \mathbb B^n)$, $\Sigma\Pi^n(\Psi)$ is uniformly dense on compacta in $C(\mathbb R^n)$ and $\mu_\rho$-dense in $M(\mathbb R^n)$ (Lemma \ref{lemma:hornik-sigma-pi-squashing}). 

Because of their simpler structure, $\Sigma$ networks have similar approximation capabilities (Lemma \ref{lemma:hornik-sigma}): For every squashing function $\Psi$ and and every probability measure $\rho$ on $(\mathbb R ^n, \mathbb B^n)$, $\Sigma^n(\Psi)$ is uniformly dense on compacta in $C(\mathbb R^n)$ and $\mu_\rho$-dense in $M(\mathbb R^n)$. 

In other words, standard feedforward networks with a single hidden layer can approximate any continuous function uniformly on any compact set and any measurable function arbitrarily well, regardless of the squashing function $\Psi$ (continuous or not). Therefore, $\Sigma$ networks are universal approximators.
\end{proof}

\begin{remark}
Hornik and his colleagues extended Funahashi's theorem for non-continuous activation functions.
\end{remark}

Neural networks can approximate functions that are not differentiable in the classical sense, but possess only a generalized derivative, as is the case for certain piecewise differentiable functions \cite{hornik1990}. Hornik established the general conditions ensuring that networks with sufficiently smooth activation functions are capable of arbitrarily accurate approximation to a function and its derivatives \cite{hornik1991}.\footnote{Why this matters in practice: In Economics, for instance, the derivative of a function represents crucial concepts like marginal utility, elasticities, and marginal rates of substitution. If artificial neural networks also learn derivatives, then they can be used to forecast elasticities and economic reactions accurately, not just raw output values. Approximation of derivatives also permit sensitivity analyses, where the relative effects on outputs of small changes in input variables can be studied.}

For broad classes of multilayer feedforward networks, the set $\Sigma(G)$ is dense in general spaces of functions where distance between functions is measured taking into account differences between the derivatives of the functions \cite{hornik1990}.
In fact, whenever $\Psi$ is continuous, bounded and nonconstant, standard multilayer feedforward networks with activation function $\Psi$ can approximate any continuous function on compact subsets $X \subset \mathbb R^n$ arbitrarily well with respect to uniform distance $\rho_{\mu,x}$ provided that sufficiently many hidden units are available \cite{hornik1991}.

Whereas \cite{hornik1990} makes use of a Fourier integral representation
for single hidden layer feedforward networks with a continuum of hidden units, as proposed by \cite{irie1988}, \cite{hornik1991} improves their results following Cybenko's original approach \cite{cybenko1989}: the application of the Hahn-Banach theorem (Theorem \ref{thm:hahn-banach}) combined with representation theorems (Theorem \ref{thm:riesz}) for continuous linear functionals on the function spaces under consideration.

The set of all functions implemented by a one-hidden-layer neural network with $h$ hidden units and a single output unit, which is assumed to be linear, is defined as
$$
\mathfrak{N}_n^{(h)}[\Psi] 
  = \left\{ f:\mathbb R^n \to \mathbb R | f(x)=\sum_{j=1}^h \beta_j \Psi(w_j \cdot x + b_j) \right\}
$$

The set of all functions implemented by such networks with an arbitrary number of hidden units is
$$
\mathfrak{N}_n [\Psi] = \bigcup_{h=1}^\infty \mathfrak{N}_n^{(h)}[\Psi] 
$$

For $1 \leq p < \infty$, the $L^p$ norm is defined as 
$$
||f||_{p,\mu} = \left[ \int_{\mathbb R^n} |f(x)|^p d\mu (x)\right]^{1/p}
$$

The $L^p$ distance (or metric) defines the distance between two functions:
$$
\rho_{p,\mu} (f,g) = || f-g||_{p,\mu} 
$$

$L^p(\mu)$ is the space of all functions such that $||f||_{p,\mu} < \infty$, i.e. is the set of all measurable functions whose total size (measured by the \(L^{p}\) norm) is a finite number. For a function to belong to \(L^p(\mu)\), its \(p\)-th power must be integrable over the entire domain, which forces the function to be well-behaved in two main ways: controlled spikes (even with vertical asymptotes, it cannot shoot up to infinity too sharply) and fast decay at infinity (must shrink toward zero quickly enough).

A subset $S \subset L^p(\mu)$ is dense in $L^p(\mu)$ if, for arbitrary $f \in L^p(\mu)$
and $\epsilon > 0$, there is a function $g \in S$  such that $\rho_{p,\mu} (f,g) < \epsilon$.

\begin{theorem}
\label{thm:hornik-nn-finite-measures}

If $\Psi$ is bounded and nonconstant, then $\mathfrak{N}_n [\Psi]$ is dense in $L^p(\mu)$ for all finite measures $\mu$ on $\mathbb R^n$.
\end{theorem}

\begin{corollary}
    As convergence in $L^p(\mu)$ implies convergence in $\mu$ measure, whenever $\Psi$ is bounded and nonconstant, all measurable functions on $\mathbb R^n$ can be approximated by functions in $\mathfrak{N}_n [\Psi]$. 
\end{corollary}

$C(X)$ is the space of all continuous functions on $X$. A subset $S \subset C(X)$ is dense in $C(X)$ if, for arbitrary $f \in C(X)$
and $\epsilon > 0$, there is a function $g \in S$  such that $\rho_{p,X} (f,g) < \epsilon$.

\begin{theorem}
\label{thm:hornik-nn-compact}

If $\Psi$ is continuous, bounded, and nonconstant, then $\mathfrak{N}_n [\Psi]$ is dense in $C(X)$ for all compact subsets $X \subset \mathbb R^n$.
\end{theorem}

\begin{proof}[Proof of Theorems \ref{thm:hornik-nn-finite-measures} and \ref{thm:hornik-nn-compact}]

Since $\Psi$ is bounded, $\mathfrak{N}_n [\Psi]$ is a linear subspace of $L^p(\mu)$ for all finite measures $\mu$ on $\mathbb R^n$. If, for some measure $\mu$, $\mathfrak{N}_n [\Psi]$ is not dense in $L^p(\mu)$, there is a nonzero continuous linear functional $\Lambda$ on $L^p(\mu)$ that vanishes on $\mathfrak{N}_n [\Psi]$. 

$\Lambda$ is of the form $\Lambda (f) = \int_{\mathbb R^n} f g \, d\mu$ for some $g \in L^q(\mu)$ where $q$ is the conjugate exponent $q=p/(p-1)$.\footnote{For $p=1$, $q\to \infty$. $L^\infty(\mu)$ is the space of all functions $f$ for which the $\mu$ essential supremum is finite ($||f||_{\infty,\mu}$), i.e. the space of $\mu$ essentially bounded functions.}

By Hölder's inequality, $ \| fg \|_1 \le \| f \|_p \| g \|_p$. Given $\sigma(B)=\int_B g \, d\mu$:
\begin{align*}
|\sigma(B)|  &= \left| \int_{\mathbb R^n } \mathbf 1_B g \, d\mu \right| \\
             &\le \| 1_B \|_{p,\mu} \| g \|_{q,\mu} \\
             &\le \sqrt[p]{\mu(\mathbb R^n)} \, \| g \|_{q,\mu} < \infty
\end{align*}
Therefore, $\sigma$ is a nonzero finite signed measure on $\mathbb R^n$ such that $\Lambda (f) = \int_{\mathbb R^n} f g \, d\mu = \int_{\mathbb R^n} f \, d\sigma$. Since $\Lambda$ vanishes on $\mathfrak{N}_n [\Psi]$, for all $w\in \mathbb R^n$ and $b \in \mathbb R$:
$$
\int_{\mathbb R^n} \Psi(wx+b) \, d\sigma(x) = 0
$$

Similarly, suppose that $\Psi$ is continuous and, for some compact subset $X$ of $R^n$, $\mathfrak{N}_n [\Psi]$ is not dense in $C(X)$. Following Cybenko's proof (Theorem \ref{thm:cybenko}), there exists a nonzero finite signed measure $\sigma$ on $\mathbb R^n$ ($\sigma$ is actually concentrated on $X$) such that, for all $w\in \mathbb R^n$ and $b \in \mathbb R$:
$$
\int_{\mathbb R^n} \Psi(wx+b) \, d\sigma(x) = 0
$$

In both cases, we reduce our problem to answer whether there exists a nonzero finite signed measure $\sigma$, i.e. whether $\Psi$ is discriminatory or not (Definition \ref{def:function-discriminatory}). Cybenko \cite{cybenko1989} proved that, when $\Psi$ is sigmoidal, $\Psi$ is discriminatory. Hornik \cite{hornik1991} proved a stronger result: whenever $\Psi$ is bounded and nonconstant, $\Psi$ is discriminatory (see Theorem \ref{thm:hornik-discriminatory} below). Combining Hornik's result with our argument, the proof of Theorems \ref{thm:hornik-nn-finite-measures} and \ref{thm:hornik-nn-compact} is complete.
\end{proof}

\begin{theorem}
\label{thm:hornik-discriminatory}
Whenever $\Psi$ is bounded and nonconstant, $\Psi$ is discriminatory.
\end{theorem}

\begin{proof}
Hornik's proof \cite{hornik1991} is based on Fourier analysis. Suppose that $\Psi$ is bounded and nonconstant. Suppose that $\sigma$ is a finite signed measure $\sigma$ on $\mathbb R^n$ such that $\int_{\mathbb R^n} \Psi(wx+b) \, d\sigma(x) = 0$ for all $w\in \mathbb R^n$ and $b \in \mathbb R$.

Set $u\in \mathbb R^n$ and define a finite signed measure $\sigma_u$ on $\mathbb R$ induced by the transformation $x \mapsto u \cdot x$. For all Borel sets, we have $\sigma_u(B)= \sigma\{ x \in \mathbb R^n: u \cdot x \in B \}$.

For all bounded functions $\chi$ on $\mathbb R$:
$$
\int_{\mathbb R^n} \chi(u x) \, d\sigma(x) = \int_{\mathbb R} \chi(t) \, d\sigma_u(t)
$$

Hence, for all $w,b \in \mathbb R$:
$$
\int_{\mathbb R^n} \Psi(wu x+b) \, d\sigma(x) = \int_{\mathbb R} \Psi(wt+b) \, d\sigma_u(t) = 0
$$

By choosing $b$ such that $\Psi(b)\ne 0$ and $w=0$, we find that $\int_{\mathbb R} d\sigma_u(t) = \widehat \sigma_u (0) = 0$, where $\widehat \sigma_u$ is the Fourier transform of $\sigma_u$. 

For $u=0$, $\sigma_0$ is concentrated at $t=0$ and $\sigma_0 \{ 0 \} = \widehat \sigma_0 = 0$. Hence $\sigma_0=0$.

Now suppose $u \ne 0$. Choose an integrable function $f \in L^1(\mathbb R)$ whose Fourier transform has no zero, e.g. $f(t) = e^{-t^2}$. 
As, by Hölder's inequality,
\begin{align*}
\int_{\mathbb R} \int_{\mathbb R} & |\Psi(w(s+t)+b)| \, |f(s)| \, ds \, d|\sigma_u(t)|\\
& \le \|f\|_L \, \|\sigma_u\|_M \, \sup_{t\in \mathbb R} \|\Psi(t) \| < \infty
\end{align*}
we can apply Fubini's theorem:
\begin{align*}
0 &= \int_{\mathbb R} \left[ \int_{\mathbb R} \Psi(wt+(b+ws)) \, d\sigma_u(t) \right] f(s) \, ds \\
&= \int_{\mathbb R} \int_{\mathbb R} \Psi(w(s+t)+b) \, f(s) \, ds d\sigma_u(t)\\
&= \int_{\mathbb R} \Psi(wt+b) \, d(f *\sigma_u)(t)\\
\end{align*}
where $f *\sigma_u$ denotes the convolution of $f$ and $\sigma_u$.
Since $L^1(\mathbb R)$ is a closed ideal in $M(\mathbb R)$, for $f \in L^1(\mathbb{R})$ and $\mu \in M(\mathbb{R})$, the convolution $f * \mu = \mu * f \in L^1(\mathbb{R})$.
In particular, $f * \sigma_u$ is absolutely continuous with respect to Lebesgue measure. Let $h \in L^1(\mathbb{R})$ the corresponding Radon-Nikodym derivative.
Then, $\widehat h = \widehat f \cdot \widehat \sigma_u$ and $\widehat h (0) = 0$.
The above equation is, therefore, equivalent to 
$$
\int_{\mathbb R} \Psi(wt+b) \, h(t) \, dt = 0
$$

Let $\alpha \ne 0$ and $\gamma \in \mathbb R$. By replacing $w$ by $1/\alpha$ and $b$ by $\gamma/\alpha$ and performing the change of variables $t \mapsto \alpha t - \gamma$:
$$
\int_{\mathbb R} \Psi(t) \, h(\alpha t-\gamma) \, dt = 0
$$
This equation implies that $\int_{\mathbb R} \Psi(t) f(t) dt$ vanishes for all $f$ in the closed translation subspace $I$ spanned by the family $M_\alpha h$, where $M_\alpha h(t) = h(\alpha t)$, for $\alpha \ne 0$. As a consequence, $I$ is an ideal in $L^1(\mathbb R)$.

Define $Z(f)$ as the set of all $\omega \in \mathbb R$ where the Fourier transform $\widehat f(\omega)$ vanishes. When $I$ is an ideal, $Z(I)$, the zero set of $I$, is the set of $\omega$ where the Fourier transforms of all functions in $I$ vanish.

Suppose that $h$ is nonzero. As $M_\alpha h(\omega) = \widehat h(\omega/\alpha)\alpha$, we find that $Z(I) = \{0\}$ and $I$ is the set of all integrable functions $f$ with $\int_{\mathbb R} f(t) dt = \widehat f(0) = 0$.
\begin{quote}
For all functions $f \in I$, we trivially have $\{0\} = Z(I) \subseteq Z(f)$. Conversely, suppose that $f$ has zero integral. As the intersection of the boundaries of $Z(I)$ and $Z(f)$ equals $\{0\}$ and contains no perfect set, $f \in I$.
\end{quote}

When $h$ is nonzero, the fact that the integral $\int_{\mathbb R} \Psi(t)f(t) dt$ vanishes for all integrable functions which have zero integral implies that $\Psi$ is
constant, which was ruled out by our initial assumption. Hence $h = 0$ and $\widehat h = \widehat f \cdot \widehat \sigma_u$ is identically zero. In turn, $\widehat \sigma_u$ vanishes identically, because $\widehat f$. Therefore, $\sigma_u = 0$ for all $u \in \mathbb R^n$.

Then, the Fourier transform of $\sigma$ at $u$ is
\begin{align*}
\widehat \sigma (u) &= \int_{\mathbb R^n}e ^{i u x}d\sigma(x) \\
                    &= \int_{\mathbb R}e ^{i t} d\sigma_u(x) \\
                    &= 0
\end{align*}
Since $\widehat \sigma=0$, then $\sigma=0$ and our proof is complete: $\Psi$ is discriminatory, i.e. no nonzero finite signed measure $\sigma$ exists such that $\int \Psi(wx+b) \, d\sigma(x)=0$.
\end{proof}

A multi-index is a k-tuple $\alpha = (\alpha_1, ..,\alpha_k)$ of non-negative integers. The order of the multi-index is $|\alpha| = \sum_{i=1}^k \alpha_i$ and its corresponding partial derivative is
$$
D^\alpha f(x) = \frac{ \partial^{\alpha_1..\alpha_k} f}{\delta \xi_1^{\alpha_1} .. \partial \xi_k^{\alpha_k} }(x)
$$
for a sufficiently smooth function $f$ of $x=(\xi_1, ..,\xi_k) \in \mathbb R^k$.

$C^m(\mathbb R^k)$ is the space of all functions $f$ which, with all their partial derivatives $D^\alpha f$ of order $|\alpha| \geq m$, are continuous on $\mathbb R^k$.

For all subsets $X \subset \mathbb R^k$ and $f \in C^m(\mathbb R^k)$:
$$
||f||_{m,\mu,X} = \max_{|\alpha| \leq m} \sup_{x \in X} | D^\alpha f(x)|
$$

A subset $S$ of $C^m(\mathbb R^k)$ is uniformly dense on compacta in $C^m(\mathbb R^k)$ if, for all compact subsets $X$ of $\mathbb R^k$ and all $\epsilon > 0$, there is a function $g(f,X,\epsilon) \in S$  such that $|| f-g||_{m,\mu,X} < \epsilon$.

For $f \in C^m(\mathbb R^k)$, $\mu$ a finite measure on $\mathbb R^k$, and $1 \leq p < \infty$, 
$$
||f||_{m, p,\mu} = \left[ \sum_{|\alpha| \leq m} \int_{\mathbb R^k} |D^\alpha f|^p d\mu \right]^{1/p}
$$

The weighted Sobolev space $C^{m,p}(\mu)$ is defined by
$$
C^{m,p}(\mu) = \left\{ f \in C^m(\mathbb R^k): ||f||_{m, p,\mu} < \infty \right\}
$$

It should be noted that $C^{m,p}(\mu) = C^m(\mathbb R^k)$ if $\mu$ has compact support.

A subset $S$ of $C^{m,p}(\mu)$ is dense in $C^{m,p}(\mu)$ if, for all $f \in C^{m,p}(\mu)$ and $\epsilon > 0$, there is a function $g(f,\epsilon) \in S$ such that $|| f-g||_{m,p,\mu} < \epsilon$.

\begin{lemma}
Let $w$ be the familiar function in $C^\infty(\mathbb R^n)$ with support in the unit sphere given by
$$
w(x) = \begin{cases}
  c \, e^{-1/(1-|x|^2)} & \text{if } |x|<1 \\
  0  & \text{if } |x| \ge 1
\end{cases}
$$
where $|x|$ is the Euclidean length of $x$ and $c$ a constant chosen so that $\int_{\mathbb R^n} w(x)dx=1$.
For $\epsilon>0$,
$$
w_\epsilon (x) = \epsilon^n \, w(x/\epsilon)
$$

Suppose that $f$ and $\sigma$ satisfy one of the two following conditions:
\begin{itemize}
    \item 
    $f$ is continuous and $\sigma$ is a finite signed measure with compact support.
    \item 
    $f$ is bounded and continuous and $\sigma$ is a finite signed measure.
\end{itemize}
If $T_y f(x) = f(x+y)$ denotes translation by $y$:

\begin{align*}
\int_{\mathbb R^n} f \mathbf J_\epsilon \sigma \, dx
&= \int_{\mathbb R^n} \left[ \int_{\mathbb R^n} T_yf \, d\sigma \right] w_\epsilon(y) \, dy \\
&= \int_{\mathbb R^n} J_\epsilon f d\sigma
\end{align*}

$$
\lim_{\epsilon \to 0} \int_{\mathbb R^n} f \mathbf J_\epsilon \sigma \, dx
= \int_{\mathbb R^n} f \, d\sigma 
$$
where, for a locally finite signed measure $\sigma$ on $\mathbb R^n$, $J_\epsilon \sigma$ is the convolution $w_\epsilon * \sigma$.
\end{lemma}

\begin{proof}
The integral equation is a straightforward application of Fubini's theorem using the symmetry of $w_\epsilon$. The limit follows from Lebesgue's bounded convergence theorem (Theorem \ref{thm:lebesgue}).
\end{proof}

\begin{remark}
    If $f$ is is a locally integrable function on $\mathbb R^n$, let $J_\epsilon f$ be the convolution $w_\epsilon * f$:

    \begin{itemize}
    \item 
    $J_\epsilon f = w_\epsilon * f \in C^\infty (\mathbb R^n)$.
    \item
    $\| J_\epsilon f \|_u \le \| f \|_u$, where $\| f \|_u = \sup_{\mathbb R^n} |f(x)|$.
    \item
    If $f$ is bounded, then $J_\epsilon f$ is uniformly bounded in $x$ and $\epsilon$. 
    \item
    If $\sigma$ is a locally finite signed measure, $J_\epsilon \sigma = w_\epsilon * \sigma \in C^\infty (\mathbb R^n)$.
    \item
    If $\sigma$ has compact support, $J_\epsilon \sigma$ has compact support.
    \end{itemize}
\end{remark}
\begin{theorem}
\label{thm:hornik-cm-density}

If $\Psi \in C^m(\mathbb R^n)$ is nonconstant and bounded, then $\mathfrak{N}_n [\Psi]$ is uniformly m-dense on compacta in $C^{m,p}(\mu)$ for all finite measures $\mu$ on $\mathbb R^n$ with compact support.
\end{theorem}

\begin{proof}
If $\mathfrak{N}_n [\Psi]$ is not uniformly m-dense on compacta in $C^m(\mathbb R^n)$, by the dual space argument, there exists a collection $\sigma_\alpha$, $|\alpha|\le m$, of finite signed measures with support in some compact subset $X$ of $\mathbb R^n$such that the functional
$$
\Lambda[f] = \sum_{|\alpha| \le m} \int_{\mathbb R^n } D^\alpha f \, d \sigma_\alpha
$$
vanishes on $\mathfrak{N}_n [\Psi]$, but not identically on $C^m(\mathbb R^n)$.

For $\epsilon > 0$, let us define the functionals $\Lambda_\epsilon$ as
$$
\Lambda_\epsilon[f] = \sum_{|\alpha| \le m} \int_{\mathbb R^n } D^\alpha f \, J_\epsilon \sigma_\alpha \, dx
$$
The integrals exist because $J_\epsilon \sigma_\alpha$ have compact support. For all $f \in C^m(\mathbb R^n)$, using the lemma above:
\begin{align*}
    \Lambda_\epsilon[f] 
    &= \int_{\mathbb R^n} \left[ \sum_{|\alpha| \le m}\int_{\mathbb R^n} D^\alpha T_y f \, d\sigma_\alpha \right] w_\epsilon(y) \, dy \\
    &= \int_{\mathbb R^n} \Lambda[T_y f] w_\epsilon(y) \, dy
\end{align*}
and
$$
\lim_{\epsilon \to 0} \Lambda_\epsilon[f]  = \Lambda[f] 
$$

Integration by parts yields
$$
\Lambda_\epsilon[f] = 
\int_{\mathbb R^n} \left[ \sum_{|\alpha| \le m} (-1)^\alpha  D^\alpha J_\epsilon \sigma_\alpha \right] f \, dx
$$
If we define $h_\epsilon = \sum_{|\alpha| \le m} (-1)^\alpha  D^\alpha J_\epsilon \sigma_\alpha$, then $\Lambda_\epsilon[f] = \int_{\mathbb R^n} h_\epsilon f \, dx $.

Let us write $\Psi_{\theta, \theta_0} (x) = \Psi(\theta \cdot x+\theta_0)$. Suppose that $\Lambda$ vanishes on $\mathfrak{N}_n [\Psi]$. Since $\Psi_{\theta, \theta_0} \in \mathfrak{N}_n [\Psi]$, we infer $\Lambda[\Psi_{\theta, \theta_0}]=0$. Given that $T_y \Psi_{\theta, \theta_0} = \Psi_{\theta, \theta_0 - \theta \cdot y}$, it follows that
\begin{align*}
\int_{\mathbb R^n} \Psi_{\theta, \theta_0}  h_\epsilon  \, dx 
  & = \Lambda_\epsilon [\Psi_{\theta, \theta_0}] \\
  & =  \int_{\mathbb R^n} \Lambda[T_y \Psi_{\theta, \theta_0}] w_\epsilon(y) \, dy \\
  & = 0
\end{align*}

By assumption, $\Psi$ is bounded and nonconstant, so Theorem \ref{thm:hornik-discriminatory} implies that $h_\epsilon \equiv 0$.

As a consequence, for all functions $f \in C^m(\mathbb R^n)$, $\Lambda_\epsilon[f]=\int_{\mathbb R^n} h_\epsilon f \, dx$ vanishes  and $\Lambda[f] = \lim_{\epsilon \to 0} \Lambda_\epsilon[f] = 0$, which was ruled out by assumption.

Therefore, $\mathfrak{N}_n [\Psi]$ is uniformly m-dense on compacta in $C^m(\mathbb R^n)$. We must now complete our proof by showing that, for every function $f \in C^m(\mathbb R^n)$ and $\epsilon > 0$, there is a function $g \in \mathfrak{N}_n [\Psi]$ such that $\| f-g \|_{m,p,\mu} < \epsilon$.

Let $X$ be a compact set containing the support of $\mu$. Then we can establish a standard bound showing that a measure-based \(L^{p}\) norm is controlled by a uniform (supremum) norm over a compact set.
$$
\| f-g \|_{m,p,\mu} \le \gamma \| f-g \|_{m,\mu,X}
$$
where $\gamma^p = \mu(\mathbb R^n) \# \{ \alpha: |\alpha| \le m\}$ and $\# \{ \alpha: |\alpha| \le m\}$ is the cardinality (total count) of all multi-indices of order at most $m$ (i.e. the total number of partial derivatives of $f - g$ from order $0$ up to order $m$).

If we choose $g \in \mathfrak{N}_n [\Psi]$ such that $\| f-g \|_{m,\mu,X} < \epsilon/\gamma$ (something we can do because we just proved that $\mathfrak{N}_n [\Psi]$ is uniformly m-dense on compacta), then $\| f-g \|_{m,p,\mu} < \epsilon$. QED.
\end{proof}

\begin{theorem}

If $\Psi \in C^m(\mathbb R^n)$ is nonconstant and and all its
derivatives up to order $m$ are bounded, then $\mathfrak{N}_n [\Psi]$ is dense in $C^{m,p}(\mu)$ for all finite measures $\mu$ on $\mathbb R^n$.
\end{theorem}

\begin{remark}
The conditions imposed on $\Psi$ are very general and they are satisfied by all common smooth squashing activation functions.
\end{remark}

\begin{proof}
The proof of this theorem is parallel to the proof of Theorem \ref{thm:hornik-cm-density}. Let $C^{m,\mu}(\mathbb R^n)$ be the space of all functions $f \in C^m(\mathbb R^n)$ that, with their derivatives up to order $m$, are bounded:
$$
C^{m,\mu}(\mathbb R^n) = \{ f \in C^m(\mathbb R^n): \|D^\alpha f \|_\mu < \infty, |\alpha| \leq m \}
$$

$C^{m,\mu}(\mathbb R^n)$ is a dense subset of $C^{m,p}(\mu)$. Since $\Psi \in C^m(\mathbb R^n)$, then $\mathfrak{N}_n [\Psi] \subset C^{m,\mu}(\mathbb R^n) \subset C^{m,p}(\mu)$.

If $\mathfrak{N}_n [\Psi]$ is not dense in $C^{m,p}(\mu)$, the dual space argument yields the existence of functions $g_\alpha \in L^q(\mu)$, with $|\alpha| \leq m$ and $q=p/(p-1)$, such that the functional $\Lambda [f] = \sum_{|\alpha| \leq m} \int_{\mathbb R^n} D^\alpha f g_\alpha \, d\mu$ vanishes on $\mathfrak{N}_n [\Psi]$, but not identically on $C^{m,p}(\mu)$.

To complete the proof, just follow the proof of Theorem \ref{thm:hornik-cm-density} with the finite signed measures $\sigma_\alpha$ given by $d\sigma_\alpha = g_\alpha d\mu$ and $C^{m,\mu}(\mathbb R^n)$ instead of $C^{m}(\mathbb R^n)$.
\end{proof}

\repthm{thm:hornik-derivative}

\begin{proof}
The theorem is essentially an immediate consequence of Theorem \ref{thm:hornik-cm-density} once we identify $C^k(K)$ with the corresponding $C^{k,p}(\mu)$ space for a suitable measure $\mu$. A concise argument is as follows:

Let $K\subset\mathbb R^n$ be compact and let $f\in C^k(K)$. Choose any finite measure $\mu$ whose support is exactly $K$; for example, one may take a finite Borel measure satisfying $\operatorname{supp}(\mu)=K$.

Theorem \ref{thm:hornik-cm-density} states that if $\Psi\in C^k(\mathbb R^n)$ is bounded and nonconstant, then the neural-network class $\mathfrak N_n[\Psi]$ is uniformly k-dense on compacta in $C^{k,p}(\mu)$. By definition of uniform k-density, for every $f\in C^{k,p}$ and every $\epsilon>0$, there exists a neural network $f_N\in\mathfrak N_n[\Psi]$ such that
$$
\max_{|\alpha|\le k}\sup_{x\in K} \bigl|D^\alpha f(x)-D^\alpha \tilde f_N(x)\bigr| <\epsilon
$$

Since $\operatorname{supp}(\mu)=K$, the space $C^{k,p}(\mu)$ may be identified with $C^k(K)$, and the quantity on the left is precisely the $C^k(K)$-norm of the approximation error:
$$
\|f-\tilde f_N\|_{C^k(K)} = \max_{|\alpha|\le k}\sup_{x\in K} |D^\alpha f(x)-D^\alpha \tilde f_N(x)|
$$

Therefore, for every $f\in C^k(K)$ and every $\epsilon>0$, there exists a neural network $\tilde f_N$ such that
$$
\|f-\tilde f_N\|_{C^k(K)} <\epsilon
$$

This is exactly the statement of the theorem, which is just a specialization of the density result to the case $m=k$ and a compactly-supported measure whose support is $K$.
\end{proof}

Therefore, a single-layer feedforward network can learn the true underlying function and its partial derivatives, making no ``avoidable'' systematic errors \cite{gallant1988}.

\subsection*{Leshno's Universal Approximation Theorem}

In 1993, Moshe Leshno, Vladimir Ya. Lin, Allan Pinkus, and Shimon Schocken generalized previous universal approximation theorems with the following result: ``A standard multilayer feedforward network with a locally bounded piecewise continuous activation function can approximate any continuous function to any degree of accuracy if and only if the network's activation function is not a polynomial'' \cite{leshno1993}. They also remarked that the threshold or bias is essential, without it their theorem does not hold.

Previous approximation results stated that, whenever activation functions satisfy some assumptions, which varied from paper to paper, multilayer neural networks are universal approximators. Leshno et al. \cite{leshno1993} established the necessary and sufficient conditions for universal approximation.

The function $f: \mathbb R^n \to \mathbb R$ computed by a single-hidden-layer neural network with $n$ inputs, $h$ hidden neurons, and a single linear output neuron is given by:
$$
f(x) = \sum_{j=1}^h \alpha_j \sigma \left( \sum_{i=1}^n w_{ij} x_i  + b_j \right) = \sum_{j=1}^h \alpha_j \sigma \left( w_j \cdot x  + b_j \right)
$$
The parameters of the hidden layer consist of the number of hidden neurons ($h$), a set of weights ($\{w_{ij}\}$), a set of biases/thresholds ($\{b_j\}$), and an activation function ($\sigma: \mathbb R \to \mathbb R$). Its parameter space is $\Lambda= \langle h, \{w_{ij}\}, \{b_j\}, \sigma \rangle$. A particular configuration of the hidden layer is described by the particular quadruple $\omega \in \Lambda$.
The hidden layer characterized by $\omega$ is $\mathcal N_\omega$, the function computed by $\mathcal N_\omega$ is $f_\omega: \mathbb R \to \mathbb R^n$, and the family of such functions is $\mathcal F = \{ f_\omega | \omega \in \Lambda \}$.

Discovering all the functions that can be approximated by networks with a hidden layer of the form $\mathcal N_\omega$ is equivalent to characterizing the closure $\overline{\mathcal F}$ of $\mathcal F$. This closure is the complete set of functions that a neural network can either produce directly or approximate with arbitrarily small error. Whether a target function belongs to the closure depends entirely on the metric (the rule used to measure the distance or error between two functions from $\mathbb R^n$ to $\mathbb R$).

\begin{definition}[Metric.]
A metric on a set $S$ is a function $d: S \times S \to \mathbb R$ that satisfies the following four properties:
\begin{itemize}
    \item
    Non-negativity: $d(s,t) \ge 0$.
    \item
    Identity of indiscernibles: $s=t \Leftrightarrow d(s,t)=0$.
    \item
    Symmetry: $d(s,t)=d(t,s)$.
    \item 
    Triangle inequality: $d(s,t)\le d(s,u) + d(u,t)$.
\end{itemize}
\end{definition}

\begin{definition}[Closure.]
The closure of a set $S$ of a metric space $(X, d)$ is defined as
$$
\overline{S} 
= \{ r \, | \, \forall \epsilon>0, \exists s \in S, d(s,t) < \epsilon \}
$$
\end{definition}

A function $f\in L^\infty(\Omega)$ is essentially bounded on a measurable set $\Omega$ with respect to the essential supremum norm $\|f\|_{L^{\infty}(\Omega)}$. 
A function $f\in L^\infty_{loc}(\Omega)$ is locally essentially bounded on a domain $\Omega$ (i.e. an open set in $\mathbb R^n$) if, for every compact set $K \subset \Omega$, $f\in L^\infty (K)$.

\begin{definition}
    A set $F$ of functions in $L^\infty_{loc}(\Omega)$ is dense in $C(\mathbb R^n)$ if, for every function $f\in C(\mathbb R^n)$ and every compact set $K \subset \mathbb R^n$, there exists a sequence of functions $f_j \in F$ such that
    $$
    \lim_{j \to \infty} \| f - f_j \|_{L^\infty_{loc}(K)} = 0
    $$
\end{definition}

If a set of functions $F$ is dense in $C(\mathbb R^n)$, for every continuous function $f \in C(\mathbb R^n)$ there is a function $\tilde f \in F$ that is a good approximation to $f$.

Let $M$ be the set of functions in $L^\infty_{loc}(\mathbb R)$ whose closure
of the set of points of discontinuity is of zero Lebesgue measure (i.e. they are Riemann integrable on every compact interval). For any $\sigma \in M$, interval $[a,b]$, and $\epsilon>0$, there is a finite number of open intervals, whose union $U$ is of measure $\delta$, such that $\sigma$ is uniformly continuous on $[a,b] \setminus U$. 

\begin{theorem}
\label{thm:leshno-span}

For $\sigma \in M$, let the span\footnote{The span of a set of functions is the collection of all new functions you can create by scaling (multiplying by constants) and adding those original functions together.}
$$
\Sigma_n = \operatorname{span} \{ \sigma(w \cdot x + b): w \in \mathbb R^n, b \in \mathbb R \}
$$
$\Sigma_n$ is dense in $C(\mathbb R^n)$ if and only if $\sigma$ is not an algebraic polynomial.
\end{theorem}

\begin{proof} \cite{leshno1993}

\noindent
{\it
If $\sigma$ is a polynomial, then $\Sigma_n$ is not dense in $C(\mathbb R^n)$.
}
\begin{quote}
If $\sigma$ is a polynomial of degree $k$, then $\sigma(w \cdot x + b)$ is
a polynomial of degree $k$. Then, $\Sigma_n$ is exactly the set of algebraic polynomials of degree at most $k$. Therefore $\Sigma_n$ cannot be dense in $C(\mathbb R^n)$.\footnote{By the Stone–Weierstrass theorem (Theorem \ref{thm:stone-weierstrass}), the set of all algebraic polynomials (of arbitrarily high degree) is dense in \(C(\mathbb{R}^n)\). However, once the polynomial degree is limited ($\le k$), both higher-degree polynomials and non-polynomial functions are unreachable.}
\end{quote}

\noindent
Hence, we will always assume that $\sigma$ is not a polynomial. Now, we can reduce our problem to $\Sigma_1$:

\noindent
{\it
If $\Sigma_1$ is dense in $C(\mathbb R)$, then $\Sigma_n$ is dense in $C(\mathbb R^n)$.
}
\begin{quote}
    The space $V =\operatorname{span} \{ f(a \cdot x)| a\in \mathbb R^n, f\in C(\mathbb R^n)\}$ is dense in $C(\mathbb R^n)$. For any compact set $K\subset \mathbb R^n$, V is dense in $C(K)$.
    Let $g \in C(\mathbb R^n)$. For $\epsilon > 0$, there exists $f_i \in C(\mathbb R)$ such that
    $$
    | g(x) - \sum_{k=i}^k f_i (a_i \cdot x) | < \epsilon/2
    $$
    for all $x \in K$. where $a_i \in \mathbb R^n$. $\{ a_i \cdot x |x \in K\} \subseteq [\alpha_i, \beta_i]$ for some finite interval $[\alpha_i, \beta_i]$.

    Given that $\Sigma_1$ is dense in $[\alpha_i, \beta_i]$, there exist constants $c_{ij}, w_{ij}, b_{ij}$ such that
    $$
    | f_i(x) - \sum_{j=i}^{m_i} c_{ij} \sigma (w_{ij} y + b_{ij}) | < \epsilon/2k
    $$
    for all $y \in [\alpha_i, \beta_i]$. Therefore,
    $$
    | f(x) - \sum_{k=i}^k\sum_{j=i}^{m_i} c_{ij} \sigma (w_{ij} (a_i \cdot x) + b_{ij}) | < \epsilon
    $$
    for all $x \in K$. Therefore, $\Sigma_1$ is dense in $C(\mathbb R)$ implies that  $\Sigma_n$ is dense in $C(\mathbb R^n)$.
\end{quote}

\noindent
Let us now complete the proof using the following steps:

\begin{itemize}

\item
If $\sigma \in C^\infty(\mathbb R)$, then $\Sigma_1$ is dense in $C(\mathbb R)$.

\item
For each $\varphi \in C_0(\mathbb R)$, $\sigma * \varphi \in \overline{\Sigma_1}$

\item
If, for some $\varphi \in C_0^\infty (\mathbb R)$, $\sigma * \varphi$ is not a polynomial, then $\Sigma_1$ is dense in $C(\mathbb R)$.

\item
If, for all $\varphi \in C_0^\infty (\mathbb R)$, $\sigma * \varphi$ is a polynomial, then $\exists d\in \mathbb{N}$ such that $\sigma * \varphi$ is a polynomial of degree at most $d$ for all $\varphi \in C_0^\infty (\mathbb R)$.

\item
If $\sigma * \varphi$ is a polynomial of degree at most $d$ for all $\varphi \in C_0^\infty (\mathbb R)$, then $\sigma$ is a polynomial of degree at most
$d$ (a.e.).

\end{itemize}

\noindent
{\it
Step 1.\\ 
If $\sigma \in C^\infty(\mathbb R)$, then $\Sigma_1$ is dense in $C(\mathbb R)$.
}

\begin{quote}
    If $\sigma \in C^\infty(\mathbb R)$, $(\sigma((w+h)x+b)-\sigma(wx+b))/h \in \Sigma_1$ and, therefore, $(d/dw) \sigma(wx+b) \in \overline{\Sigma_1}$ for every $h \ne 0$.

    By the same argument, $(d^k/dw^k) \sigma(wx+b) \in \overline{\Sigma_1}$ for all $k\in \mathbb N$.

    $(d^k/dw^k) \sigma(wx+b) = x^k \sigma^{(k)}(wx+b)$, where $\sigma^{(k)}$ is the $k$-th derivative of $\sigma$. Since $\sigma$ is not a polynomial, there exists $\theta_k \in \mathbb R$ such that $\sigma^{(k)}(\theta_k) \ne 0$. Hence,
    $$
    x^k \sigma^{(k)}(\theta_k) = \frac{d^k}{dw^k} \sigma(wx+b) \vert{}_{w=0, b=\theta_k} \in \overline{\Sigma_1}
    $$
    This implies that $\overline{\Sigma_1}$ contains all polynomials.\footnote{Because \(\sigma^{(k)}(\theta_k)\) is a non-zero constant, we can divide by it to obtain \(x^{k}\) as follows
    $$
    x^{k}=\frac{1}{\sigma ^{(k)}(\theta _{k})}\Big(x^{k}\sigma ^{(k)}(\theta _{k})\Big)\in \overline{\Sigma _{1}}
    $$
    Since \(\overline{\Sigma _{1}}\) contains \(1, x, x^2, x^3, \dots\) for all \(k \ge 0\), it contains every linear combination \(\sum _{k=0}^{m}a_{k}x^{k}\), i.e. all polynomials.}
    By Weierstrass Approximation Theorem,\footnote{Weierstrass Approximation Theorem: Every continuous function on a closed and bounded (compact) interval \(K \subset \mathbb{R}\) can be uniformly approximated as closely as desired by a polynomial.} it follows that $\overline{\Sigma_1}$ contains $C(K)$ for all compact sets $K \subset \mathbb R$. In other words, $\Sigma_1$ is dense in $C(\mathbb R)$.
\end{quote}

\noindent
{\it
Step 2.\\ 
For each $\varphi \in C_0(\mathbb R)$, $\sigma * \varphi \in \overline{\Sigma_1}$
}

\begin{quote}
    Assume that $\operatorname{supp} \varphi \subseteq [-\alpha,\alpha]$. We will prove that we can uniformly approximate $\sigma * \varphi$ from $\Sigma_1$ on $[-\alpha,\alpha]$: $\sum_i \sigma(x-y_i)\varphi(y_i) \Delta y_i$ uniformly converges to $\sigma * \varphi$ on $[-\alpha,\alpha]$, where $y_i=-\alpha+2i\alpha/m$ and $\Delta y_i = 2\alpha/m$ with $i=1..m$.

    Given $\epsilon>0$, we choose $\delta>0$ so that $10\delta \| \sigma \|_{L^\infty [-2\alpha, 2\alpha]} \| \varphi \|_{L^\infty} \le \epsilon$.

    For this $\delta>0$, there is a finite number $r(\delta)$ of intervals, the measure of whose union $U$ is $\delta$, such that $\sigma$ is uniformly continuous on $[-2\alpha, 2\alpha]$. We choose $m$ sufficiently large, so that $m\delta < \alpha r(\delta)$.

    If $|s-t| \le 2\alpha/m$, by the uniform continuity of $\varphi$:
    $$
    | \varphi(s) - \varphi(t) | \le \frac{\epsilon}{2\alpha \| \sigma \|_{L^\infty [-2\alpha, 2\alpha]}}
    $$

    If $s,t \in [-2\alpha, 2\alpha]/U$ and $|s-t| \le 2\alpha/m$, by the uniform continuity of $\sigma$ on $[-2\alpha, 2\alpha]/U$:
    $$
    | \sigma(s) - \sigma(t) | \le \frac{\epsilon}{\|\varphi \|_{L^1}}
    $$

    For $x\in [-\alpha, \alpha]$, set $\Delta_i = [y_{i-1},y_i]$ with $y_0=\alpha$. Since $\operatorname{supp} \varphi \subseteq [-\alpha,\alpha]$:
    \begin{align*}
      & \left| \int \sigma(x-y)\varphi(y)dy - \sum_{i=1}^m \int_{\Delta_i} \sigma(x-y_i)\varphi(y)dy \right| \\
      & \le \sum_{i=1}^m \int_{\Delta_i} | \sigma(x-y) - \sigma(x-y_i)| \, |\varphi(y)| \, dy
    \end{align*}

    If $x-\Delta_i$ does not intersect $U$, from our inequality on $| \sigma(s) - \sigma(t) |$:
    \begin{align*}
    \int_{\Delta_i} & | \sigma(x-y) - \sigma(x-y_i)| \, |\varphi(y)| \, dy \\
    & \le \frac{\epsilon}{\|\varphi \|_{L^1}} \int_{\Delta_i} |\varphi(y)| \, dy 
    \end{align*}

    When $(x-\Delta_i) \cap U \ne \emptyset$, we denote such intervals as $\tilde \Delta_i$. $U$ has measure $\delta$ and is composed of $r(\delta)$ intervals. The total length of the $\tilde \Delta_i$ intervals is, at most, $\delta+(4\alpha/m) r(\delta)$. By our choice of $m$, $\delta+(4\alpha/m) r(\delta) \le 5\delta$. From our choice of $\delta$:
    \begin{align*}
    \int_{\tilde \Delta_i} & | \sigma(x-y) - \sigma(x-y_i)| \, |\varphi(y)| \, dy \\
    & \le 2 \| \sigma \|_{L^\infty [-2\alpha, 2\alpha]} \| \, \varphi \|_{L^\infty} \, 5\delta\le \epsilon
    \end{align*}

    Finally,
    \begin{align*}
    & \left| \sum_{i=1}^m \int_{\Delta_i} \sigma(x-y_i)\varphi(y)dy 
           - \sum_{i=1}^m \sigma(x-y_i)\varphi(y_i) \Delta y_i \right| \\
    & = \left| \sum_{i=1}^m \int_{\Delta_i} \sigma(x-y_i) [\varphi(y)-\varphi(y_i)] dy \right| \\
    & \le \sum_{i=1}^m \int_{\Delta_i} |\sigma(x-y_i)| \, |\varphi(y)-\varphi(y_i)| \, dy \\
    & \le \sum_{i=1}^m \int_{\Delta_i} |\sigma(x-y_i)| \, dy
     \left[ \frac{\epsilon}{2\alpha \| \sigma \|_{L^\infty [-2\alpha, 2\alpha]}} \right] \le \epsilon
    \end{align*}
    where the last inequality comes from our expression on $| \varphi(s) - \varphi(t) |$.

    Therefore, we obtain our desired approximation of $\sigma * \varphi$:
    $$
    \left| \int \sigma(x-y)\varphi(y)dy - \sum_{i=1}^m \sigma(x-y_i)\varphi(y_i) \Delta y_i \right| \le 3\epsilon
    $$

\end{quote}

\noindent
{\it
Step 3.\\ 
If, for some $\varphi \in C_0^\infty (\mathbb R)$, $\sigma * \varphi$ is not a polynomial, then $\Sigma_1$ is dense in $C(\mathbb R)$.
}

\begin{quote}
    From step 2 above, $\sigma * \varphi \in \overline{\Sigma_1}$. Then $(\sigma * \varphi)(wx+b) \in \overline{\Sigma_1}$. For any $\varphi \in C_0^\infty (\mathbb R)$, $\sigma * \varphi \in C^\infty (\mathbb R)$. From step 1, if $\sigma * \varphi$ is not a polynomial, $\Sigma_1$ is dense in $C(\mathbb R)$.
\end{quote}

\noindent
We now assume that $\sigma * \varphi$ is a polynomial for all $\varphi \in C_0^\infty (\mathbb R)$. We will conclude from this fact that $\sigma$ is itself a polynomial (a.e.).

\noindent
{\it
Step 4.\\ 
If, for all $\varphi \in C_0^\infty (\mathbb R)$, $\sigma * \varphi$ is a polynomial, then $\exists d\in \mathbb{N}$ such that $\sigma * \varphi$ is a polynomial of degree at most $d$ for all $\varphi \in C_0^\infty (\mathbb R)$.
}

\begin{quote}
    For any $a<b$, the set $C_0^\infty[a,b]$ is the set of all $C_0^\infty$ functions with support in $[a,b]$. 

    In the case of $\varphi \in C_0^\infty[a,b]$, we define a metric $\rho$ on $C_0^\infty[a,b]$ as 
    $$
    \rho(\varphi_1,\varphi_2) 
    = \sum_{n=0}^\infty 2^{-n} 
       \frac{ \| \varphi_1-\varphi_2 \|_n}{1+ \| \varphi_1-\varphi_2 \|_n}
    $$
    where $\|\varphi\|_n = \sum_{j=0}^n \sup_{x\in [a,b]} |\varphi^{(j)}(x)|$. $C_0^\infty[a,b]$ with the metric $\rho$ is a Fréchet space (a complete metric vector space).

    By our assumption, $\sigma * \varphi$ is a polynomial for every $\varphi \in C_0^\infty[a,b]$. We define the space
    $$
    V_k = \{ \varphi \in C_0^\infty[a,b] \, | \, degree(\sigma * \varphi)\le k \}
    $$
    $V_k$ is a closed subspace, $V_k \subseteq V_{k+1}$, and $\cup_{k=0}^{\infty} V_k = C_0^\infty[a,b]$.

    Since $C_0^\infty[a,b]$ is a complete metric space, by Baire's Category Theorem,\footnote{Baire's Category Theorem states that every non-empty complete metric space is of the second category in itself. The first category includes meager (small) sets, which can be written as a countable union of nowhere dense sets (e.g. $\mathbb Q$ is meager in $\mathbb R$). The second category contains non-meager (topologically large) sets. $C_0^\infty[a,b]$ is of the second category and, therefore, some $V_d$ contains a non-void open set. Because $V_d$ is a vector space, $V_d = C_0^\infty[a,b]$.}
    there is an integer $d$ such that $V_d = C_0^\infty[a,b]$. This completes the proof for $\varphi \in C_0^\infty[a,b]$.

    For the general case, the degree $d$ does not depend on $[a,b]$ yet, by translation, $d$ depends at most of the length of the interval. For $\varphi \in C_0^\infty[A,B]$, we can find $\varphi_i \in C_0^\infty[a_i,b_i]$ such that $[A,B]\subseteq \cup_{i=1}^k [a_i,b_i]$, $b_i-a_i=b-a$, and $\varphi = \sum_{i=1}^k \varphi_i$. Then $\sigma * \varphi = \sum_{i=1}^k \sigma * \varphi_i$ and every $\sigma * \varphi_i$ is a polynomial of degree $\le d$, so $degree(\sigma * \varphi) \le d$.
\end{quote}

\noindent
{\it
Step 5.\\ 
If $\sigma * \varphi$ is a polynomial of degree at most $d$ for all $\varphi \in C_0^\infty (\mathbb R)$, then $\sigma$ is a polynomial of degree at most
$d$ almost everywhere (a.e.).
}

\begin{quote}
    From step 4, for all $\varphi \in C_0^\infty(\mathbb R^n)$,
    $$
    \sigma * \varphi^{(d+1)}= \int \sigma(x-y)\varphi^{(d+1)}(y)dy = 0
    $$
    From standard results in Distribution Theory, if smoothing a function \(\sigma \) always produces a polynomial of degree at most \(d\), then \(\sigma \) itself must already be a polynomial of degree at most \(d\) almost everywhere.\footnote{Think of \(\sigma(x)\) as a raw 1D signal (e.g. an audio waveform) that might be noisy, jagged, or have sharp corners (like the ReLU activation function). Convolving \(\sigma \) with a smooth bump \(\eta _{\epsilon }\) is applying a Gaussian-like low-pass filter of window width \(\epsilon \). With a tiny window width, the blurred output \(\sigma_\epsilon(x)\) stops averaging out wide neighborhoods and locks onto the true local value of the raw signal \(\sigma(x)\). Why ``almost everywhere''? Unless \(\sigma(x)\) has a bizarre, infinitely thin spike at a single isolated point, which carries zero energy and vanishes under integration, \(\sigma_\epsilon(x)\) converges to \(\sigma(x)\) as \(\epsilon \to 0\).} Therefore, $\sigma$ is itself a polynomial of degree at most $d$ (a.e.).

    In neural network theory, this is the key step that extends Universal Approximation Theorems from smooth activations to non-smooth activations like ReLU or step functions. Why? Because, when an activation function \(\sigma \) is not differentiable, we cannot take its derivatives directly. Instead, we convolve \(\sigma \) with a smooth, compactly supported test function \(\varphi \in C_0^\infty(\mathbb{R})\). The convolution $\sigma * \varphi$ averages \(\sigma \) against \(\varphi \).
\end{quote}

If \(\sigma \) is not a polynomial, we want to guarantee that \(\sigma * \varphi\) is not a polynomial either. Proving the contrapositive (that \(\sigma * \varphi\) being a polynomial forces \(\sigma \) to be a polynomial) closes the argument and completes our proof.
\end{proof}

\begin{remark}
As mentioned in \cite{leshno1993}, an alternative proof of Theorem \ref{thm:leshno-span} can be based on Schwartz's theory of mean-periodic functions. That theory establishes that, if $\sigma \in C(\mathbb R)$ is not a polynomial, then the span of $\{ \sigma(x+\theta): \theta \in \mathbb R \}$ contains in its closure a function of the form $e^{\lambda x} \cos \nu x$ for some real $(\lambda, \nu) \ne (0,0)$. Since those functions are in $C^\infty(\mathbb R)$, Theorem \ref{thm:leshno-span} follows.
\end{remark}
\repthm{thm:leshno}
\begin{proof}
    Leshno's universal approximation theorem for non-polynomial activation functions can be proved directly from Theorem \ref{thm:leshno-span} just by connecting the topological definitions of $C(\mathbb{R}^n)$ and $C(K)$, once we identify the set of functions $L(x)$ with the span $\Sigma _{n}$ and address the continuity of $\sigma$:

    \begin{itemize}

    \item
    By definition of linear span, $\Sigma_n$ is the set of functions of the form $L(x)$.

    \item Since $\sigma: \mathbb{R} \to \mathbb{R}$ is continuous, it is automatically bounded on every compact set and has zero discontinuities, so
    $\sigma \in M$.
    
    \end{itemize}

    \noindent {\it Sufficiency ($\Leftarrow$):} Suppose \(\sigma \) is not a polynomial almost everywhere, so it is certainly not an algebraic polynomial everywhere:

    \begin{itemize}

    \item
    By Theorem \ref{thm:leshno-span}, \(\Sigma _{n}\) is dense in \(C(\mathbb{R}^n)\).

    \item
    Restrict to \(K\): Let \(K \subset \mathbb{R}^n\) be an arbitrary compact set and let \(f : K \to \mathbb{R} \in C(K)\).

    \item
    Extend \(f\) to \(\mathbb{R}^{n}\): Using Tietze Extension Theorem (Theorem \ref{thm:tietze-extension}), the continuous function \(f \in C(K)\) can be extended to a continuous function \(\tilde{f} \in C(\mathbb{R}^n)\) such that \(\tilde{f}(x) = f(x)\) for all \(x \in K\).

    \item
    Approximate on $K$: Because $\Sigma_n$ is dense in $C(\mathbb{R}^n)$, for $\tilde{f} \in C(\mathbb{R}^n)$, compact set $K$, and $\epsilon > 0$, there exists a function $L \in \Sigma_n$ such that $\sup _{x\in K}|{}L(x)-\tilde{f}(x)|<\epsilon$.

    \item
    Since \(\tilde{f}(x) = f(x)\) on \(K\) and \(L\) has the exact form given in the theorem, we have \(\Vert{}L - f\Vert{}_K < \epsilon\). Thus, the set of functions \(L(x)\) is dense in \(C(K)\) for every compact \(K \subset \mathbb{R}^n\).

    \end{itemize}
    
    \noindent {\it Necessity ($\Rightarrow$):}
    Suppose \(\Sigma _{n}\) is dense in \(C(K)\) for every compact \(K \subset \mathbb{R}^n\).

    \begin{itemize}

    \item
    Density in \(C(\mathbb{R}^n)\): Let \(g \in C(\mathbb{R}^n)\), \(K \subset \mathbb{R}^n\), and  \(\epsilon > 0\). The restriction of \(g\) to \(K\), denoted \(g\vert{}_K\), belongs to \(C(K)\). By assumption, \(\Sigma _{n}\) is dense in \(C(K)\), so there exists \(L \in \Sigma_n\) such that:
    $$
    \sup _{x\in K}|{}L(x)-g(x)|{}=\|{}L-g|{}_{K}\|{}_{K}<\epsilon
    $$
    Because this holds for every \(g \in C(\mathbb{R}^n)\), every compact \(K \subset \mathbb{R}^n\), and every \(\epsilon > 0\), it follows that \(\Sigma _{n}\) is dense in \(C(\mathbb{R}^n)\).

    \item
    Application of Theorem \ref{thm:leshno-span}: Because \(\Sigma _{n}\) is dense in \(C(\mathbb{R}^n)\), \(\sigma \) cannot be an algebraic polynomial.

    \item
    The ``almost everywhere'' condition: Suppose, for the sake of contradiction, that \(\sigma \) is equal to an algebraic polynomial \(P(t)\) almost everywhere on \(\mathbb{R}\).

    \begin{itemize}

    \item
	The set \(E = \{t \in \mathbb{R} : \sigma(t) \neq P(t)\}\) has Lebesgue measure zero.
	\item 
    The complement \(\mathbb{R} \setminus E = \{t \in \mathbb{R} : \sigma(t) = P(t)\}\) is dense in \(\mathbb{R}\), since any non-empty open interval in \(\mathbb{R}\) has strictly positive Lebesgue measure and cannot be contained in \(E\).
	\item
    Because both \(\sigma \) and the polynomial \(P\) are continuous functions on \(\mathbb{R}\) and they agree on a dense subset \(\mathbb{R} \setminus E\), they must agree everywhere on \(\mathbb{R}\) (i.e., \(\sigma(t) = P(t)\) for all \(t \in \mathbb{R}\)).
    \item
	This means \(\sigma \) is an algebraic polynomial everywhere, which leads us to a contradiction, so \(\sigma \) is not a polynomial almost everywhere.
    \end{itemize}

    \end{itemize}

\end{proof}
In conclusion, feedforward neural network can approximate any continuous function if and only if its activation function is not polynomial.
\subsection*{Barron's Approximation Error Bounds}


Andrew Barron \cite{barron1993} studied how the approximation error is related to the number of hidden neurons in a feedforward neural network. His results hold for approximating functions whose Fourier transform first moment is bounded: an average of the norm of the frequency vector weighted by the Fourier magnitude distribution is used to measure how functions oscillate (i.e. their smoothness).

\begin{definition}[Barron norm.]
Given a function $f: \mathbb R^n \to \mathbb R$, its Barron norm is given by 
    $$
    C_f = \int_{\mathbb R^n} \| \omega\| \, |\mathcal F[f](\omega) |\, d\omega
        = \int_{\mathbb R^n} \| \omega\| \, |\widehat f(\omega) |\, d\omega
    $$
    $$
    \| \omega\| = \| \omega \|_{L^2} = \sqrt{\omega \cdot \omega}
    $$
\end{definition}

\begin{definition}[Barron function.]
  $\Gamma$ is the class of functions on $\mathbb R^n$ for which their Barron norm $C_f= \int |\omega| \mathcal F (d\omega)$ is finite.
\end{definition}

The finiteness of $C_f$ imposes that the function $f$ has a gradient with an integrable Fourier representation. Functions with finite $C_f$ are continuously differentiable on $\mathbb R^n$ and their gradient holds $\mathcal F[\nabla f] (\omega) = i \omega \widehat f(\omega)$.

\begin{definition}[$\Gamma_C$.]
    For each $C>0$, $\Gamma_C$ is the set of functions $f: \mathbb R^n \to \mathbb R$ such that $C_f \le C$.
\end{definition}

\begin{definition}[$\Gamma_B$.]
  Analogously to $\Gamma$, $\Gamma_B$ is defined for functions that are approximated on bounded measurable subsets $B$ of their domain in $\mathbb R^n$. $\Gamma_B$ is the set of functions $f: \mathbb B \to \mathbb R$ such that $C_f \le C$.
\end{definition}

Linear functions $f(x)=w \cdot x$ and the class of infinitely differentiable functions on $\mathbb R*n$ are not contained in $\Gamma$ but are in $\Gamma_B$ for any bounded set $B$. Containment in $\Gamma_B$ for every bounded set $B$ is denoted by $\Gamma_* = \cap_B \Gamma_B$.

\begin{definition}[$\Gamma_{C,B}$]
  is the class of functions $f$ in $\Gamma_B$ for which $C_f= \int |\omega|_B \mathcal F (d\omega) \le C$, where $|\omega|_B = \sup_{x \in B} | \omega \cdot x|$.
\end{definition}

For the ball $B_r= \{ x: |x| \le r \}$, the norm simplifies to $|\omega|_B = r |\omega|$.

An upper bound of the approximation error by linear combinations of sigmoidal functions can be obtained if we take into account that, when the function has an integrable representation in terms of parameterized basis functions, then a random sample of the parameters of the basis functions can lead to an accurate approximation.

The approximation error is measured by integrating the squared error with respect to an arbitrary probability measure $\mu$ on the ball $B_r = \{ x: |x| \le r\}$ of radius $r>0$.

\begin{lemma}[Convex combinations in a Hilbert space.]
\label{lemma:barron-hilbert}

If $\overline f$ is the closure of the convex hull of a set $G$ in a Hilbert space, with $\|g\|<b$ for each $g \in G$, for every $h\ge1$ and every $c'>b^2-\| \overline f\|^2$, there is an $f_h$ in the convex hull of $h$ points in G such that 
$$
\| \overline f - f_h \|^2 \le \frac{c'}{h}
$$
\end{lemma}

\begin{proof}
Given $n \ge 1$ and $\delta >0$, let $f^*$ a point in the convex hull of $G$ with $\| \overline f - f^* \| \le \delta/h$. Therefore, $f^*$ is of the form $\sum_{k=1}^{m} \gamma_k g_k^*$, with $g_k^* \in G$, $\gamma_k \ge 0$, $\sum_{k=1}^{m} \gamma_k=1$, for some sufficiently large $m$.

Let $g$ be randomly drawn from the set $\{ g_1^*.. g_m^* \}$ with $P(g=g_k^*)=\gamma_k$. Let $g_1.. g_h$ be independently drawn from the same distribution as $g$. Their sample average is $f_h=(1/h)\sum_{i=1}^{h} g_i$. Then $E[f_h]=f^*$. The expected value of the squared norm of the error is $E[\|f_h-f^* \|^2] = (1/h) E[\| g-f^* \|^2] = (1/h) (E[\| g \|^2] - \| f^* \|^2)$, which is bounded by $(1/h) (b^2 - \| f^* \|^2)$.

Since the expected value is bounded, there must exist $g_1.. g_h$ for which $\|f_h-f^* \|^2 \le (1/h) (b^2 - \| f^* \|^2)$. The proof of Lemma \ref{lemma:barron-hilbert} is completed using the triangle inequality and $\| \overline f - f_h \|^2 \le \delta/h$ for a sufficiently small $\delta$. By combining the approximation of $\bar f$ by $f^*$ with the probabilistic approximation of $f^*$ by $f_h$:
$$
\|f_h-f^*\|^2 \le \frac1h\bigl(b^2-\|f^*\|^2\bigr) \le \frac{b^2}{h}
$$
We apply the triangle inequality:
$$
\|\bar f-f_h\| \le \|\bar f-f^*\|+\|f^*-f_h\| \le \frac{\delta}{h}+\frac{b}{\sqrt h}
$$
Squaring gives
$$
\|\bar f-f_h\|^2 \le \left(\frac{\delta}{h}+\frac{b}{\sqrt h}\right)^2 = \frac{\delta^2}{h^2} +\frac{2b\delta}{h^{3/2}} +\frac{b^2}{h}
$$
Since
$$
\frac{\delta^2}{h^2} +\frac{2b\delta}{h^{3/2}} \in o\!\left(\frac1h\right)
$$
for sufficiently large $h$. Then, for any $\varepsilon>0$:
$$
\|\bar f-f_h\|^2 \le \frac{b^2+\varepsilon}{h}
$$
Equivalently, there exists a constant $c'>0$ such that
$$
\| \bar f-f_h\|^2\le \frac{c'}{h}
$$
\end{proof}

\begin{lemma}
\label{lemma:barron-gamma}
For every function in $\Gamma_{C,B}$ and every sigmoidal function $\sigma$, the function $\overline f(x) = f(x)-f(0)$ is in the closure of the convex hull of $G_\sigma$, where the closure is taken in $L_2(\mu,B)$.
\end{lemma}

\begin{proof}
    Let $\widehat F(d\omega)=e^{i \Theta(\omega)} F(d\omega)$ be the magnitude and phase decomposition of the Fourier transform of an extension of the function $f$ on $B$ for which $\int |\omega|_B F(d\omega) \le C$.
    Let $\Omega=\{\omega \in \mathbb R^n: \omega \ne 0\}$. Since $f(x)$ is real-valued:
    \begin{align*}
     f(x)-f(0) 
     &= \operatorname{Re} \int (e^{i \omega \cdot x} - \mathbf 1 ) \widehat F (d\omega)\\
     &= \operatorname{Re} \int (e^{i \omega \cdot x} - \mathbf 1 ) e^{i \Theta(\omega)} F(d\omega)\\
     &= \int_\Omega (\cos(\omega \cdot x + \Theta(\omega))-\cos(\Theta(\omega))) F(d\omega)\\
     &= \int_\Omega g(x,\omega) \Lambda(d\omega)
    \end{align*}
    for $x \in B$, where $C_{f,B}=\int |\omega|_B F(d\omega) \leq C$, $\Lambda(d\omega)=|\omega|_B F(d\omega)/ C_{f,B}$ is a probability distribution, and
    $$
    g(x,\omega) = \frac{C_{f,B}}{|\omega|_B} ( \cos(\omega \cdot x + \Theta(\omega))-\cos(\Theta(\omega))) 
    $$
    For $x \in B$ and $\omega \ne 0$, $g(x,\omega) \le C | \omega \cdot x | / |\omega|_B \le C$.

    The integral above represents $\overline f$ as an infinite convex combination of functions in the class $G_{cos}$:
    \begin{align*}
     G_{cos} = \left\{
      \frac{\gamma}{|\omega|_B} (\cos(\omega \cdot x + b)-\cos(b)): \right.
      \\ \left. \omega \ne 0, |\gamma| \le C, b \in \mathbb R \right\}
    \end{align*}

    \noindent
    It follows that  $\overline f$ is in the closure of the convex hull $G_{cos}$: 
    {\it    
    For each $f \in \Gamma_{C,B}$, the function $f(x)-f(0)$ is in the closure of the convex hull of $G_{cos}$.
    }
    
    \begin{quote}
    If $\omega_1 .. \omega_h$ is a random sample of $h$ points, independently drawn from the distribution $\Lambda$, by Fubini's theorem
    \begin{align*}
    E & \int_{B_r} \left( f(x)- \sum_{i=1}^{h} g(x,\omega_i) \right)^2 \mu(dx) \\
      & = \int_{B_r} E \left( f(x)- \sum_{i=1}^{h} g(x,\omega_i)\right)^2 \mu(dx) \\
      & = \frac{1}{h} \operatorname{var}(g(x,\omega)) \mu(dx) \\
      & \le \frac{C^2}{h}
    \end{align*}
    Therefore, the mean value of the squared $L_2(\mu,B)$ norm of a convex combination in $G_{cos}$ converges to zero as $h \to \infty$, at rate $O(1/h)$ in accordance to Lemma \ref{lemma:barron-hilbert}. 
    \end{quote}

    \noindent
    {\it
    $G_{cos}$ is in the closure of the convex hull of $G_{step}$.
    }

    \begin{quote}
    Each function in $G_{cos}$ is a composition of a 1D sinusoidal function $g(z)= \gamma/|\omega|_B (\cos(|\omega|_B x + b)-\cos(b))$ and a linear function $z(x)= \alpha \cdot x$, where $\alpha=\omega/|\omega|_B$ for some $\omega \ne 0$. For $x\in B$, $z(x) \in [-1, 1]$. Then, it suffices to analyze the approximation of the sinusoidal function in $[-1,1]$

    Since $g$ is uniformly continuous in $[-1,1]$, it is uniformly well approximated by piecewise constant functions, for any sequence of partitions of $[-1,1]$ into intervals of maximum width tending to 0. Such piecewise functions can be represented as linear combinations of step functions.

    Consider $g(z)$ restricted to $z\in [0,1]$, with $g(0)=0$. For a partition $0=t_0 < t_1 < .. < t_k=1$,
    $$
    g_{k,+}(z) = \sum_{k=1}^{k-1} (g(t_i)-g(t_{i-1})) \, 1_{\{z \ge t_i\}}
    $$
    interpolates the function $g$ at the points $t_i$ as a linear combination of step functions. The derivative of $g$ is bounded by $|\gamma| \le C$, so the sum of the absolute values of the coefficients is also bounded by $C$, i.e. $\sum_i |g(t_i)-g(t_{i-1})| \le C$. 

    In a similar way,
    $$
    g_{k,-}(z) = \sum_{k=1}^{k-1} (g(-t_i)-g(-t_{i-1})) \, 1_{\{z \le -t_i\}}
    $$    

    Adding $g_{h,+}(z)+g_{h,-}(z)$ yields a sequence of piecewise constant functions on $[-1,1]$ that are uniformly close to $g(z)$. They are linear combinations of step functions and the sum of the absolute values of their coefficients is bounded by $2C$. 

    The functions $g(z)$ are, therefore, in the closure of the convex hull of the functions $\gamma \operatorname{step}(z-t)$ and $\gamma \operatorname{step}(-z-t)$ with $|\gamma|\le 2C$ and $t \le 1$, where $\operatorname{step}(z)=1_{z \ge 0}$ denotes the step function:
    \begin{align*}
     G_{step} = \left\{
      \gamma \operatorname{step}(\alpha \cdot x - t):  \right.
      \\ \left. |\gamma| \le 2C,  |\alpha|_B = 1, |t| \le 1 \right\}
    \end{align*}
    
    \end{quote}

    \noindent
    {\it
    $G_{cos}$ is in the closure of the convex hull of $G^\mu_{step}$.
    }

    \begin{quote}
    The result for $G_{step}$ holds if, for each $\alpha$, the parameter $t$ is restricted to a subset $T_\alpha$ that is dense in $[-1,1]$.

    We can restrict $t$ to the continuity points of the distribution $z=\alpha \cdot x$ induced by the measure $\mu$ on $\mathbb R^n$ to obtain the class $G^\mu_{step}$.
    \end{quote}

    \noindent
    {\it
    $G^\mu_{step}$ is in the closure of the convex hull of $G_{\sigma}$.
    }

    \begin{quote}
    The sequence of sigmoidal functions $\sigma(|a|(\alpha \cdot x -t))$, with $|a| \to \infty$, has pointwise limit equal to $ \operatorname{step}(\alpha \cdot x - t)$, except possibly for $x$ in the set with $\alpha \cdot x - t=0$, which has measure zero by the restriction imposed on $t$. By the dominated convergence theorem, the limit also holds in $L_2(\mu,B)$: Functions in $G^\mu_{step}$ are in the closure in $L_2(\mu,B)$ of the class of sigmoidal functions.
    \end{quote}

    Putting together the previous closure results:
    $$
    \Gamma_C^0 
        \subset \overline{G_{cos}} 
        \subset \overline{G^\mu_{step}}
        \subset \overline{G_{\sigma}}
    $$
    where $\overline{G}$ denotes the closure in $L_2(\mu,B)$ of the convex hull of $G$ and $\Gamma_C^0$ is the set of functions in $\Gamma_C$ with $f(0)=0$.
    This completes our proof.

\end{proof}

\begin{theorem}
\label{thm:barron-approximation}
For every function $f \in \Gamma_{C,B}$, every sigmoidal function $\sigma$, every probability measure $\mu$, and every $h \ge 1$, there exists a linear combination of sigmoidal functions $f_h(n)$ such that
$$
\int_B (f(x)-f_h(x))^2 \mu(x) dx \le \frac{(2C)^2}{h}
$$
\end{theorem}

\begin{proof}

The proof is based on convex combinations in Hilbert spaces (Lemma \ref{lemma:barron-hilbert}) and the convex hull of sigmoidal functions (Lemma \ref{lemma:barron-gamma}).

Given a measurable set $B$ containing the point $x=0$, if, for functions in the class $\Gamma_{C,B}$, $\overline f(x)=f(x)-f(0)$ is in the closure of the convex hull of $G_\sigma$ in $L_2(\mu,B)$. Then, by Lemma \ref{lemma:barron-hilbert}, there exists a convex combination of $h$ sigmoidal functions such that the square of the $L_2(\mu,B)$ norm is bounded by a constant divided by $h$.

The constant can be determined to be $(2C)^2$ as follows:

\begin{itemize}

\item
The approximation bound is trivially true if $\| \overline f\|=0$, when $f(x)$ is a constant almost everywhere in $B$. So suppose that $\| \overline f\| > 0$...

\item
When the sigmoidal function is bounded by 1, the functions in $G_\sigma$ are bounded by $b=2C$. Any $c' > (2C)^2-\|\overline f\|^2$ is valid for Lemma \ref{lemma:barron-hilbert}. There is a convex combination of $h$ functions in $G_\sigma$ for which the square of the $L_2(\mu,B)$ norm of the approximation error is bounded by $c'/h$.

\item
If the sigmoidal function is not bounded by 1, combine Lemma \ref{lemma:barron-hilbert} and $\Gamma_C^0 \subset \overline{G^\mu_{step}}$ to obtain a convex combination of $h$ functions in $G^\mu_{step}$ for which the square of the $L_2(\mu,B)$ norm of the approximation error is bounded by $(2C)^2-(1/2)\|\overline f\|^2$ divided by $h$. Then, using $\overline{G^\mu_{step}} \subset \overline{G_{\sigma}}$, by scaling the sigmoid function, sufficiently accurate replacements of the step functions can be obtained so that a convex combination of $h$ functions in $G_{\sigma}$ yields a square $L_2(\mu,B)$ norm of the approximation error bounded by $(2C)^2/h$.

\end{itemize}

This completes Barron's proof \cite{barron1993}.
\end{proof}

\repthm{thm:barron}

\begin{proof}
The Barron norm is
$$
\|f\|_{\mathcal B} := \int_{\mathbb R^n}\|\omega\|\, |\mathcal F(\omega)|\,d\omega 
$$

A Barron function is a function with a finite Barron norm $C_f=\|f\|_{\mathcal B} < \infty$. By the definition of the Barron space, $f \in \Gamma_{C_f,B}$. Therefore, Barron's approximation theorem (Theorem \ref{thm:barron-approximation}) applies: for every $h\ge1$, there exists a linear combination of sigmoidal units $f_h$ such that
$$
\int_B (f(x)-f_h(x))^2\,d\mu(x) \le \frac{(2C_f)^2}{h}
$$

Setting $h=H$, where $H$ is the number of hidden units:
$$
\|f-f_H\|_{L^2(B,\mu)}^2 \le \frac{4C_f^2}{H}
$$

Taking square roots yields
$$
\|f-f_H\|_{L^2(B,\mu)} \le \frac{2C_f}{\sqrt H}
$$

Just by renaming $2C_f$ as $C$, we obtain the desired result: every Barron function admits approximation by a single-hidden-layer neural network with $H$ hidden units at the rate $O(1/\sqrt{H})$.
\end{proof}

\begin{remark}
The dimension $n$ can appear indirectly through the constant $C_f$. Dependence of the constant on $n$ does not affect the convergence rate, but, when $C_f$ is exponentially large in $n$, an exponentially large value of $h$ might be needed for $C_f^2/h$ to be small enough for approximation (an unsurprising result, given that $C_f$ involves an $n$-dimensional integral).
\end{remark}

Feedforward neural networks with a single hidden layer of $h$ sigmoidal units achieve integrated equared error of order $O(1/h)$. Barron found that surprising when compared to the exponential number of parameters required  by polynomial, spline, and trigonometric expansions. The neural network just needs $(n+2)*h+1$ parameters.\footnote{Traditional expansions take a linear combination of a set of fixed basis functions (only the parameters of the linear combination are adjusted). Adjustable basis functions (as in neural networks) provide an ``striking advantage'' \cite{barron1993}.}
\subsection*{Montúfar on the Power of Deep Networks}

In a NIPS'2004 paper (now NeurIPS), Guido Montúfar, Razvan Pascanu, Kyunghyun Cho, and Yoshua Bengio studied the complexity of the functions that are computable by deep feedforward neural networks with piecewise linear activations (e.g. ReLU) in terms of the symmetries and the number of linear regions that they have \cite{montufar2014}.

\begin{definition}
A map $F$ identifies two neighborhoods $S$ and $T$ of its input domain if it maps them
to a common subset $F(S) = F(T)$ of its output domain. In this case, we say that $S$ and $T$ are identified by $F$.
\end{definition}

\begin{lemma}
The maximal number of linear regions of the functions computed by an $L$-layer neural
network with piecewise linear activations is at least $\mathcal N^L = \sum_{R \in P^{L-1}} \mathcal N_R^{L-1}$, 
where $\mathcal N_R^0 = 1$ for each region $R \subseteq \mathbb R^{n_0}$
and $P^L$ denotes the set of neighbordhoods in distinct linear regions of the function computed by the $L$-th layer.
\end{lemma}

\repthm{thm:montufar}

\begin{proof}

Montúfar's proof, included in their arXiv technical report \cite{montufar2014arxiv}, counts the number of regions by dividing the first $L-1$ network layers into $n_0$ independent parts, one for each input neuron (i.e. each part is sensitive only to one coordinate of the input-space).

The units of each layer are organized into $n_0$ non-empty groups of units of sizes
$p_1 = .. = p_{n_0} = \lfloor n_l / n_0 \rfloor$ for a layer of width $n_l$. 

The input weights of each group are defined in such a way that the units in that group are sensitive to only one coordinate of the $n_0$-dimensional input-space. 
By choosing the input and bias weights in the right way, the alternating sum of the activations of the $p$ units within one group folds their input-space coordinate $p$ times into itself.

\begin{quote}
    Consider a layer of $n$ rectifiers with $n_0$ input variables, where $n \ge n_0$. We partition the set of rectifier units into $n_0$ (non-overlapping) subsets of cardinality $p = \lfloor n/n_0 \rfloor$. Weights and biases in each group are chosen as follows:
    \begin{align*}
      h_1(x) &= \max \{ 0, wx \} \\
      h_2(x) &= \max \{ 0, 2wx - 1\} \\
      h_3(x) &= \max \{ 0, 2wx - 2\} \\
      &... \\
      h_p(x) &= \max \{ 0, 2wx - (p-1)\} 
    \end{align*}
    where $w$ is a vector whose $j$-th entry is 1 and all other entries are 0. The inner product $wx$ selects the $j$-th coordinate of $x$.

    Adding these rectifiers with alternating signs, we obtain the scalar function $\tilde h_j(x)$:
    \begin{align*}
    \mathbf 1_{alt} &= [1,-1,1,...,(-1)^{p-1}] \\
    \tilde h_j(x) &= \mathbf 1_{alt} \cdot [h_1(x), h_2(x), h_3(x),..., h_p(x)]
    \end{align*}

    The function $\tilde h_j(x)$ acts only on the $j$-th coordinate of $x$, so it can be interpreted as a scalar function. This scalar function acts on $p$ intervals: $(-\infty,0], [0,1], [1,2]...[p-1,\infty)$. Each of those intervals are mapped by $\tilde h(x)$ onto the interval $(0,1)$.

    If we consider the $n_0$ sets of rectifiers, the function $\tilde h = [\tilde h_1, \tilde h_2, \tilde h_3, ..., \tilde h_p]$ is locally symmetric about each hyperplane with a fixed $j$-th coordinate equal to $x_j = 1 .. p$. A periodic pattern emerges and the function $\tilde h$ identifies $p^{n_0}$ hypercubes delimited by these hyperplanes.

    NOTE: $\tilde h$ arises from $h$ by composition with a linear function $\mathbf 1_{alt}$ (alternating sums). This linear function can be effectively absorbed in the preactivation function of the next layer. Hence, we can treat $\tilde h$ as the function computed by the current layer. Computations by deeper layers, as functions of the unit hypercube output of this rectifier layer, are replicated on each of the $p^{n_0}$ input-space hypercubes.
\end{quote}

Once the parameters of each group are chosen so that they fold its 1D input-space many times into itself, the alternating sum $\tilde h = h_1 - h_2 + h_3 - .. \pm h_p$ 
of the activations of the $p$ units in a group as the activation of a fictitious intermediary unit. The compound output of all these $n_0$ intermediary units partitions the input-space, $\mathbb R^{n_0}$, into a grid of $\prod_{i=1}^{n_0} p_i$ regions.  Each of these input-space regions is mapped to the $n_0$-dimensional unit cube in the output-space of the intermediary layer. This unit cube is the effective input for the next hidden layer. With each layer of the network, the number of identified regions is multiplied by $\prod_{i=1}^{n_0} p_i$.

The function $\tilde h$ computes an alternating sum that folds the input-coordinate $p$ times into the interval $(0, 1)$. The function $\tilde h_j(x)$ identifies the input-space strips with j-th coordinate $x_j$ restricted to the intervals $(0,1), (1,2) ... (p-1, p)$. Consider one of those intervals $(k-1,k)$. For $x\in(k-1,k)$, $\tilde h (x) =x + 2\sum_{i=1}^{k-1} (-1)^i (x-i) = (-1)^{k-1}(x-(k-1)) - ((-1)^{k-1}-1)/2$. Therefore $\tilde h(k-1) = - ((-1)^{k-1}-1)/2$ and $\tilde h(k) = (-1)^{k-1}(x-(k-1))$. One of them is always $0$, the other is $1$, so $\tilde h(\{ k-1, k\}) = \{0,1\}$. Since the function is linear in the interval $(k-1,k)$, $\tilde h$ maps the interval $(k-1,k)$ onto the inverval $(0,1)$.

The number of input-space neighborhoods that are mapped by the first $L-1$ layers onto the (open) unit hypercube $(0,1)^{n_0}$ of the (effective) output space of the $(L-1)$-th layer is given by
$$
\mathcal N_{(0,1)^{n_0}}^{L-1} = \prod_{l=0}^{L-1} \prod_{i=1}^{n_0} p_{l,i}
$$
where $p{l,i}$ is the number of units in the $i$-th group of units in the $l$-th layer.

The inputs and bias of the last hidden layer can be chosen in such a way that the function $h_L$ partitions its (effective) input neighborhood $(0,1)^{n_0}$ by an arrangement of $n_L$ hyperplanes in general position, i.e., into $\sum_{j=0}^{n_0} {n_L \choose j}$ regions.

\begin{quote}
Rectifier units can either be constant or linear. The boundary between those two behaviors is given by a hyperplane. The collection of hyperplanes derived from all the units in the same layer forms a hyperplane arrangement that splits the input-space into several regions. An arrangement of $n$ hyperplanes in $\mathbb R^{n_0}$ has at most $\sum_{j=0}^{n_0} {n \choose j}$ regions \cite{pascanu2013arxiv}.

This implies that shallow rectifier neural networks with $n_0$ inputs and $n_1$ hidden units can only identify $\sum_{j=0}^{n_0} {n_1 \choose j}$ different regions.
\end{quote}

Let $m_l$ denote the remainder of $n_l$ divided by $n_0$. Choosing $p_{l,1} = .. = p_{l,n_0-m_l} = \lfloor n_l / n_0 \rfloor$ and $p_{l,n_0-m_l+1} = .. = p_{l,n_0} = \lfloor n_l / n_0 \rfloor+1$, we obtain a total of linear regions
$$
\left( \prod_{l=1}^{L-1} 
      \left\lfloor \frac{n_l}{n_0} \right\rfloor^{n_0-m_l}
      \left( \left\lfloor \frac{n_l}{n_0} \right\rfloor + 1\right)^{m_l}
\right) \sum_{j=0}^{n_0} {n_L \choose j}
$$

This is equal to the bound given in theorem when all remainders $m_l = n_l - n_0 \lfloor n_l / n_0 \rfloor$ are zero. Otherwise, it is larger. QED.
\end{proof}

The previous result, stated in terms of the maximal number of linear regions of the functions computable by deep neural networks, can be expressed in terms of the number of parameters of neural networks:

\begin{itemize}

\item
For shallow models (single-hidden-layer networks) with $n$ hidden units, the number of parameters is $\Theta(Ln)$ and the number of linear regions per parameter is given by
$$
O \left( L^{n_0-1} n^{n_0-1} \right)
$$ 

\item
For deep models of $L$ layers of width $n$, the number of parameters grows as $\Theta(Ln^2)$ and the number of linear regions per parameter grows as 
$$
\Omega \left( \left(\frac{n}{n_0} \right)^{n_0(L-1)} \frac{n^{n_0-2}}{L} \right)
$$
\end{itemize}

\begin{proof}

For shallow models:

\begin{itemize}
\item
Number of regions: $L=1$, so 
$$
R \ge \sum_{j=0}^{n_0}\binom{n}{j} \in \Theta(n^{n_0})
$$
given that 
$$
\binom{n}{n_0} = \frac{n^{n_0}}{n_0!} +O(n^{n_0-1})
$$

\item
Number of parameters: A single-hidden-layer network with input dimension $n_0$, width $n$, and output dimension $1$ has $P=(n_0+1)n+(n+1) \in \Theta(n)$ parameters, $\Theta(Ln)$ given that $L=1$

\item
Efficiency:
\begin{align*}
\frac{R}{P} \in \Theta\!\left( \frac{n^{n_0}}{Ln} \right) 
  & = \Theta\!\left( L^{-1} n^{n_0-1} \right) \\
  & \subset  O\!\left( L^{\,n_0-1}n^{\,n_0-1} \right)
\end{align*}

\end{itemize}

For deep models:

\begin{itemize}

\item
Regions:
$$
R \ge \left( \left\lfloor\frac{n}{n_0}\right\rfloor^{n_0} \right)^{L-1} \sum_{j=0}^{n_0}\binom{n}{j}
$$
Since
$$
\left\lfloor\frac{n}{n_0}\right\rfloor \in \Theta\!\left(\frac{n}{n_0}\right)
$$
and
$$
\sum_{j=0}^{n_0}\binom{n}{j} \in \Theta(n^{n_0})
$$
then
$$
R \in \Omega\!\left( \left(\frac{n}{n_0}\right)^{n_0(L-1)} n^{n_0} \right)
$$

\item
Parameters: A depth-$L$ network of constant width $n$ has $P = n_0n +(L-1)n^2 + n + O(Ln) \in \Theta(Ln^2)$ parameters.

\item
Efficiency:
\begin{align*}
\frac{R}{P} 
&\in \Omega\!\left( \frac{ \left(\frac{n}{n_0}\right)^{n_0(L-1)} n^{n_0} }{ Ln^2 } \right) \\
&= \Omega\!\left( \left(\frac{n}{n_0}\right)^{n_0(L-1)} \frac{n^{n_0-2}}{L} \right)
\end{align*}

\end{itemize}

\end{proof}

Shallow networks achieve only polynomial growth in the number of linear regions per parameter, whereas deep networks achieve an exponential growth in depth $L$, demonstrating the substantially greater expressivity of deep ReLU architectures. 


\bibliographystyle{acm}
\bibliography{references}

\end{document}